\documentclass[10pt,twocolumn,letterpaper]{article}

\usepackage{wacv}              

\definecolor{wacvblue}{rgb}{0.21,0.49,0.74}
\usepackage[pagebackref,breaklinks,colorlinks,allcolors=wacvblue]{hyperref}

\usepackage[accsupp]{axessibility}  

\usepackage{kotex}
\usepackage{comment}
\usepackage{mathtools}
\usepackage{amssymb}
\usepackage{graphicx}
\usepackage{algpseudocode}
\usepackage{caption}
\usepackage{subcaption}
\usepackage{multirow}
\usepackage{tablefootnote}
\usepackage{wrapfig}
\usepackage{textcomp}
\usepackage{siunitx}
\usepackage{multirow}
\usepackage{wrapfig}
\usepackage{booktabs}
\usepackage{graphics}
\usepackage{multibib}
\usepackage{algorithm}
\usepackage{arydshln}
\usepackage{nicematrix}
\usepackage{soul}

\newtheorem{define}{Definition}

\newcommand{\jy}[1]{\textcolor{blue}{#1}}
\newcommand{\yr}[1]{\textcolor{olive}{#1}}
\newcommand{\nj}[1]{\textcolor{TealBlue}{#1}}

\def\wacvPaperID{996} 
\def\confName{WACV}
\def\confYear{2026}

\title{Bi-ICE: An Inner Interpretable Framework for Image Classification via Bi-directional Interactions between Concept and Input Embeddings}

\author{Jinyung Hong\textsuperscript{1}\thanks{Both contributed equally to this work as co–first authors. email: \url{jhong53@asu.edu}, \url{yerim1656@snu.ac.kr}}, 
Yearim Kim\textsuperscript{2*}, Keun Hee Park\textsuperscript{1}, Sangyu Han\textsuperscript{3}, Nojun Kwak\textsuperscript{2, 3}\thanks{Both provided equal contributions in advising this work.}, Theodore P.~Pavlic\textsuperscript{1, 4$\dagger$} \\
\textsuperscript{1}School of Computing and Augmented Intelligence \\
\textsuperscript{4}School of Life Sciences \\
Arizona State University, Tempe, AZ 85281\\
\textsuperscript{2}Department of Intelligence and Convergence \\
\textsuperscript{3} Interdisciplinary Program in Artficial Intelligence \\
Seoul National University, Seoul, 08826, South Korea \\
}

\begin{document}
\maketitle
\begin{abstract}
Inner interpretability is a promising field aiming to uncover the internal mechanisms of AI systems through scalable, automated methods.
While significant research has been conducted on large language models, limited attention has been paid to applying inner interpretability to large-scale image tasks, focusing primarily on architectural and functional levels to visualize learned concepts.
%
In this paper, we first present a conceptual framework that supports inner interpretability and multilevel analysis for large-scale image classification tasks.
Specifically, we introduce the \textbf{Bi-directional Interaction between Concept and Input Embeddings~(Bi-ICE) module}, which facilitates interpretability across the computational, algorithmic, and implementation levels.
This module enhances transparency by generating predictions based on human-understandable concepts, quantifying their contributions, and localizing them within the inputs.
Finally, we showcase enhanced transparency in image classification, measuring concept contributions, and pinpointing their locations within the inputs. Our approach highlights algorithmic interpretability by demonstrating the process of concept learning and its convergence.
\end{abstract}    
\section{Introduction}
\label{sec:intro}
In recent years, the area of study,~\emph{Inner Interpretability}~\cite{rauker2023toward}, has been a growing body of research focused on understanding the internal structural components~\cite{geva2021transformer, elhage2021mathematical}, functions~\cite{todd2024function}, algorithms~\cite{zhong2023clock}, and representations~\cite{hernandez2023inspecting} of neural networks. 
The field primarily seeks to explain how the internal mechanisms of these models contribute to their capabilities. 
In practical applications, however, many studies on inner interpretability appear to equate mechanistic explanations exclusively with identifying model components that causally influence specific behaviors of interest~\cite{meng2022locating}.
Moreover, the mechanistic proposals presented in these studies often contain gaps, rendering them incomplete~\cite{merullo2023circuit}.
Incomplete mechanistic explanations can easily lead to the mischaracterization of elements within the mechanism~\cite{craver2006mechanistic}.
To address these issues, there have been recent works~\citep{vilasposition, he2024multilevel} on inner interpretability frameworks integrated with multilevel analysis inspired by cognitive science, enabling us to build better and more complete mechanistic explanations.  

Multilevel analysis offers a principled framework to systematically examine model behavior across three distinct but interrelated levels: computational, algorithmic, and implementation~\citep{marr1976understanding, marr2010vision}.
At the computational level, the focus is on defining what problem the model is solving, providing a high-level description of its functional objectives.
The algorithmic level then describes how the model processes and transforms information to achieve these objectives, identifying the structures and operations responsible for learning and reasoning. 
Finally, the implementation level investigates how these transformations are instantiated within the model’s architecture, examining how neurons, circuits, and weight distributions support the computations described at higher levels~\citep{bechtel2015non, vilasposition, griffiths2024bayes}.

By explicitly linking these levels, multilevel analysis prevents the mischaracterization of mechanistic components and ensures causal explanations reach beyond isolated interventions.
This structured approach not only enforces internal consistency across different levels of explanation but also enables a deeper and more coherent understanding of how neural networks generate their emergent capabilities.
Integrating multilevel analysis into inner interpretability thus represents a crucial step toward developing more complete mechanistic explanations of deep learning models~\citep{vilasposition, he2024multilevel}.
To date, frameworks have largely centered on LLMs; image models are comparatively underexplored.
\begin{figure}[t!]
    \centering
    \includegraphics[width=0.98\columnwidth]{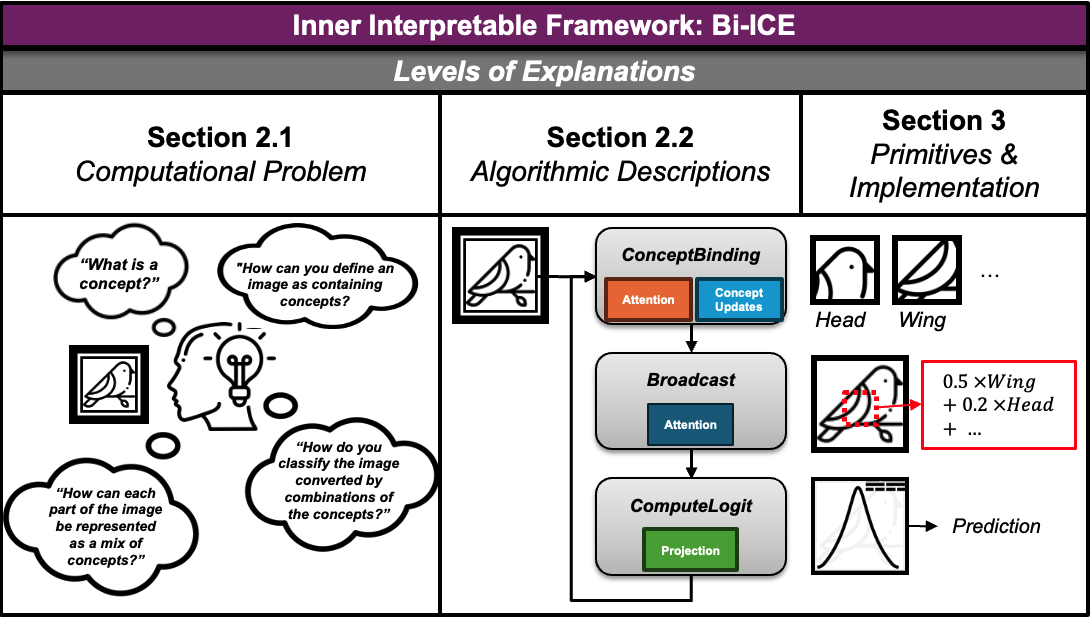}
    \caption{Illustration of our inner interpretability framework and our proposed module, Bi-ICE.}
    \label{fig:main}
    \vspace{-5mm}
\end{figure}

Therefore, we aim to move beyond these fragmented approaches by integrating all three levels—computational, algorithmic, and implementation—offering a more complete and mechanistic understanding of the behavior of models trained by image data~(Fig.~\ref{fig:main}). 
Here, we establish a theoretical foundation by defining Computational problem~(Sec.~\ref{subsec:computational}), 
and then provide Algorithmic descriptions (Sec.~\ref{subsec:algorithmic}), including the core procedures, \textsc{ConceptBinding}, \textsc{Broadcast}, and \textsc{ComputeLogit}.

Furthermore, we bridge the gap between theoretical frameworks and practical methods by proposing \textbf{Bi-directional Interaction between Concept and Input Embeddings~(Bi-ICE)} (Sec.~\ref{sec:impl}). 
Acting as a central hub in the model, our Bi-ICE collects image embeddings from the backbone models, processes them collectively, and broadcasts the refined information back to the network, enabling coordinated decision making.
By attaching Bi-ICE to the pretrained models, we can address all levels of inner interpretability, validating the capability of our module across multilevel analysis (Sec.~\ref{sec:exp}).
We can improve model transparency by making predictions based on human-understandable concepts and showing the process of concept learning and convergence.

\vspace{-2mm}
\section{Conceptual Multilevel Framework}
\label{sec:multilevel_framework}

Here, we present a conceptual framework inspired by \citet{vilasposition} and \citet{he2024multilevel} for our problem, \textbf{image classification via concept decomposition}.
We address the computational (Sec.~\ref{subsec:computational}) and algorithmic (Sec.~\ref{subsec:algorithmic}) levels, which define the underlying problems and outline the method to solve them.
These foundational levels set the stage for the practical implementation, detailed in Sec.~\ref{sec:impl}.


\subsection{Computational Problems}
\label{subsec:computational}
We aim to solve an image classification task via concept decomposition.
Let $\mathcal{S} = \{ (x^{l}, y^{l}) \}_{l=1}^{M}$ be the training set,
where each input image $x^{l} \in \mathcal{X} \subset \mathbb{R}^{n}$ is associated with a class label $y^{l} = \{ 1, \cdots, N \} \in \mathcal{Y}$, a one-hot vector of length $N$, the number of classes.
For simplicity, we omit the instance index $l$ henceforth, focusing on a single image and its patches. 
Thus, $x, y$, and $z$ will denote an individual image, its label, and its embedding, respectively.

Given an input image $x$, the task is to use its embedding $z$, which is reconstructed with the mixture of the learned concept representations, to predict the conditional probability of a class label $y$.
This leverages the concept decomposition of the embedding, making the classification process more interpretable.
To define our computational problem, we first establish a series of fundamental definitions.
\begin{define}[Concept]
\label{def:concept}
A \emph{concept} $c \in \mathbb{C}$ is an interpretable semantic feature that captures a distinct and meaningful characteristic within the data.
The set of all concepts is denoted as $\mathbb{C} = \{c_1, ..., c_K\}$, where $K$ denotes the total number of distinct concepts.
\end{define}
\begin{define}[Concept Vector]
\label{def:concept vector}
A \emph{concept vector} $\boldsymbol{\zeta} \in \mathbb{R}^D$ is the 
representation of a concept $c \in \mathbb{C}$.
The mapping $R:\mathbb{C} \rightarrow \mathbb{R}^D$ assigns each concept $c$ to its corresponding vector representation $\boldsymbol{\zeta}_c = R(c)$, capturing the semantic meaning of $c$ in the embedding space of dimension $D$.
\end{define}
\begin{define}[Concept Score]
\label{def:concept score}
Given a sample embedding $\boldsymbol{z} = f(x) = \{ z_{1}, \dots, z_{L} \}   \in \mathbb{R}^{L \times D}$, where $f(\cdot)$ is the backbone network to produce the embedding representation of the input, $L$ is the number of input patches, and $D$ is the embedding dimension, the \emph{concept score} for each patch embedding with respect to a concept $c$ is defined as $s(z_{i}, c) \triangleq \operatorname{Sim}(z_{i}, \boldsymbol{\zeta_c})$.
Here, $\operatorname{Sim}(\cdot)$ denotes a similarity measure, such as cosine similarity or dot product.
\end{define}
\begin{define}[Image-patch-specific Concept Decomposition]
\label{def:concept decomposition}
Given a sample with its associated concept set $C \subseteq \mathbb{C}$, each patch embedding $z_i \in \mathbb{R}^D$ can be expressed as a weighted sum of concept representations:
\begin{equation*}
    z_{i} = \sum_{c \in C} \lambda_{i,c} ~ \zeta_c, \; \text{s.t.} \; \lambda_{i,c} = s(z_{i}, c) \geq 0 \;, \; \sum_{c \in C} \lambda_{i,c} = 1.
\end{equation*}
\end{define}
\begin{define}[Classification with Concept Decomposition]
\label{def:classification}
Given a sample embedding $\boldsymbol{z} \in \mathbb{R}^{L \times D}$ and a projection layer $P: \mathbb{R}^{D} \rightarrow \mathbb{R}^{N}$ for $\mathcal{Y}= \{1, 2, ..., N\}$, the classification logit for class $n \in \mathcal{Y}$ is calculated by: $\operatorname{Pr}(y | \boldsymbol{z}) = 
     \operatorname{softmax}(\frac{1}{L} \sum_{l=1}^L z_{l} \cdot P_{:, n}),$
where $P_{:, n} \in \mathbb{R}^{D}$ means the $n$-th column of the linear weight $P \in \mathbb{R}^{D \times N}$.
\end{define}

Building upon the definitions above, we formulate the core computational problem of our framework: image classification based on concept decomposition.
The goal is to use the concept-based representation of the image to predict the class label.
We outline this formal problem as follows:
\noindent 
\begin{description}
    \item \textbf{Problem:} \textsc{Classification-based-concept $C$} \\
        Input: embedding $\boldsymbol{z}$ from an image $x \in \mathcal{X}$ \\
        Output: conditional probability of a class $y, \operatorname{Pr}(y | \boldsymbol{z})$
        \label{def:classification_based_c}
\end{description}

Based on the above definitions and problem setup, we analyze our model by evaluating how each patch embedding comprises concepts for decision-making~(Sec.~\ref{subsubsec:comp}).

%

\subsection{Algorithmic Descriptions}
\label{subsec:algorithmic}
The Concept Decomposition algorithm~(Algorithm~\ref{alg:concept_extraction}) is our candidate algorithm for the computational problem \textsc{Classification-Based-Concept $C$}. 
We employ three key procedures in this algorithm: \textsc{ConceptBinding}, \textsc{Broadcast}, and \textsc{ComputeLogit}. 
It consists of a series of steps that extract the decomposed concepts related to the input embedding and then update the embedding representation by the combinations of the concepts. 
This bi-directional embedding update for both concept $\boldsymbol{\zeta}$ and input embedding $\boldsymbol{z}$ is the key computation to satisfy the definition of the computation problem, and the interaction between the input can be achieved by a working memory module that will be explained in detail in Sec.~\ref{sec:impl}.
Finally, the last step outputs the classification logits based on the updated embeddings for predictions.

This allows us to evaluate the learning dynamics of concept convergence by observing the interaction between embedding and concept vectors at every epoch~(Sec.~\ref{subsubsec:alg}).

\begin{algorithm}[h!]
\caption{Classification via Concept Decomposition}
\label{alg:concept_extraction}
\begin{algorithmic}
\Require image embedding $\boldsymbol{z}$, initial concept vectors $\boldsymbol{\zeta}_{0} \in \mathbb{R}^{K \times D}$
\For{$t = 1, \dots, T$}
    \State $\boldsymbol{\zeta}_{t} \leftarrow \operatorname{ConceptBinding}(\boldsymbol{z}, \boldsymbol{\zeta}_{t-1})$ 
    \Comment{Sec.~\ref{subsubsec: conceptbinding}}
    \State $\bar{\boldsymbol{z}} \leftarrow \operatorname{Broadcast}(\boldsymbol{z}, \boldsymbol{\zeta}_{t})$ 
    \Comment{Sec.~\ref{subsubsec: broadcast}}
    \State $l_{y} \leftarrow \operatorname{ComputeLogit} (\bar{\boldsymbol{z}})$
    \Comment{Sec.~\ref{subsubsec: logitcompute}}
\EndFor 
\end{algorithmic}
\end{algorithm}
\section{Primitives \& Implementation}
\label{sec:impl}
In this section, we present methodological strategies based on the conceptual framework~(Sec.~\ref{sec:multilevel_framework}).
The implementation level characterizes how the primitives and operations are implemented in the model.
Here, we present our deep-learning-based framework inspired by Shared Global Workspace~(SGW)~\citep{goyal2021coordination}.
This framework includes a working memory module that acts as an information bottleneck, enabling interaction among specialized computing modules via shared memory~\citep{goyal2021coordination}.
Research shows that attention-based methods~\citep{goyal2021coordination, hong2024concept} effectively implement SGW-inspired deep learning, making these submodules well-suited for essential operations within our framework.

\begin{figure*}[t!]
     \centering
     \begin{subfigure}[b]{0.495\textwidth}
         \centering
         \includegraphics[width=\textwidth]{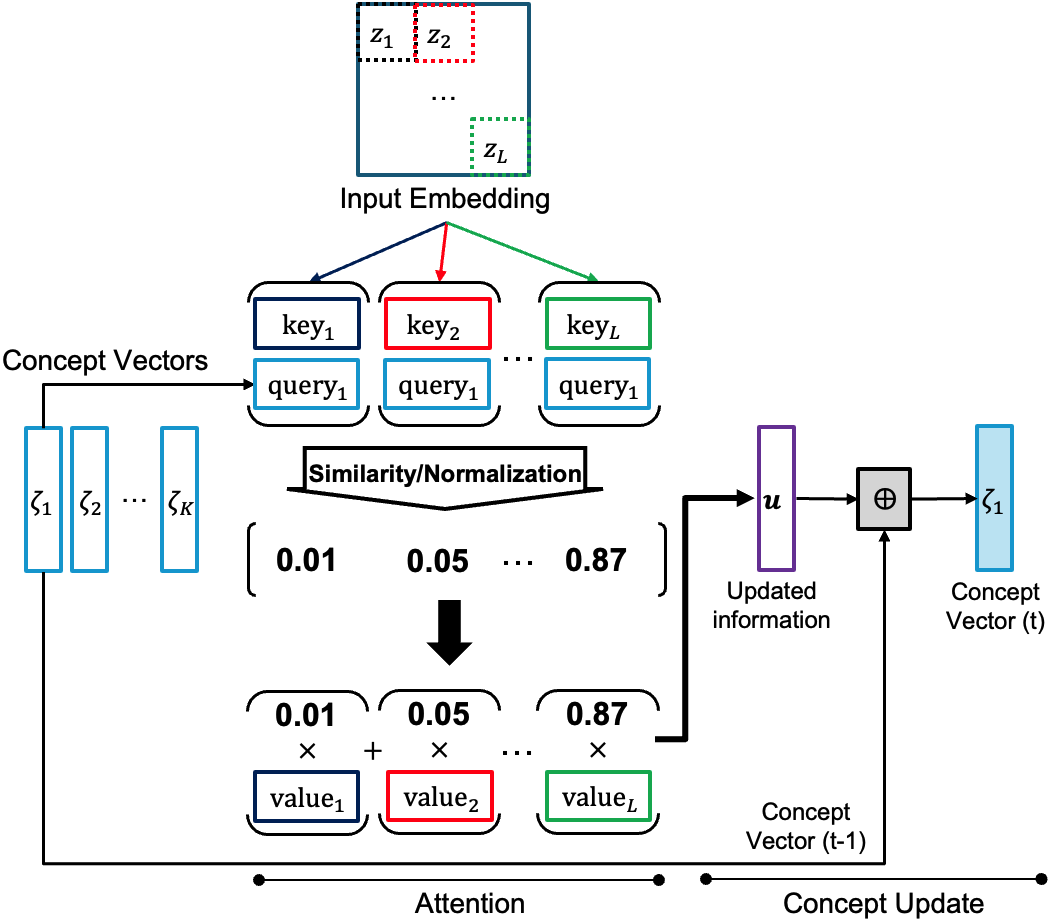}
         \caption{Primitives and Operations for \textsc{ConceptBinding}}
         \label{fig:primitive_concept-binding}
     \end{subfigure}
     \hspace{2pt}
     \begin{subfigure}[b]{0.40\textwidth}
         \centering
         \includegraphics[width=0.9\textwidth]{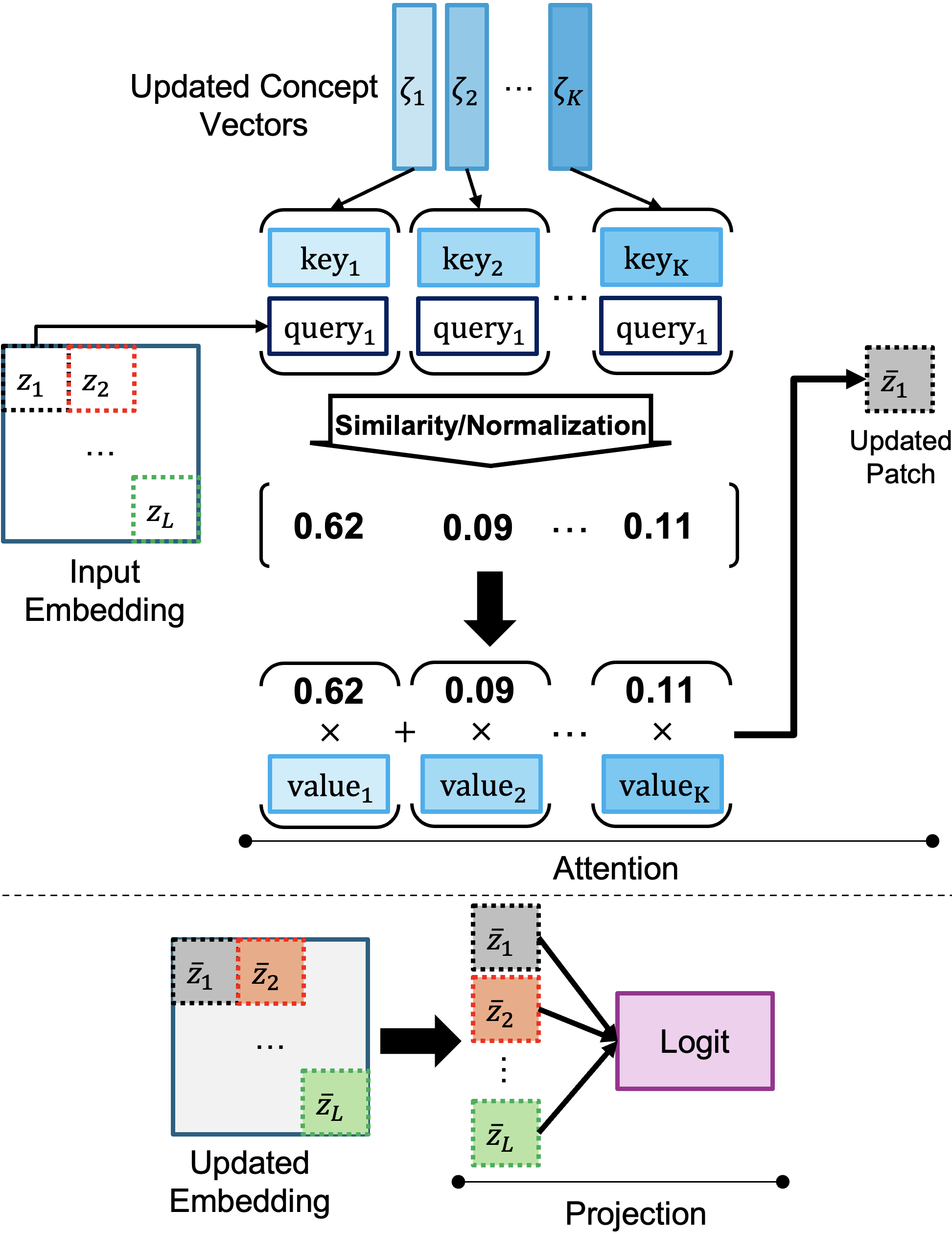}
         \caption{Primitives and Operations for \textsc{Broadcast} (\textbf{Top}) and \textsc{ComputeLogit} (\textbf{Bottom})}
    \label{fig:primitive_broadcast_compute-logit}
     \end{subfigure}
    \caption{Primitives and Operations used as the essential components to implement our Bi-ICE module. } 
    \label{fig:primitives}
    \vspace{-5mm}
\end{figure*}
%

\subsection{Primitive Representations \& Operations}
\label{subsec:primitive}
In our problem, image classification via concept decomposition, the primitive candidate 
is our \textbf{Bi-directional Interaction between Concept and Input Embeddings~(Bi-ICE)} module using an attention mechanism~\citep{vaswani2017attention}~(Fig.~\ref{fig:primitives}). 

The process begins with the \textsc{ConceptBinding} step~(Fig.~\ref{fig:primitive_concept-binding}), where the concept vectors $\boldsymbol{\zeta}_t$ at epoch $t$ are updated based on the interaction between the former concept vectors $\boldsymbol{\zeta}_{t-1}$ and the input image embedding $\boldsymbol{z}$.
This procedure refines the concept representations by associating them with relevant features from the input, allowing the model to dynamically adjust its understanding of the concepts.
Next, the \textsc{Broadcast} step~(Top, Fig.~\ref{fig:primitive_broadcast_compute-logit}) integrates the refined concept vectors $\boldsymbol{\zeta}_t$ back into the image embedding ${\boldsymbol{z}}$, producing an enhanced representation $\bar{\boldsymbol{z}}$.
This shared representation facilitates developmental interpretability by capturing the updating interactions between the concept vectors and image embedding at every epoch and then showing the evolution of concept learning over time.
Finally, the \textsc{ComputeLogit} procedure~(Bottom, Fig.~\ref{fig:primitive_broadcast_compute-logit}) generates the classification logits $l_{y}$ with the updated embedding, $\bar{\boldsymbol{z}}$.

\subsection{Implementation Detail}
\label{subsec:impl_detail}

\begin{figure}
    \centering
    \includegraphics[width=0.95\columnwidth]{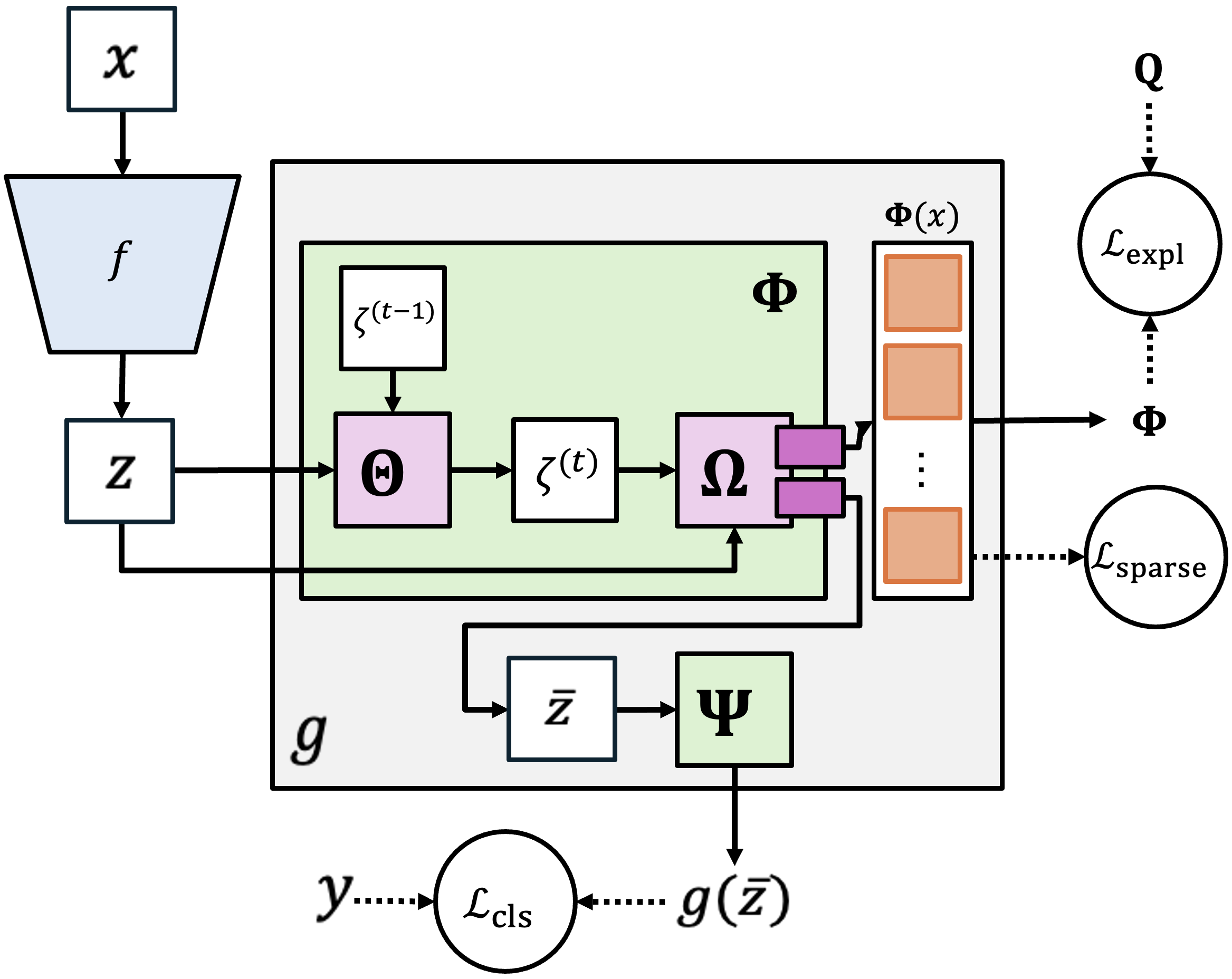}
    \caption{Implementation overview. Each component is detailed in Sec.~\ref{subsec:impl_detail}. The training objectives are explained in Sec.~\ref{subsec:objectives}.}
    \label{fig:method}
    \vspace{-5mm}
\end{figure}

With the primitives and their operations above, we now turn to the implementation details that support the model's interpretability~(Fig.~\ref{fig:method}).
The following outlines how the design of the proposed system inherently enhances interpretability, enabling clear and structured pathways for concept representation and decision-making.

\subsubsection{Concept Binding and Refinement Module \texorpdfstring{$\boldsymbol{\Theta}$}{TEXT}}
\label{subsubsec: conceptbinding}

With the number of concepts $K$, the initial learnable concept vectors $\boldsymbol{\zeta}^{(0)} = \{ \zeta_{1}, \dots, \zeta_{K} \} \in \mathbb{R}^{K \times D}$ perform competitive attention on the input features $\boldsymbol{z} \in \mathbb{R}^{L \times D}$ with the linear projections, $k^{\boldsymbol{\Theta}}(\boldsymbol{z}), q^{\boldsymbol{\Theta}}(\boldsymbol{\zeta})$, and $v^{\boldsymbol{\Theta}}(\boldsymbol{z})$.
This produces the attention matrix $\mathbf{A}^{\boldsymbol{\Theta}} \in \mathbb{R}^{K \times L}_{+}$, where each entry $\mathbf{A}_{k,l}^{\boldsymbol{\Theta}}$ is the attention weight of concept vector ${\zeta}_k$ attending over the input vector $z_l$.
We then normalize $\mathbf{A}^{\boldsymbol{\Theta}}$ by applying softmax across concept vectors, i.e., along the axis $K$:
\begin{equation}
    \mathbf{A}^{\boldsymbol{\Theta}} = \operatorname{softmax}_{K}(q^{\boldsymbol{\Theta}}(\boldsymbol{\zeta}) \cdot k^{\boldsymbol{\Theta}}(\boldsymbol{z})^{\top}/ \sqrt{D}). 
\end{equation}
This implements a form of competition among concept vectors to attend to each input embedding. 

Next, we group and aggregate the attended embeddings to quantify the semantic representation of each concept. 
For this, we normalize 
$\mathbf{A}^{\boldsymbol{\Theta}}$ 
along the axis $L$ and multiply it by input embedding values $v^{\boldsymbol{\Theta}}(\boldsymbol{z}) \in \mathbb{R}^{L \times D}$.
This produces the attention readout in the form of a matrix $\boldsymbol{u} \in \mathbb{R}^{K \times D}$ where each row $\boldsymbol{u}_{k} \in \mathbb{R}^{D}$ is the readout for concept vector $\zeta_k$:
\begin{equation}
\mathbf{A}_{k,l}^{\boldsymbol{\Theta}} = \mathbf{A}_{k,l}^{\boldsymbol{\Theta}}/ \sum_{i=1}^{L} \mathbf{A}_{k,i}^{\boldsymbol{\Theta}},\quad
\boldsymbol{u} = \mathbf{A}^{\boldsymbol{\Theta}} \cdot v^{\boldsymbol{\Theta}}(\boldsymbol{z}).
\end{equation}
We use the readout information obtained from concept binding and update each concept vector.
This aggregated update $\boldsymbol{u}$ is used to refine the concept vectors through a recurrent update, where we use a Gated Recurrent Unit~(GRU)~\cite{cho2014learning}:
$\boldsymbol{\zeta} = \mathtt{GRU}(\boldsymbol{\zeta}, \boldsymbol{u})$.
Thus, the module $\boldsymbol{\Theta}$ produces the updated concept vectors as follows:
%
\begin{equation}
    \boldsymbol{\zeta}^{(t)} = \boldsymbol{\Theta}(\boldsymbol{\zeta}^{(t-1)}, \boldsymbol{u}) \in \mathbb{R}^{K \times D}.
    \label{eq:concept_refinement}
\end{equation}
%

\subsubsection{Broadcast Module \texorpdfstring{$\boldsymbol{\Omega}$}{TEXT}} 
\label{subsubsec: broadcast}
The cross-attention mechanism in the \textsc{Broadcast} module $\boldsymbol{\Omega}$ integrates the refined concept vector $\zeta$ back into the input embedding $z$.
The concept composition score $\boldsymbol{\Phi}(x)$ is calculated as:
%
\begin{equation}
    \boldsymbol{\Phi}(x) \triangleq
    \operatorname{softmax}_{L}\left(q^{\boldsymbol{\Omega}}(\boldsymbol{z}) \cdot k^{\boldsymbol{\Omega}}(\boldsymbol{\zeta})^{\top} / \sqrt{D}\right) \in \mathbb{R}^{L \times K}_{+}
    \label{eq:concept_activation}
\end{equation}
between each patch--concept vector pair. 
The resultant matrix represents patch-wise non-negative concept relevance scores conditioned on input $x$. 
The updated embedding $\Bar{\boldsymbol{z}}$ is then obtained by aggregating the concept values:
%
\begin{equation}
    \bar{\boldsymbol{z}} = 
    \boldsymbol{\Phi}(x) \cdot v^{\boldsymbol{\Omega}}(\boldsymbol{\zeta}) \in \mathbb{R}^{L \times D}.
    \label{eq:broadcast}
\end{equation}
%

Then, the following module $\boldsymbol{\Psi}$ produces classification logits using the updated embedding $\Bar{\boldsymbol{z}}$. 
We evaluate our model using the concept composition score $\boldsymbol{\Phi}$ by visualizing the most important concept for each patch in an input image~(Sec.~\ref{subsubsec:impl}).

\subsubsection{Logit Computational Module \texorpdfstring{$\boldsymbol{\Psi}$}{TEXT}}
\label{subsubsec: logitcompute}
The \textsc{ComputeLogit} module $\boldsymbol{\Psi}$ computes prediction logits using the updated embedding $\Bar{\boldsymbol{z}}$ from Eq.~(\ref{eq:broadcast}).
The output of $\boldsymbol{\Psi}$ is derived by multiplying the updated embedding $\Bar{\boldsymbol{z}}$ by the weights in a linear layer $\boldsymbol{P} \in \mathbb{R}^{D \times N}$.
For $i=1,\dots, N$:
\begin{equation*}
\begin{split}
    \operatorname{logit}_{i}   
    & = \boldsymbol{\Psi}\left(\bar{\boldsymbol{z}}_l; \boldsymbol{P}\right) =  
    \frac{1}{L}\textstyle\sum_{l=1}^{L} \bar{\boldsymbol{z}}_{l, :} \cdot \boldsymbol{P}_{:, i} \\
    &=
    \sum_{k=1}^{K}
    \left(
        \frac{1}{L} \sum_{l=1}^{L} \boldsymbol{\Phi}_l(x)
    \right)
    \cdot
    \left(
        v^{\boldsymbol{\Omega}}(\boldsymbol{\zeta}) \cdot \boldsymbol{P}
    \right)_{k, i}. 
\end{split}
\end{equation*}
%


\subsection{Training Objectives for Interpretability}
\label{subsec:objectives}

Training of concept composition function $\boldsymbol{\Phi}$ proceeds under two distinct scenarios based on concept annotation availability: unsupervised and supervised.
In the supervised case, $\boldsymbol{\Phi}$ uses sample-specific ground-truth concept labels, typically represented as binary vectors indicating concept presence or absence.
In the unsupervised scenario, training relies on loss functions designed to induce properties that jointly optimize interpretability and predictive performance.



\paragraph{Explanation Loss with Concept Annotations.}
In the supervised scenario, the model leverages ground-truth concept annotations $\mathbf{Q}_{i} \in \{ 0, 1 \}^{L \times K}$, where each element indicates whether a specific concept is present in a given input patch.
To align the concept activation with these annotations, we introduce an explanation loss term, $\mathcal{L}_{\text{expl}}$.
This loss penalizes discrepancies between the concept activation $\boldsymbol{\Phi}(x_i)$ and the ground-truth annotations $\mathbf{Q}_{i}$, guiding to align the activation with human-interpretable annotations.

The concept activations $\boldsymbol{\Phi}$ from our proposed module are used as a regularization term by adding an \emph{explanation cost} to the objective function: $\mathcal{L}_{\text{expl}} = \lVert \boldsymbol{\Phi} - \mathbf{Q} \rVert^{2}_{F},$
where 
$\lVert \cdot \rVert$ is the Frobenius norm. 



\paragraph{Sparsity Loss based on Entropy.}
When concept annotations are absent, i.e., $\mathbf{Q}_{i} = \phi$, as is common in real-world cases, interpretability is achieved through the binding and refinement process of $\boldsymbol{\Theta}$. 
Thanks to the concept decomposition and refinement capability~(Eq.~(\ref{eq:concept_refinement})) of the module $\boldsymbol{\Theta}$, the process iteratively binds concepts to input features through competitive attention, refining concept representations at each step, even without explicit supervision. 

We enforce sparse concept activations by minimizing the entropy of the attention mask:
\begin{equation*}
    \mathcal{L}_{\text{sparse}}
    =
    H(\boldsymbol{\Phi}) = H(a_{1}, \dots, a_{|\boldsymbol{\Phi}|})
    =
    \frac{1}{|\boldsymbol{\Phi}|} \sum_{i} -a_{i} \cdot \log(a_{i})
\end{equation*}
where $a_{i}$ is the $i$-th concept composition value in $\boldsymbol{\Phi}$ and $|\boldsymbol{\Phi}|$ its cardinality.
This entropy-based sparsity loss applies in scenarios with and without ground-truth explanations.

\paragraph{Final Loss.}
Thus, the final loss in training the model becomes $\mathcal{L} = \mathcal{L}_{\text{cls}} + \lambda_{\text{expl}} \mathcal{L}_{\text{expl}} + \lambda_{\text{sparse}} \mathcal{L}_{\text{sparse}}$, where $\mathcal{L}_{\text{cls}}$ denotes the conventional classification loss, such as cross-entropy loss.
Notice that the constant $\lambda_{\text{expl}} \geq 0$ controls the relative contribution of the explanation loss to the total loss, and so our model can be applied with ground-truth explanations~($\lambda_{\text{expl}} > 0$) or without them~($\lambda_{\text{expl}} = 0$). 
Finally, the constant $\lambda_{\text{sparse}}$ scales the influence of sparsity loss. 
\section{Experiment}
\label{sec:exp}

%
\begin{table}[t!]
    \caption{Performance evaluation on CIFAR100, ImageNet, and CUB200. We used ViT-T for CIFAR100, ViT-L for CUB200, and ViT-S for ImageNet and Places365. For the experimental results of Bi-ICE on CUB200, the left is the result using concept annotations, whereas the right is the one without the annotations.}
    \centering
    \resizebox{0.9\linewidth}{!}{%
    \begin{NiceTabular}{lcccc}
    \toprule
        Model & CIFAR100 & ImageNet & CUB200 & Places365 \\
        \midrule
        ResNet50 & $83.5_{\pm 0.1}$ & $76.1_{\pm 0.2}$ & $79.4_{\pm 0.2}$ & $44.8_{\pm 0.1}$ \\
        (+ Bi-ICE) & $83.1_{\pm 0.1}$ & $75.9_{\pm 0.2}$ & $79.7/79.4_{\pm 0.2}$ & $45.2_{\pm 0.1}$ \\
        \midrule
        ViT & $84.9_{\pm 0.1}$ & $76.5_{\pm 0.2}$ & $87.7_{\pm 0.1}$ & $48.5_{\pm 0.2}$ \\
        (+ Bi-ICE) & $85.1_{\pm 0.1}$ & $77.4_{\pm 0.1}$ & $90.3/90.1_{\pm 0.1}$ & $50.5_{\pm 0.2}$ \\ 
    \bottomrule
    \end{NiceTabular}
    }
    \vspace{-3mm}
    \label{tab:perform_eval}
\end{table}
\begin{figure}
\centering 
\includegraphics[width=0.98\columnwidth]{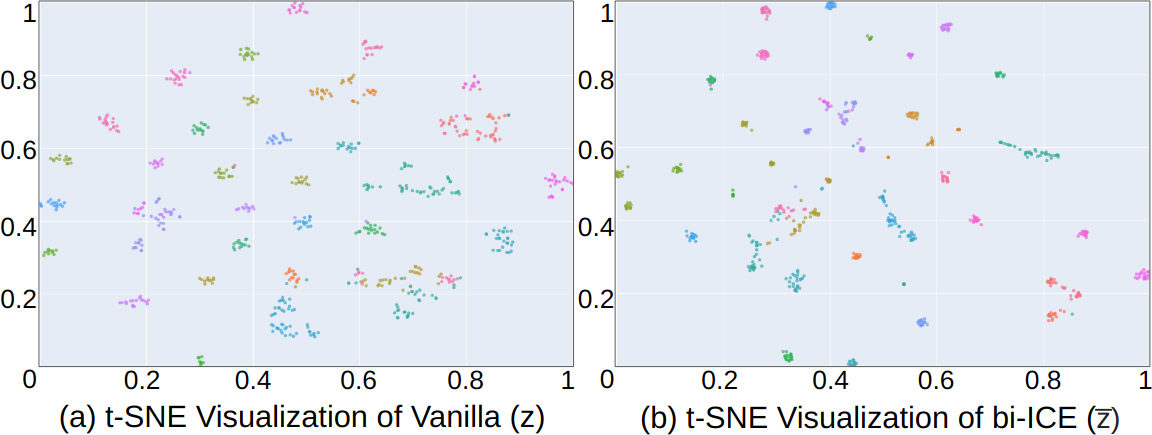}
\caption{t-SNEs of (a)~vanilla ViT~($\boldsymbol{z}$), V-score: $0.9124$ vs. (b)~Bi-ICE~($\bar{\boldsymbol{z}}$), V-score: $\boldsymbol{0.9366}$ on CUB w.o. concept annotations.}
\vspace{-3mm}
\label{fig:tsne_z_z-bar}
\end{figure}
\begin{figure*}[t!]
    \centering
    \includegraphics[width=0.93\textwidth]{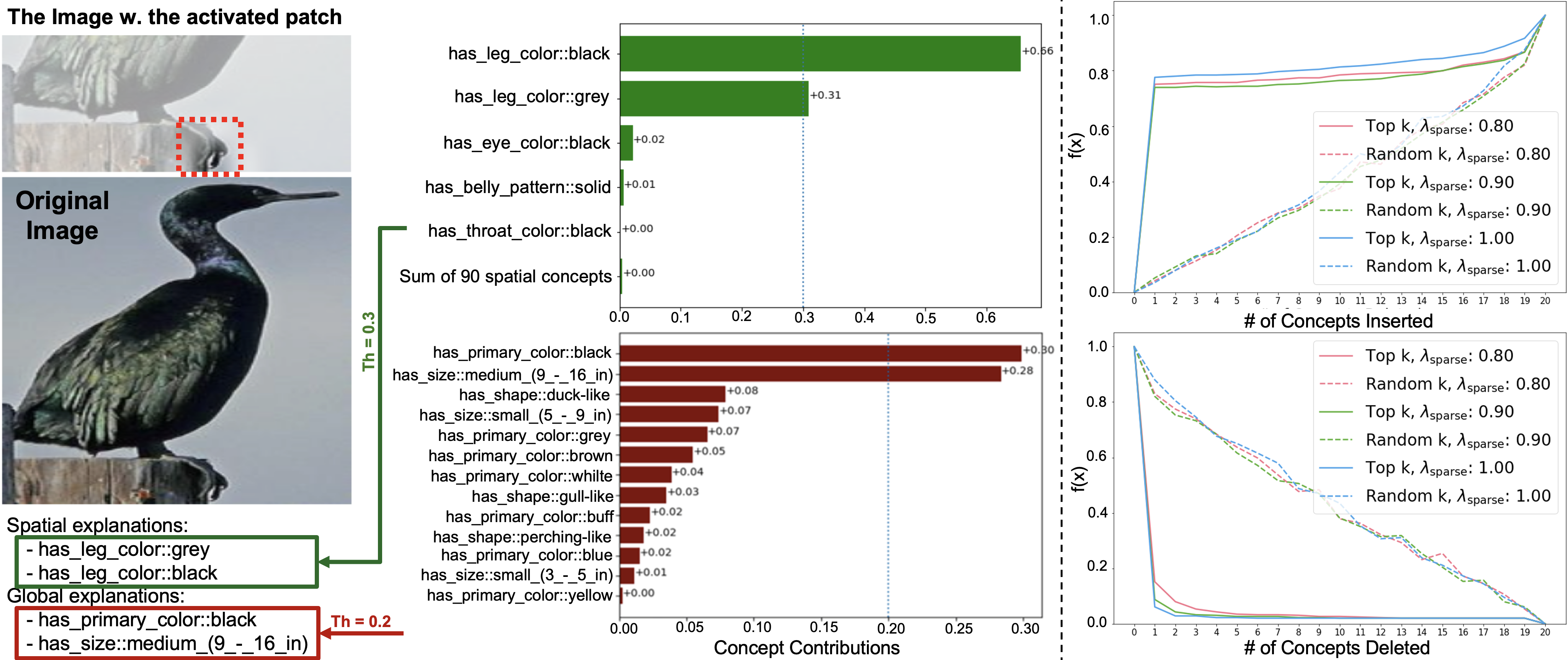}
    \caption{Computational-level Interpretability Analysis. \textbf{Left}: Concept Contributions of the class \textit{`Brandt Cormorant'}, showing their impact on the model's decision. The dashed red box indicates the activated patch ($>$ 0.6). Green for spatial, and red for global. \textbf{Right}: Concept-Insertion and Deletion graphs of randomly selected 50 classes in CUB. $f(x)$ is the normalized test accuracy with the range $[0, 1]$. }
    \vspace{-3mm}
    \label{fig:1_computational}
\end{figure*}
%

\subsection{Performance Evaluation}
\label{subsec:results}

We evaluate the performance of the models on four datasets: CIFAR100~\cite{krizhevsky2009learning}, CUB-200-2011~\cite{wah2011caltech}, ImageNet-1K~\cite{deng2009imagenet}, and Places365~\cite{zhou2017places}.
For baseline, we fine-tuned ResNet50~\cite{he2016deep} and Vision Transformers~(ViT-Tiny, ViT-Small, and ViT-Large)~\cite{dosovitskiy2020image}.
For Bi-ICE, we replaced the last classifier with our Bi-ICE and trained the model following the same training setups for the baseline.
We adjusted the number of concepts to match each dataset's characteristics.
Specifically, 
$50$ for CIFAR100, $150$ for ImageNet, and $200$ for Places365.
Following prior work~\cite{rigotti2021attention, hong2024concept}, we configured $13$ global concepts and $95$ spatial concepts for CUB using the concept annotations. In contrast, for CUB without annotations, we set $\lambda_{\text{exp}}$ to 0 so that the model is trained without the explanation loss.

Table~\ref{tab:perform_eval} presents a performance comparison of the models.
It indicates that integrating our module, Bi-ICE, maintains or even slightly improves the performance of the original backbone models, all while adding interpretability.
This demonstrates that our approach provides interpretability without compromising, and sometimes even improving, the model’s predictive accuracy.
Additionally, Fig.~\ref{fig:tsne_z_z-bar} shows why our approach performed better than vanilla ViT on CUB. 
We measured the V-score~\cite{rosenberg2007v}, representing homogeneity and completeness, with higher values indicating better clustering. 
The embeddings from the vanilla ViT backbone~($\boldsymbol{z}$) are disentangled based on classes, whereas $\bar{\boldsymbol{z}}$ processed by Bi-ICE are qualitatively and quantitatively better clustered and semantically structured, leading to enhanced classification performance.
Additional evaluations including ablation study and comparison with other baselines can be found in Appendix~\ref{appsubsec:performance}.


\subsection{Analysis on Inner Interpretability}
\label{subsec:analysis}

Here, we perform an in-depth analysis of inner interpretability using the task of classifying bird species based on their concepts, a common topic in explainable AI.
Our analysis primarily focuses on the CUB-200-2011 dataset, which includes ground-truth concept annotations tailored for fine-grained bird species classification.
Yet, it can also be extended to other datasets, such as ImageNet, where ground-truth concept annotations are unavailable~(Appendix.~\ref{appsubsec:analysis}).

The CUB dataset involves $312$ concepts distributed unevenly across images, and so we utilize a pre-processing method from \citet{rigotti2021attention} and \citet{hong2024concept}. 
The concepts are then categorized into: 
(i)~\emph{global} annotations, which indicate the presence of a particular concept within the entire input~(e.g., size of a bird), and (ii)~\emph{spatial} annotations, which denote the presence of specific concepts in various image patches of the input~(e.g., eye color of a bird).
Let $\boldsymbol{Q}_{i} \in \{0, 1\}^{L \times K}$ represent the concept annotations for a given $x_i$, where $L$ is the number of image patches and $K$ is the number of concepts, consisting of $K^\text{global}$ and $K^\text{spatial}$ (i.e. $K = K^{\text{global}} + K^{\text{spatial}}$).
Then, $\boldsymbol{Q}_{i} = \{ \boldsymbol{Q}^{\text{global}}_{i}, \boldsymbol{Q}^{\text{spatial}}_{i} \}$, where $\boldsymbol{Q}^{\text{global}}_{i} \in \{0, 1\}^{K^{\text{global}}}$ and $\boldsymbol{Q}^{\text{spatial}}_{i} \in \{ 0, 1 \}^{L \times K^\text{spatial}}$.


%
\begin{figure*}[t!]
    \centering
    \includegraphics[width=0.98\textwidth]{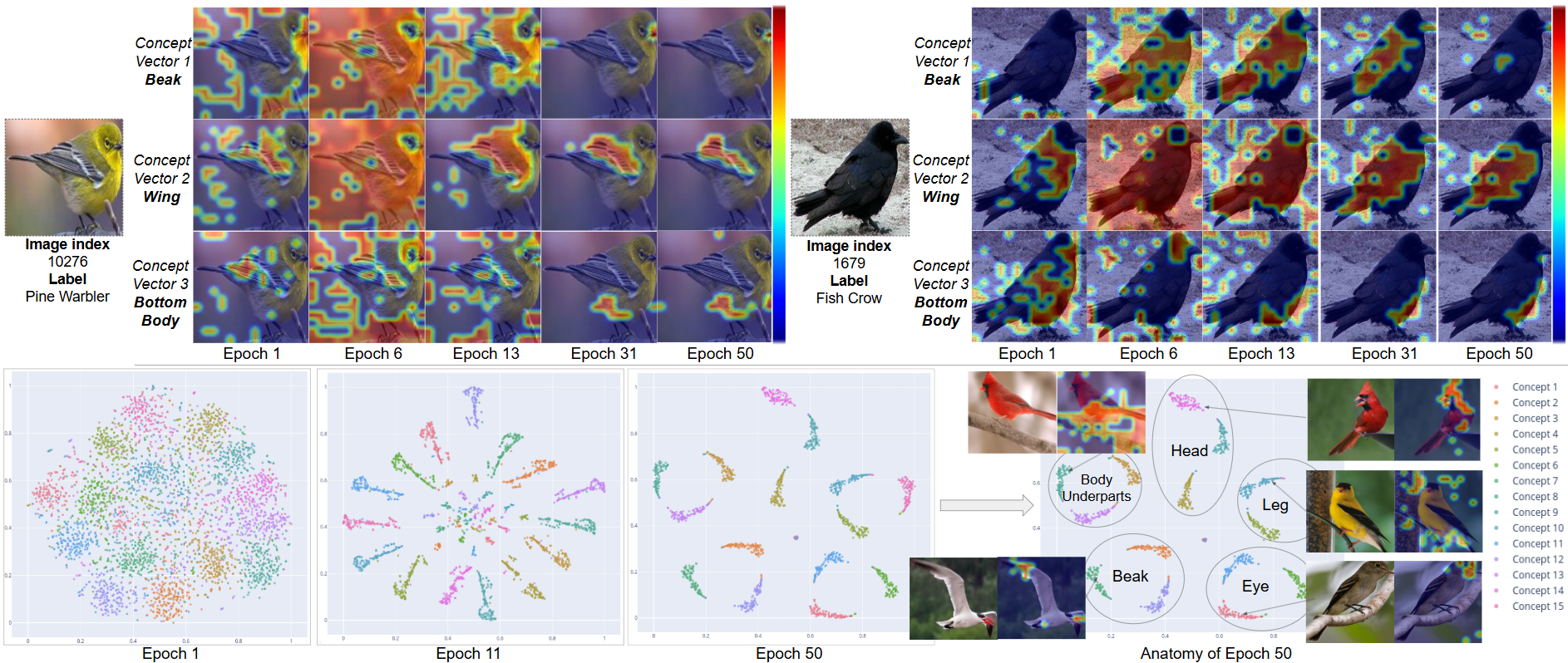}
    \caption{Algorithmic-level Interpretability Analysis for Concept Vector Convergence Across Training Epochs.
    \textbf{Top}: Attention map evolution for concept vectors.
    Each row represents a concept vector highlighting semantic parts like ``Beak," ``Wing," and ``Bottom Body," with warmer colors indicating stronger focus. The attention maps become more refined over epochs.
    \textbf{Bottom}: t-SNE plots of concept vector distributions over epochs, showing semantic differentiation over time.
    By epoch 50, similar semantic concepts have converged into tight, well-separated clusters, each associated with a specific part of the image, such as ``Beak," ``Leg," or ``Body Underparts".} 
    \label{fig:2_algorithmic}
    \vspace{-3mm}
\end{figure*}

\subsubsection{Computational Level}
\label{subsubsec:comp}
At the computational level, we investigate the functional role of concept representations in the model's classification decision, addressing \textbf{`why'} a model makes a particular prediction.
This involves analyzing the contribution of concepts, including spatial and global concepts to classification, and evaluating whether the identified concepts align with actual prediction.
Furthermore, we can explain a counterfactual intervention case (Appendix~\ref{appsubsec:usecase1}).

\paragraph{Concept Importance Score.}
With $\boldsymbol{\Phi}={\boldsymbol{\Phi^{\text{global}}}+\boldsymbol{\Phi^{\text{spatial}}}}$ (Eq.~(\ref{eq:concept_activation}), we can easily visualize the concept contributions in each image.
We obtain $\boldsymbol{\Phi}^{\text{global}}$ by averaging the axis of image patches: $\boldsymbol{\Phi}^{\text{global}} \triangleq \operatorname{average}_{L} (\boldsymbol{\Phi}_{:, :K^{\text{global}}})  \in \mathbb{R}^{K^{\text{global}}}$, and $\boldsymbol{\Phi^{\text{spatial}}}$ by slicing the column of $\boldsymbol{\Phi}$: $\boldsymbol{\Phi}^{\text{spatial}} \triangleq \boldsymbol{\Phi}_{K^{\text{global}}:} \in \mathbb{R}^{L \times K^{\text{spatial}}}$.
For simplicity, we compute the global and spatial scores in separate, parallel ways.

The left side of Fig.~\ref{fig:1_computational} displays concept contributions for predicting the class `Brandt Cormorant' alongside its activated patch.
We consider a patch to be activated if its spatial concept activation exceeds 0.6 (dashed red box).
As indicated in the green chart, the concept {\small\texttt{has\_leg\_color::black}} holds the highest contribution score at $0.66$, followed by {\small\texttt{has\_leg\_color::grey}} at $0.31$, which is an interesting observation, as the patch in the original image indeed contains grey in the leg area.
Additionally, among global concepts, {\small\texttt{has\_primary\_color::black}} contributes most significantly with a score of $0.3$, closely followed by the `medium size' concept with a score of $0.28$.



\paragraph{C-deletion \& C-insertion.}
To further validate the estimated concept contribution scores, we employed the Concept insertion and deletion evaluations~\citep{fel2024holistic}, which serve as key metrics for assessing the sparsity and faithfulness of concept representations in our model. 
We assess model performance changes when concepts are sequentially added or removed based on their importance scores---significant performance shifts (increases upon insertion and decreases upon deletion) indicate that the model effectively identifies and utilizes key concepts in its predictions.

For this experiment, we randomly selected 50 classes and trained our method without using any concept annotations, setting the number of concepts~($K$) to~20 for analytical simplicity.
Additionally, we varied the sparsity regularization parameter $\lambda_{\text{sparse}}$ with values 0.8, 0.9, and 1.0 to observe its impact on the model's behavior.

As shown in Fig.~\ref{fig:1_computational}, our method (solid line) shows a steeper increase in performance than a random insertion order (dashed line), indicating more accurate importance estimation.
Specifically, our model with higher sparsity regularization (larger $\lambda_\text{sparse}$) shows steeper performance gains, indicating that enforcing sparsity ensures the model to rely on a minimal yet essential set of concepts.
Similarly, in the C-deletion experiment, our method shows a significant performance drop compared to random removal, underscoring the accuracy of our importance estimations.
This steep drop further validates the faithfulness of our method, as it confirms that the identified concepts are indeed critical for the model's predictions.
Again, our model with higher sparsity regularization exhibits a more significant drop.

\subsubsection{Algorithmic Level}
\label{subsubsec:alg}
At the algorithmic level, interpretability outlines the human-understandable steps that enable the model to perform its tasks, revealing the intermediate reasoning processes that connect the input data to the final output, providing insights into \textbf{`how'} of the model’s interpretability.

As shown in Fig.~\ref{fig:2_algorithmic}, we conducted two experiments to examine how concept vectors evolve throughout training epochs to capture distinct, semantically meaningful regions without ground-truth concept annotations. 
In the first experiment~(Top, Fig.~\ref{fig:2_algorithmic}), we tracked the progression of 15 spatial concept vectors by highlighting their concept composition score $\boldsymbol{\Phi}$ using a colorbar. 
The resulting heatmaps display how each concept vector gradually refines its focus over time for two bird images, with specific vectors focusing on features such as the beak, wing, and lower body.
The second experiment (Bottom, Fig.~\ref{fig:2_algorithmic}) examines the distribution of concept vectors in the embedding.
Using a subset of 50 classes with 15 spatial concept vectors, we visualized the convergence of concept vectors over epochs with t-SNE~\cite{van2008visualizing}.
At the beginning of training~(Epoch 1), the vectors form a densely entangled cluster, reflecting undifferentiated representations.
However, by epoch 50, they form well-defined clusters corresponding to distinct semantic features.
Therefore, these results demonstrate that our model autonomously learns interpretable concept representations, even without explicit concept explanations.
With this analysis, we can provide empirical guidance for determining the optimal number of concepts (Appendix~\ref{appsubsec:usecase2}).


\subsubsection{Implementation Level}
\label{subsubsec:impl}

Implementation-level of interpretability is achieved by decomposing the network at an appropriate level of abstraction to investigate the underlying mechanisms.
This involves identifying specific \textbf{components} within the model that exerts a causal influence on the output, thereby narrowing down the set of basic operations for model behavior. 

\begin{figure}
    \centering
    \includegraphics[width=0.98\columnwidth]{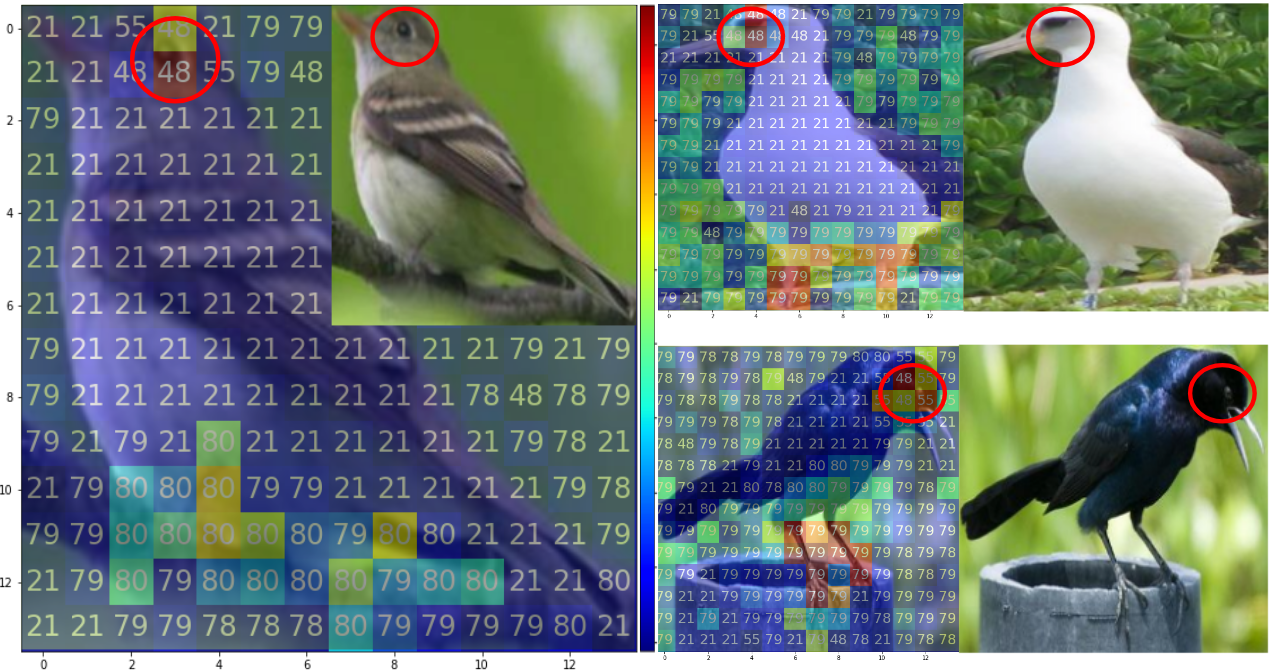}
    \caption{Implementation-level of Interpretability for Concept Localization. The image shows three birds with a $14 \times 14$ grid overlay. Each cell is color-coded with a colorbar, where red indicates a high concept contribution score, and blue indicates its low value. 
    }
    \label{fig:3_implementation}
    \vspace{-5mm}
\end{figure}

Fig.~\ref{fig:3_implementation} illustrates the concept localization of Bi-ICE with ground-truth concept annotations, highlighting key regions where specific attributes are detected.
%
Concept 48, {\small\texttt{has\_eye\_color::black}}, consistently aligns with eye regions, showing accurate identification of black eyes, 
while Concepts 79~({\small\texttt{has\_leg\_color::grey}}) and 80~({\small\texttt{has\_leg\_color::black}}) align with leg regions. 
Concept 21, {\small\texttt{has\_underparts\_color::white}}, appears in deep blue across the grid, indicating minimal activation in these areas.
This mapping offers a clear view of the concepts that the model uses in its decision-making process.
A user study, detailed in Appendix~\ref{app:user_study}, further confirmed the human-interpretability of these concepts.


\section{Related Work}
\label{sec:related_work}


\paragraph{Existing XAI Methods in Image Models} 

The existing XAI approaches reveal significant gaps in their analytical depth compared to the inner interpretability framework with multilevel analysis.
For example, post-hoc approaches, such as feature attribution~\cite{sundararajan2017axiomatic, selvaraju2017grad, han2024respect, bach2015lrp} and concept attribution~\cite{kim2018interpretability, ghorbani2019towards, fel2023craft, kim2024decomposemodel}, operate entirely from the outside.
They approximate input–output associations by either highlighting important features or aligning predictions with human concepts, but remain blind to the internal mechanisms.

In contrast, intrinsically interpretable methods, including CBM~\cite{koh2020concept, yang2023language, yuksekgonul2022post, semenov2024sparse} and prototype~\cite{snell2017prototypical, Nauta_2023_CVPR, DBLP:journals/corr/abs-2112-02902, DBLP:journals/corr/abs-1806-10574, DBLP:journals/corr/abs-2111-15000, ma2024looks} based models build interpretability directly into the model architecture itself by forcing predictions through an explicit concept layer or providing interpretable exemplars.
While these approaches provide transparency by exposing human-aligned intermediate variables, their scope is largely confined to the implementation level. 
They reveal what is represented, but not how it is algorithmically constructed or computationally motivated, leaving the causal and structural dynamics of the model unexplained.

Inner interpretability, by contrast, aims to uncover a model's internal mechanisms through a principled, multilevel analysis~\cite{rauker2023toward, vilasposition}.
Rather than stopping at external associations or implementation-level, it seeks to connect the computational, algorithmic, and implementation levels to explain why and how a model arrives at its decisions.
Our framework diverges from both post-hoc and intrinsic paradigms by adopting inner interpretability perspective.
To our knowledge, ours is the first to apply a multilevel analytical interpretability framework to image classification tasks, analyzing model behavior across all levels.




\paragraph{Deep Learning with Shared Global Workspace.}
Neural networks incorporating memory mechanisms have been widely explored to enhance the ability to store and process information, enhancing task performance.
Much of this research draws inspiration from cognitive neuroscience, particularly from theories of human cognition such as Global Workspace Theory~(GWT)~\cite{baars1993cognitive, dehaene1998neuronal, mashour2020conscious}, 
which proposes a centralized framework for sharing information among specialized processing units.
Building on the principles of GWT, the Shared Global Workspace~(SGW)~\cite{goyal2021coordination} has emerged as an interpretation of GWT within deep learning.

Recent approaches inspired by GWT and SGW have been developed~\cite{dehaene1998neuronal, goyal2019recurrent, jaegle2021perceiver, munkhdalai2019metalearned, santoro2018relational}, including a fine-tuning method that integrates a Mixture-of-Experts to address intrinsic challenges related to reducing token uncertainty~\cite{wu2024gw}.
Other advances include dual-memory transformer architectures designed for multi-modal question-answering and reasoning~\cite{zeng2024sdmtr}, an interpretable framework with intrinsic explanations~\cite{hong2024concept}, and a debiasing approach that learns unbiased and interpretable representations of attributes without requiring predefined bias types~\cite{hong2024debiasing}.

In this paper, we leverage the SGW to implement the working memory module containing concept vectors and the learning process with bi-directional interactions between the input and the memory module. 
\section{Conclusion}
\label{sec:conclusion}
We presented an inner interpretability framework with multilevel analysis for image classification, utilizing a Bi-directional Interaction between Concept and Input Embeddings~(Bi-ICE) module. 
It incorporates a working memory module for concept learning, featuring an intuitive training process that facilitates bi-directional interactions between concepts and input embeddings. 
With our multilevel analysis, we demonstrated that Bi-ICE improves transparency by generating predictions based on human-understandable concepts, quantifying their contributions, and localizing them within the inputs.
Additionally, it also exhibits algorithmic and developmental interpretability by illustrating the process of concept learning and convergence.

We plan to extend the proposed method in various ways for future work.
Aligned with recent work on understanding learning dynamics~\citep{jacot2018neural, lee2019wide, simon2023stepwise}, we aim to develop theoretical fundamentals to support the developmental interpretability of concept learning.
Additionally, we will enhance our method by mapping concept features onto the latent space of a pretrained generative model, thus improving the visualization of concept activations and intervention analysis through the image generated with concepts. 
\section*{Acknowledgment}
The ASU researchers were supported by NSF award 2223839, while the SNU researchers were supported by the Korean Government through grants from NRF (2021R1A2C3006659) and IITP (RS-2021-II211343, RS-2022-II220953, RS-2025-25442338).

{
    \small
    \bibliographystyle{ieeenat_fullname}
    \bibliography{main}
}

\clearpage
\appendix
\section*{Supplementary Material}
\label{sec:appendix}
\renewcommand\thefigure{A-\arabic{figure}}\setcounter{figure}{0}
\renewcommand\thetable{A-\arabic{table}}
\setcounter{table}{0}
\setcounter{section}{0}
\pagenumbering{Alph}

\section{Reproducibility}
\label{appsec:reproduce}
All the source code, figures, and models will be available at~\url{https://github.com/jyhong0304/bi-ICE}.







\section{Comparison to Other Intrinsically Interpretable Methods}
We emphasize the novelties of our Bi-ICE framework compared to Concept Bottleneck Models~(CBMs) and prototype-based approaches. In general, both CBMs and prototype-based models are restricted to providing interpretability only at the implementation level.

\subsection{Limitations of CBMs and Prototype-based Approaches}

CBMs suffer from a strong dependence on annotations, as they require either ground-truth concept labels or highly accurate concept predictors. They also face scalability challenges, often showing degraded performance when applied to large-scale datasets, such as ImageNet or Places365. In contrast, Bi-ICE achieves strong performance across both small- and large-scale settings, and operates effectively with or without concept supervision.
From the standpoint of concept computation, Bi-ICE adopts a more general formulation than CBMs. Whereas concepts in CBMs are often assumed to be independently predictive in a relatively simple way by representing each concept as a scalar value~\cite{xu2024energy}, our framework models concepts as high-dimensional vector embeddings. This richer representation allows the model to (i) capture semantic meaning in a distributed and relational manner through patch-wise alignments, (ii) preserve spatial grounding of concepts~(e.g., the concept \emph{beak} aligning with head patches), and (iii) support bi-directional interactions, where image patches refine concepts and concepts in turn broadcast back to refine image embeddings.

Prototype-based approaches face significant scalability challenges. Since each prototype corresponds to a stored exemplar, the number of prototypes must increase as dataset size and class diversity grow. This inevitably leads to \emph{prototype explosion}, where hundreds or even thousands of prototypes are required for large-scale datasets, such as ImageNet. In practice, these prototype-based approaches are rarely applied with backbones larger than ViT-S, as the computational burden becomes prohibitive and the risk of overfitting to individual prototypes increases with larger-capacity models. As a result, these models incur substantial memory and computational costs and lose interpretability due to redundant or overlapping prototypes.
In contrast, Bi-ICE leverages \emph{shared concept embeddings} that are reused across all training examples. Each concept is represented as a high-dimensional vector that is refined during training. This design offers: i) efficiency--far fewer parameters compared to storing per-class prototypes, ii) generalization--concepts capture semantic features~(e.g., \emph{beak, wing}) that transfer across the dataset rather than being tied to specific instances, and iii) robust scalability--Bi-ICE scales effectively to datasets like ImageNet and Places365 with small number of concepts while maintaining accuracy, where prototype-based approaches become impractical.
Moreover, prototype-based methods typically provide only qualitative faithfulness validation, relying on nearest-example visual matches. They lack systematic causal testing to ensure that prototypes are truly necessary for predictions. Bi-ICE overcomes this limitation by: i) performing concept insertion and deletion to verify causal necessity, ii) applying counterfactual interventions~(e.g., recoloring a bird’s body) to induce systematic and interpretable shifts in concept activations and predictions and
iii) offering patch-level grounding, where concepts localize to specific image regions, ensuring that explanations align with the actual input evidence.

\subsection{Additional Performance Comparison}
\begin{table}[t!]
    \caption{Performance comparison between various CBMs and ours on CIFAR100, ImageNet, CUB-200-2011, and Places365. As a backbone for our Bi-ICE, we used ViT-T~(5.7M) for CIFAR100, ViT-L~(307M) for CUB, and ViT-S~(21.7M) for ImageNet and Places365. The number of parameters of our Bi-ICE are as follows: 6M for CIFAR100, 23.9M for ImageNet, 329M for CUB200, and. 23.8M for Places365. From~\cite{semenov2024sparse}, all CBMs used CLIP-ViT-L/14~(428M) and B/32~(88.3M) as backbones, which is highly larger than our models' configurations. For the CBMs, the results are sourced from~\cite{semenov2024sparse}. For Bi-ICE, the result is from our evaluation.}
    \centering
    \small
    \resizebox{0.98\linewidth}{!}{%
    \begin{NiceTabular}{l|c|c|c|c}
    \toprule
        Model &  \textbf{CIFAR} & \textbf{ImageNet} & \textbf{CUB} & \textbf{Places} \\
        \midrule\midrule
        \text{LaBo}~\cite{yang2023language} & $69.1$ & $70.4$ & $71.8$ & $39.4$ \\
        \text{PCBM}~\cite{yuksekgonul2022post} & $57.2$ & $62.6$ & $63.9$ & $39.7$ \\
        \text{Lf-CBM}~\cite{oikarinen2023label} & $65.3$ & $72.0$ & $74.5$ & $43.7$ \\
        Sparse CBM~\cite{semenov2024sparse} & 74.9 & 71.6 & 80.0 & 41.3 \\
        \midrule
        Bi-ICE~(Ours) & $\textbf{85.1}_{\pm 0.1}$ & $\textbf{76.4}_{\pm 0.1}$ & $\textbf{90.3}_{\pm 0.1}$ & $\textbf{50.5}_{\pm 0.2}$ \\ 
    \bottomrule
    \end{NiceTabular}
    }
    \label{tab:perform_eval_cbm}
\end{table}
\begin{table}[t!]
    \caption{Performance comparison between prototype-based approaches and ours on CUB-200-2011. 
    For the prototype-based approaches, we report their best results from the original papers. 
    The backbones and parameter sizes are as follows: 
    PIP-Net uses ConvNeXt-Tiny~(28.6M), ProtoPFormer uses DeiT-S~(22M), ProtoPool uses ResNet-50~(25.6M), 
    ProtoPNet uses an ensemble of VGG19+ResNet34+DenseNet121~(173.5M), Deform-ProtoPNet uses DenseNet-161~(28.7M), 
    and ProtoConcept uses DenseNet-161~(28.7M). 
    Note that in prototype-based models, the number of prototypes is heuristically fixed for each class and cannot be shared across classes. This design often results in redundant concepts~(e.g., separate \emph{head} prototypes for different bird species) and limits scalability to larger backbones such as ViT-L. In contrast, Bi-ICE allows concept vectors to be shared across classes, leading to more compact and efficient representations. Even with a lightweight backbone~(ViT-S, 22M), Bi-ICE achieves competitive performance while offering better scalability and efficiency.}    
    \centering
    \begin{NiceTabular}{l|c}
    \toprule
        Model & \textbf{CUB-200-2011} \\
        \midrule\midrule
        ViT-S : 22M & $81.3_{\pm 0.4}$ \\
        \midrule
        PIP-Net~\cite{Nauta_2023_CVPR} & $84.3$ \\
        ProtoPFormer~\cite{xue2022protopformer} & $84.9$ \\
        ProtoPool~\cite{DBLP:journals/corr/abs-2112-02902} & $87.6$ \\
        ProtoPNet~\cite{DBLP:journals/corr/abs-1806-10574} & $84.8$ \\
        Deform-ProtoPNet~\cite{DBLP:journals/corr/abs-2111-15000} & $86.5$ \\
        ProtoConcept~\cite{ma2024looks} & $85.2$ \\
        \midrule
        Bi-ICE~(Ours): 25.4M & $85.3_{\pm 0.2}$ \\ 
    \bottomrule
    \end{NiceTabular}
    \label{tab:perform_eval_proto}
\end{table}

While we have outlined the distinctions between our Bi-ICE and CBM/prototype-based methods above, we also recognize that these architectures are widely considered important milestones in image-based XAI for their explicit emphasis on interpretability.
Accordingly, we include some representative CBMs and prototype-based models as baselines and provide a detailed comparison of their performance against our proposed approach.

Table~\ref{tab:perform_eval_cbm} compares the performance of our model, Bi-ICE, with various CBM architectures across four datasets: CIFAR100, ImageNet, CUB-200-2011, and Places365. 
Bi-ICE achieves $85.1\%$ on CIFAR100, $76.4\%$ on ImageNet, $90.3\%$ on CUB-200-2011, and $50.5\%$ on Places365, surpassing the strongest CBM competitor by clear margins on every dataset. Importantly, Bi-ICE obtains these results without relying on CLIP-like vision–language pretraining, which other CBMs used to compensate for the absence of ground-truth concept annotations. This demonstrates that Bi-ICE delivers state-of-the-art accuracy while maintaining interpretability, even under settings where other CBMs depend on external pretrained supervision.

For comparison with prototype-based approaches, we chose CUB-200-2011 only because for larger datasets such as ImageNet and Places365, the computational and memory requirements of prototype-based models grow significantly with the number of classes and prototypes, making them impractical for efficient evaluation.

Table~\ref{tab:perform_eval_proto} shows the performance comparison with the prototype-based methods of CUB-200-2011.
To ensure a fairer comparison with them, we configured Bi-ICE in an unsupervised concept learning setup with $50$ number of concepts and a ViT-S backbone~(22M parameters), closer in scale to ResNet- or DeiT-based backbones commonly used for the baseline. Even under this lightweight setup, Bi-ICE delivers $85.3\%$ accuracy on CUB-200-2011. 
This performance is notable given that prototype-based methods often depend on large and redundant ensembles~(e.g., VGG19+ResNet34+DenseNet121) or heavy backbones~(e.g., DenseNet-161).
Moreover, while prototype-based architectures offer strong interpretability by grounding predictions in visual exemplars, they usually allocate a fixed set of prototypes per class, which limits their scalability to larger or more complex settings.
For example, in the case of CUB-200-2011, most prototype-based methods allocate 10 prototypes per class, resulting in a total of 2,000 prototypes across the 200 classes. This design not only inflates the number of parameters but also produces redundant concepts~(e.g., separate all different body parts prototypes for different bird species), thereby limiting scalability. In contrast, Our Bi-ICE adopts a concept-sharing mechanism, where a compact set of concept vectors~(e.g., 50 concept vectors only) is shared across all classes. This yields far greater efficiency while preserving interpretability. Importantly, if prototype-based approaches were to increase the number of prototypes per class to match our concept capacity~(e.g., 50 prototypes per class), the resulting computational and memory costs would become prohibitive---especially when scaling to larger backbones or more complex datasets. Bi-ICE thus achieves competitive accuracy with significantly fewer resources, highlighting its scalability advantage over prototype-based models. 
\section{Experimental Details}
\label{appsec:exp}


\subsection{Dataset Statistics}
\label{appsubsec:dataset_statistics}
\begin{table*}[t!]
\caption{Benchmark Dataset Statistics. $\dagger$ indicates the rescaled size of inputs for the ViT backbone, which is different from the original sizes of the datasets. $\ddagger$ indicates the number of concepts corresponding to latent spatial concepts.}
\centering
\begin{NiceTabular}{m{0.18\textwidth}|m{0.13\textwidth}|m{0.2\textwidth}|m{0.15\textwidth}|m{0.15\textwidth}}
\textbf{Dataset} & \textbf{CIFAR100} & \textbf{CUB-200-2011} & \textbf{ImageNet} & \textbf{Places365} \\
\midrule\midrule
Input size & $3 \times 28 \times 28$ & $3 \times 224 \times 224^{\dagger}$ & $3 \times 224 \times 224^{\dagger}$ & $3 \times 224 \times 224^{\dagger}$ \\ 
\midrule
\# Classes & 100~(class) & 200~(bird species) & 1000~(objects) & 365~(scenes) \\ 
\midrule
\# Concepts & $20^{\ddagger}$ & 13~(global), 95~(spatial) & $150^{\ddagger}$ & $200^{\ddagger}$ \\
\midrule
\# Training samples & $55,000$ & $5,994$ & $255,224$ & $1,803,460$ \\ 
\midrule
\# Validation samples & $5,000$ & $1,000$ & $10,000$ & $30,000$ \\
\midrule
\# Test samples & $10,000$ & $4,794$ & $20,000$ & $36,500$ \\
\bottomrule
\end{NiceTabular}
\label{tab:dataset_stat}
\end{table*}

Tab.~\ref{tab:dataset_stat} summarizes the statistics of all benchmark datasets used in our experiments.
Because none of the datasets includes a predefined validation split, we manually selected a portion of the training data to serve as a validation set for hyperparameter tuning.

The CIFAR-100 dataset contains 100 fine-grained image classes, but there are no ground-truth concept annotations that we can leverage. 
Thus, we arbitrarily set the number of latent spatial concepts to $20$ to train our model.

For the CUB-200-2011, we followed the pre-processing steps outlined in~\cite{rigotti2021attention}.
Initially, the original dataset has $312$ binary concepts, but we filtered them to retain only those that occur in at least 45$\%$ of all samples within a given class and occur in at least $8$ classes.
Thus, we got a total of 108 concepts, which we grouped into two types of concepts: spatial and global concepts.
Specifically, we identified $13$ global and $95$ spatial concepts by looking at each concept.
For example, {\small\texttt{has\_shape::perching-like}} and {\small\texttt{has\_primary\_color::black}} are global concepts, and {\small\texttt{has\_eye\_color::black}} and {\small\texttt{has\_leg\_color::grey}} are spatial concepts.

The ImageNet and the Places365~\cite{zhou2017places} datasets do not contain the ground-truth concept annotations the same as CIFAR100. 
Therefore, we arbitrarily set the number of latent concepts to $150$ for ImageNet and $200$ for Places365, respectively.

\subsection{Model Architectures}
\label{appsubsec:model_arch}
We provide additional details of the ResNet50~\cite{he2016deep} and Vision Transformers~(ViT)~\cite{dosovitskiy2020image} used as backbones.
We employ \href{https://huggingface.co/docs/timm/index}{\texttt{timm}} Python library supported by Hugging Face\textsuperscript{\texttrademark}.
In our experiments, we leverage \texttt{ResNet50}~({\small\texttt{resnet50.tv\_in1k}}), and three variants of \texttt{ViT} 
~({\small\texttt{vit\_large\_patch16\_224}}, {\small\texttt{vit\_small\_patch16\_224}}, {\small\texttt{vit\_tiny\_patch16\_224}} in \href{https://huggingface.co/docs/timm/index}{\texttt{timm}}). These variants are defined by their number of encoder blocks, the number of attention heads on each block, and the dimension of the hidden layer.
Specifically, ViT-L has $24$ encoder blocks with $16$ heads, and the dimension of the hidden layer is $1024$.
ViT-S has $12$ encoder blocks with $6$ heads, and the dimension of the hidden layer is $384$.
Finally, ViT-T has $12$ encoder blocks with $3$ heads, and the dimension of the hidden layer is $192$, which is much more lightweight. 

\begin{figure}[!t]
    \centering
    \includegraphics[width=0.98\columnwidth]{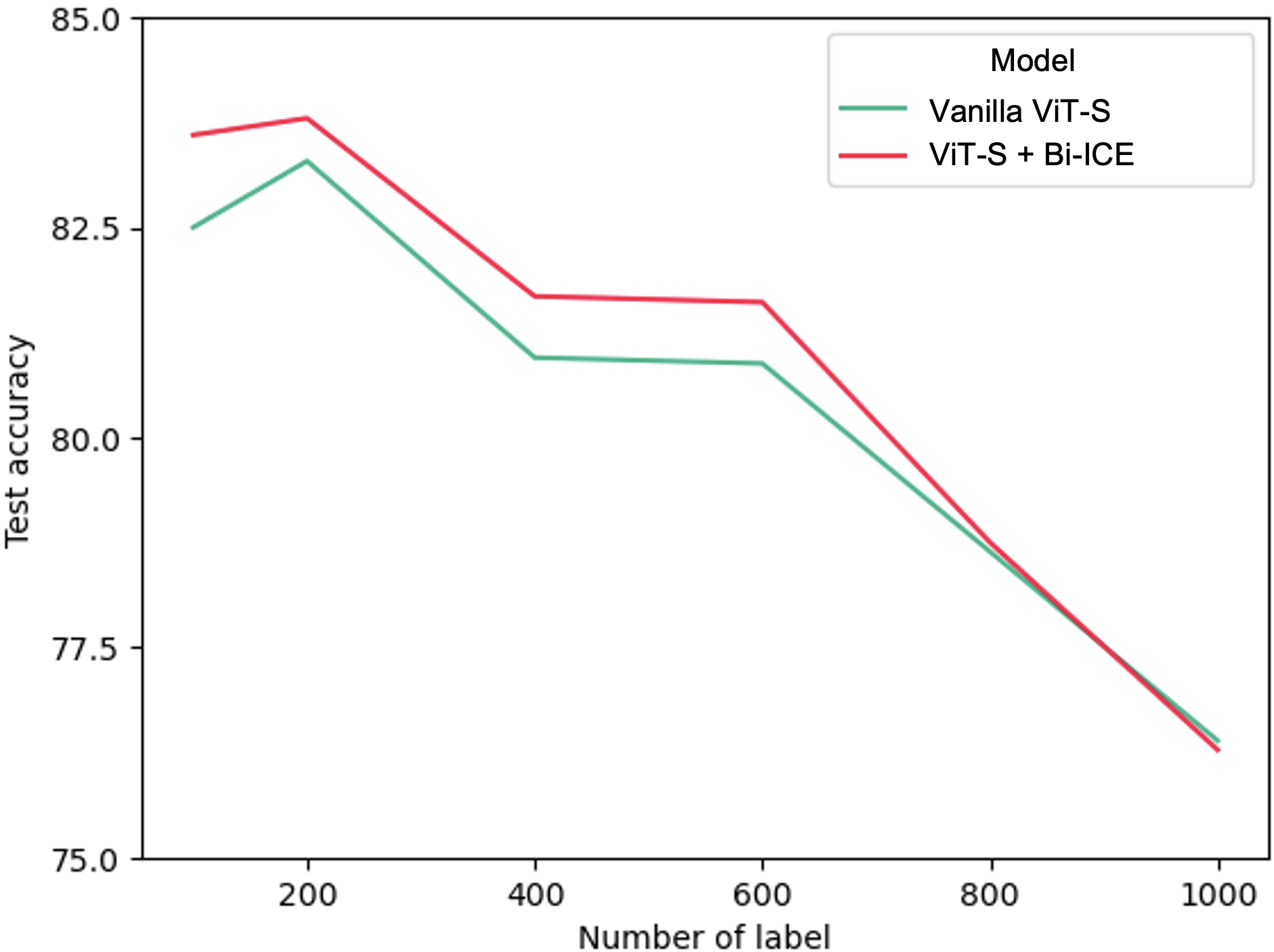}
    \caption{Performance comparison on ImageNet with different numbers of labels.}
    \label{fig:ablation_imagenet_num_label}
    \vspace{-3mm}
\end{figure}
%
%
\begin{table}[t!]
\caption{Ablation study of the number of concepts on ImageNet.}
\centering
\resizebox{0.98\linewidth}{!}{%
\begin{NiceTabular}{l||c|c|c|c}
\textbf{\# of concept vectors} & 50 & 100 & 150 & 200 \\
\midrule
\textbf{Test Acc.}~(\%) & $76.28_{\pm 0.2}$ & $76.31_{\pm 0.2}$ & $76.43_{\pm 0.1}$ & $76.35_{\pm 0.1}$ \\ 
\bottomrule
\end{NiceTabular}
}
\vspace{-3mm}
\label{tab:ablation_num_concepts_imagenet}
\end{table}

\paragraph{Training Setups in the Main Text.}
\begin{table}[t!]
    \caption{Hyperparameter setups for Vanilla ResNet50, Vanilla ViT-T, and all configurations of Bi-ICE on CIFAR100 in the main text.  }
    \centering
    \begin{NiceTabular}{c|c}
        Name & Value \\
        \midrule
        Batch size & $64$ \\
        Epochs & $20$ \\
        Warmup Iters. & $10$ \\
        Learning rate & $1e-4$ \\ 
        Explanation lambda $\lambda_{\text{expl}}$ & $1.0$ \\
        Weight decay & $1e-3$ \\ 
        Sparsity lambda $\lambda_{\text{sparse}}$& $0.5$ \\
        \bottomrule
    \end{NiceTabular}
    
    \label{tab:hyperparams_main_cifar100}
\end{table}
\begin{table}[t!]
    \caption{Hyperparameter setups for Vanilla ResNet50, Vanilla ViT-L, and all configurations of Bi-ICE on CUB-200-2011 in the main text. The learning rate for vanilla ResNet50 and ResNet50 + Bi-ICE is $1e-4$, and the learning rate for vanilla ViT-L and ViT-L + Bi-ICE is $1e-5$.}
    \centering
    \begin{NiceTabular}{c|c}
        Name & Value \\
        \midrule
        Batch size & $32$ \\
        Epochs & $50$ \\
        Warmup Iters. & $10$ \\
        Explanation lambda $\lambda_{\text{expl}}$ & $1.0$ \\
        Weight decay & $1e-3$ \\ 
        Sparsity lambda $\lambda_{\text{sparse}}$& $0.5$ \\
        \bottomrule
    \end{NiceTabular}
    \label{tab:hyperparams_main_cub}
\end{table}
\begin{table}[t!]
    \caption{Hyperparameter setups for Vanilla ResNet50, Vanilla ViT-S, and all configurations of Bi-ICE on ImageNet in the main text. The learning rate for vanilla ResNet50 and ResNet50 + Bi-ICE is $5e-5$. The learning rate for vanilla ViT-S is $1e-5$ whereas the one for ViT-S + Bi-ICE is $1e-4$.}
    \centering
    \begin{NiceTabular}{c|c}
        Name & Value \\
        \midrule
        Batch size & $64$ \\
        Epochs & $10$ \\
        Warmup Iters. & $10$ \\
        Explanation lambda $\lambda_{\text{expl}}$ & $0.$ \\
        Weight decay & $1e-3$ \\ 
        Sparsity lambda $\lambda_{\text{sparse}}$& $10.$ \\
        \bottomrule
    \end{NiceTabular}
    \label{tab:hyperparams_main_imagenet}
\end{table}
\begin{table}[t!]
    \caption{Hyperparameter setup for ViT-S + Bi-ICE on Places365}
    \centering
    \begin{NiceTabular}{c|c}
        Name & Value \\
        \midrule
        Batch size & $64$ \\
        Epochs & $10$ \\
        Warmup Iters. & $10$ \\
        Learning rate & $1e-4$ \\
        Explanation lambda $\lambda_{\text{expl}}$ & $0.$ \\
        Weight decay & $1e-3$ \\ 
        Sparsity lambda $\lambda_{\text{sparse}}$& $10.$ \\
        \bottomrule
    \end{NiceTabular}
    \label{tab:hyperparams_places365}
\end{table}

For the vanilla ResNet50 and ViT models, we replaced the last linear layer with a new one and fine-tuned it to compare their performance with our Bi-ICE module.
This is because our module can be seen as a drop-in replacement for the classifier head of any backbone network.
All configurations, including the vanilla models and our proposed module, were trained using the AdamW optimizer with a Linear Warmup Cosine Annealing scheduler.
The hyperparameter settings for CIFAR100, CUB-200-2011, and ImageNet are detailed in Tabels~\ref{tab:hyperparams_main_cifar100},~\ref{tab:hyperparams_main_cub} and~\ref{tab:hyperparams_main_imagenet}.
All models are trained on a NVIDIA GeForce Titan Xp GP102 (Pascal architecture, 3840 CUDA Cores @ 1.6 GHz, 384-bit bus width, 12 GB GDDR G5X memory.

\section{Further Experimental Results}
\subsection{Performance Evaluation}
\label{appsubsec:performance}

\paragraph{Ablation Study.}
Fig.~\ref{fig:ablation_imagenet_num_label} shows the experimental results on ImageNet with different numbers of labels to train vanilla ViT-S and ours.
Interestingly, our Bi-ICE outperforms vanilla ViT-S when the number of labels is relatively small ($< 800$), while the performance of Bi-ICE becomes comparable or slightly worse than that of the vanilla model as the number of labels is enormous ($\geq 800$). 

Table~\ref{tab:ablation_num_concepts_imagenet} shows the experimental results on ImageNet with different numbers of concepts. 
We selected the number of concepts leading to the best performance, which is 150 for the performance evaluation on ImageNet.

\paragraph{Quantitative Comparison of Clustering.}
As shown in Fig.~\ref{fig:tsne_z_z-bar} in the main text, we qualitatively and quantitatively compared the clustering quality of the embeddings between the vanilla ViT~($\boldsymbol{z}$) and our Bi-ICE~($\Bar{\boldsymbol{z}}$).
Table~\ref{tab:reliability_table_vit_bi-ice} presents the additional comparison of quantitative metrics using Expected Calibration Error~(ECE) and Negative Log Likelihood~(NLL)~\cite{guo2017calibration}, Brier Score~\cite{brier1950verification}, and V-score~\cite{rosenberg2007v}.
ECE measures calibration error, and NLL assesses probabilistic quality.
Brier Score quantifies the accuracy of probabilistic predictions. Overall, Bi-ICE embedding is better clustered quantitatively than vanilla ViT, leading to better classification performance.

\begin{table}\centering
\caption{ECE, Brier Score, NLL, and V-Score comparison of vanilla ViT and Bi-ICE embedding clusters on CUB-200-2011. The smaller the value, the better clustering for ECE, Brier Score, and NLL, while the higher the value, the better for the V-score.}
\label{tab:reliability_table_vit_bi-ice}
\begin{tabular}{lcccc}
\toprule
Model & ECE &  Brier Score & NLL & V-Score  \\ 
\midrule
Vanilla ViT & 0.005 & 0.1737 & 0.4915 & 0.9124 \\
Bi-ICE & \textbf{0.002} & \textbf{0.1496} & \textbf{0.3982} & \textbf{0.9366} \\
\bottomrule
\end{tabular}
\end{table}

\subsection{Hierarchical Classification Analysis on CIFAR100Super-class}
\label{appsubsec:hier_classification_cifar100superclass}

\paragraph{Setup.}
The CIFAR100 Super-class dataset is a modified version of the CIFAR100 image dataset introduced by~\cite{krizhevsky2009learning}.
It comprises 100 detailed~(fine-grained) image classes, which are grouped into 20 broader super-classes. For example, the fine-grained classes \emph{baby}, \emph{boy}, \emph{girl}, \emph{man}, and \emph{woman} are all categorized under the super-class \emph{people}.

Our Bi-ICE model can be trained using ground-truth concept labels by leveraging the Explanation loss detailed in Section~\ref{subsec:objectives} of the main text. This allows us to effectively evaluate the model's ability to align with these annotated concepts. Since the advent of deep learning models with embedded logical constraints~\cite{fischer2019dl2}, the CIFAR100 Super-class dataset has served as a benchmark to test how well such constraints can be incorporated into neural networks~(for extensive reviews, see~\cite{dash2022review, giunchiglia2022deep}). 
Our study demonstrates that the concepts learned by our model—and those used in Concept Transformer~(CT)~\cite{rigotti2021attention}—can be interpreted as constraints. In our experiments, we treated the fine-grained classes as global concept-level explanations for multi-class prediction across the 20 super-classes, utilizing a ViT-T backbone architecture.

In line with~\cite{hoernle2022multiplexnet}, we compare the performance of Bi-ICE against three baseline model groups. The first includes standard backbone architectures such as \emph{Wide ResNet 28-10}~~\cite{zagoruyko2016wide} and \emph{ViT-T}. The second group consists of neural models that embed logical constraints, such as the \emph{Hierarchical Model}~~\cite{hoernle2022multiplexnet}, \emph{DL2}~~\cite{fischer2019dl2}, and \emph{MultiplexNet}~\cite{hoernle2022multiplexnet}. The third group includes concept-based explainable models; however, with the exception of CT, these models are typically unable to manage both fine-grained and super-class predictions simultaneously. In our evaluation, we assess fine-grained class accuracy by comparing the top-1 predicted concept with the ground-truth annotation, based on the resulting attention scores.

\begin{figure}[t!]
    \centering
    \includegraphics[width=0.49\textwidth]{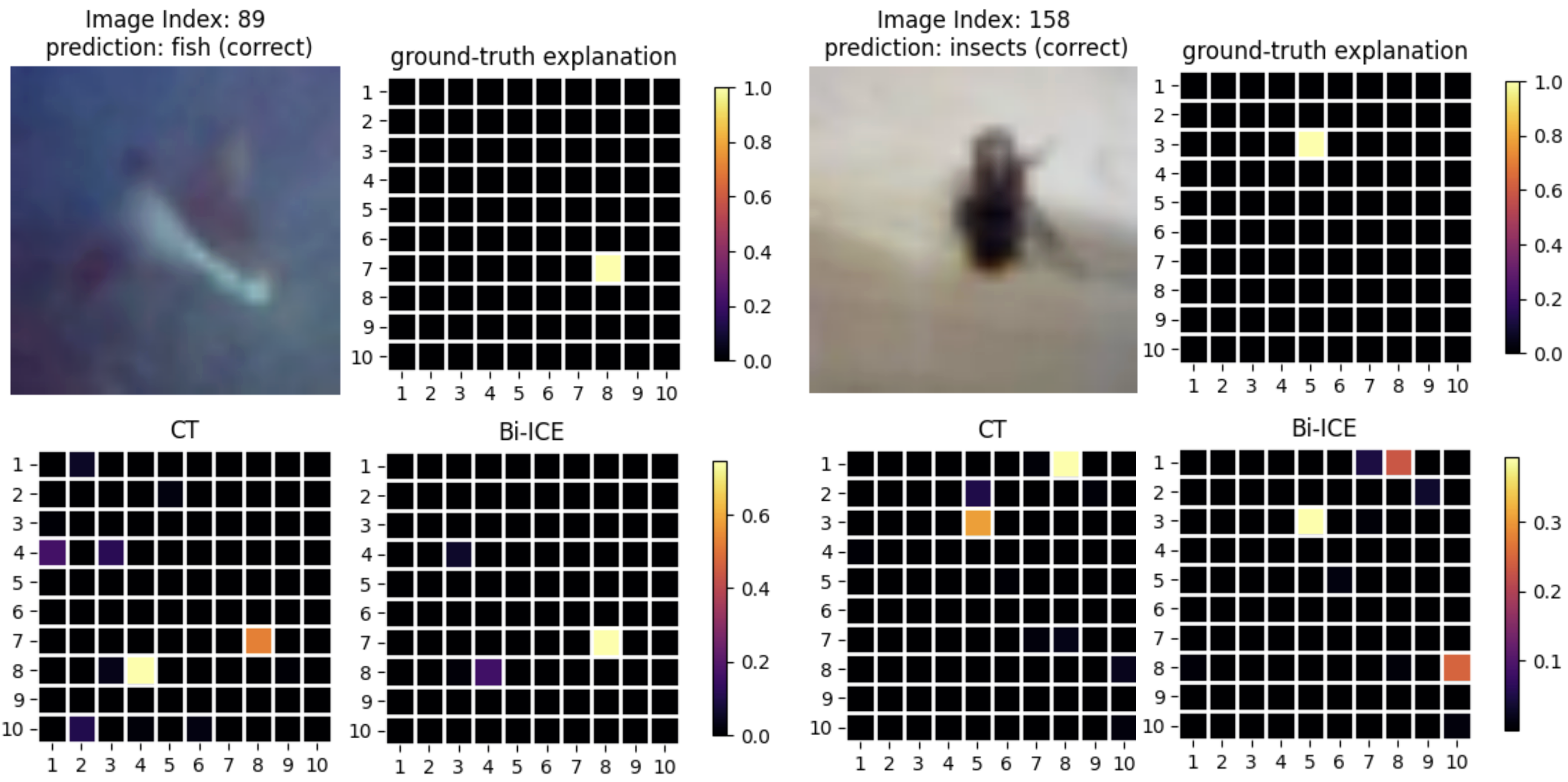}
    \caption{Comparison of class predictions for CT~\protect\cite{rigotti2021attention} and Bi-ICE in examples where both make correct CIFAR100 super-class predictions. The 100 classes are indexed from 1~(top left in 10x10~grid) to 100~(bottom right in 10x10~grid). (\textbf{Left}) Ground-truth class label is \emph{ray}~(68), but CT mispredicted as \emph{shark}~(74), whereas Bi-ICE's prediction is correct. (\textbf{Right}) Ground-truth label is \emph{cockroach}~(25), but CT incorrectly selects \emph{beetle}~(8) whereas our Bi-ICE again makes a correct class prediction.}
    \label{fig:cifar100superclass_results}
\end{figure}

\paragraph{Results.}
As illustrated in Table~\ref{tab:eval_cifar100}, our Bi-ICE method outperforms all competing baselines. Figure~\ref{fig:cifar100superclass_results} provides two illustrative examples where Bi-ICE demonstrates superior performance in both quantitative and qualitative terms compared to CT. 
One reason for CT's weaker performance on fine-grained classification may be its tendency to generate hallucinated concepts in an attempt to optimize super-class accuracy without forming meaningful latent representations. While both models achieve correct super-class predictions in these examples, their reasoning processes diverge significantly—highlighting Bi-ICE’s superior explanatory power.

\begin{table}[t!]
    \caption{Test accuracy on fine-grained class~(F.C.) and super-class~(S.C.) label prediction on CIFAR100. Notice that the classification of S.C. and F.C. are performed simultaneously and that this kind of experiment can be done in deep learning with constraints, but our method and CT are only among concept-based approaches. $\dag$ indicates results from~\cite{hoernle2022multiplexnet}. MultiplexNet used WideResNet 28-10~(36.5M) as a backbone. ProtoPNet and Deform-ProtoPNet used DenseNet-121~(8.0M) as a backbone. As a backbone for CT and our Bi-ICE, ViT-T was used~(5.7M).}
    \label{tab:eval_cifar100}
    \vspace{-2mm}
    \centering
    \resizebox{0.98\linewidth}{!}{%
    \begin{NiceTabular}{lcc}
        \toprule
        Model & F.C.\ Acc.\ (\%) & S.C.\ Acc.\ (\%) \\
        \midrule
        $\text{Vanilla WideResNet}^{\dag}$               & NA                     & $83.2_{\pm 0.2}$     \\
        Vanilla ViT-T                                & NA                     & $86.2_{\pm 0.3}$     \\
        \midrule
        $\text{Hierarchical Model}^{\dag}$          & $71.2_{\pm 0.2}$       & $84.7_{\pm 0.1}$     \\
        $\text{DL2}^{\dag}$~\cite{fischer2019dl2}    & $75.3_{\pm 0.1}$       & $84.3_{\pm 0.1}$     \\
        $\text{MultiplexNet}^{\dag}~\cite{hoernle2022multiplexnet}$ & $74.4_{\pm 0.2}$ & $85.4_{\pm 0.3}$     \\
        \midrule
        ProtoPNet~\cite{DBLP:journals/corr/abs-1806-10574} & NA & $82.3_{\pm 0.1}$\\
        Deform-ProtoPNet~\cite{DBLP:journals/corr/abs-2111-15000}  & NA & $83.7_{\pm 1.0}$ \\
        CT~\cite{rigotti2021attention}               & $73.3_{\pm 2.9}$       & $92.1_{\pm 0.2}$     \\
        \midrule
        Bi-ICE                                        & $\mathbf{83.4}_{\pm 0.1}$ & $\mathbf{93.0}_{\pm 0.1}$ \\
        \bottomrule
    \end{NiceTabular}
    }
\end{table}
%
%

\subsection{Concept-conditioned Pointing Game at Patch Resolution on CUB-200-2011}
\label{appsubsec:ccpg}

In addition to the CIFAR100Super-class experiments, we evaluate our model on \textsc{CUB-200-2011} using a pointing-game protocol tailored to concept localization.

\paragraph{Setup.}
We introduce \emph{Concept-Conditioned Pointing Game at Patch Resolution~(CPGP)}, which specifies (i)~a patch-level concept localization task, (ii)~well-defined metrics, and (iii)~reporting and interpretation guidelines. Conceptually, CPGP is a concept-level counterpart to the grid-pointing game~\cite{bohle2021convolutional}: rather than asking whether the most salient \emph{location} falls on the ground-truth object, we ask whether the most salient \emph{patch} for a \emph{given concept} lands on any ground-truth occurrence of that concept. This patch-level formulation avoids threshold tuning and makes the evaluation robust to score scaling across images and concepts.

Given a trained model, each test image is partitioned into a patch lattice of size $H_p \times W_p$ with $L = H_p \cdot W_p$ patches. As described in Sec.~\ref{subsec:analysis}, the ground-truth (GT) spatial concept annotations are represented by
$\boldsymbol{Q}^{\mathrm{spatial}}(x) \in \{0,1\}^{L \times K_{\mathrm{spatial}}}$,
where $K_{\mathrm{spatial}}$ is the number of spatial concepts and
$Q^{\mathrm{spatial}}_{i,k}(x)=1$ indicates that patch $i$ of image $x$ contains concept $k$ (otherwise $0$). Our framework produces matching-dimension concept importance scores
$\boldsymbol{\Phi}^{\mathrm{spatial}}(x) \in \mathbb{R}^{L \times K_{\mathrm{spatial}}}$ where larger values indicate stronger evidence for the concept at that patch.
We consider an image–concept pair $(x,k)$ \emph{valid} if the image contains at least one positive GT patch for $k$, i.e.,
$\sum_{i=1}^{L} \boldsymbol{1}[Q^{\mathrm{spatial}}_{i,k}(x)=1] > 0$.
Let $\mathcal{V}$ denote the set of all valid pairs.

\paragraph{Metric 1: Pointing-game accuracy (hit@1).}
For each $(x,k)\in\mathcal{V}$, we locate the most activated patch:
\begin{equation*}
    i^{\star}(x,k) \;\triangleq\; \arg\max_{\,i \in \{1,\dots,L\}} \;\Phi^{\mathrm{spatial}}_{i,k}(x).
\end{equation*}

A \emph{hit} occurs if $Q^{\mathrm{spatial}}_{i^{\star}(x,k),\,k}(x)=1$.
The overall CPGP pointing accuracy is then:
\begin{equation*}
    \mathrm{CPGP\text{-}Acc}
    \;\triangleq\;
    \frac{\sum_{(x,k)\in\mathcal{V}} \boldsymbol{1}\!\left[ Q^{\mathrm{spatial}}_{i^{\star}(x,k),\,k}(x)=1 \right]}
         {|\mathcal{V}|}.
\end{equation*}

\paragraph{Metric 2: Per-concept ROC--AUC (threshold-free ranking).}
For each concept $k$, we pool all patches across the test set into labeled score pairs
$\{(q^{(k)}_{i}, s^{(k)}_{i})\}_{i}$, where $q^{(k)}_{i}\in\{0,1\}$ comes from
$\boldsymbol{Q}^{\mathrm{spatial}}$ and $s^{(k)}_{i}=\Phi^{\mathrm{sp}}_{i,k}$.
We compute the ROC--AUC for $k$ (defined only when both positive and negative patches are present):
\begin{equation*}
    \mathrm{AUC}(k) \;\triangleq\; \mathrm{ROC\text{-}AUC}\!\left( \{(q^{(k)}_{i}, s^{(k)}_{i})\}_{i} \right).
\end{equation*}

Let $\mathcal{D}_{\mathrm{valid}}$ be the set of concepts with well-defined AUC;
we report the macro mean:
\begin{equation*}
    \mathrm{Mean\text{-}AUC} \;\triangleq\; \frac{1}{|\mathcal{D}_{\mathrm{valid}}|}
    \sum_{k \in \mathcal{D}_{\mathrm{valid}}} \mathrm{AUC}(k).
\end{equation*}

\paragraph{Results.}
Our model attains CPGP-Acc of 0.593, indicating that in 59\% of valid (image, concept) cases the single highest concept-attention peak lies on a GT-positive patch, demonstrating sharp concept-level localization.
The Mean per-concept ROC--AUC is 0.719~(averaged over concepts with both classes present), evidencing strong ranking fidelity: GT-positive patches receive higher scores than negatives even when the top patch narrowly misses.

\paragraph{Novelty relative to grid-pointing games.}
(i) \emph{Concept-conditioned evaluation:} CPGP scores localization per semantic concept (e.g., \emph{wing, color}), directly auditing concept faithfulness rather than generic class focus.
(ii) \emph{Patch-resolution localization:} Hits are computed on the native patch lattice (e.g., ViT tokens), revealing fine-grained successes and failures that coarse $2{\times}2$ or $3{\times}3$ grid pointing games may obscure.

\subsection{Analysis on Inner Interpretability}
\label{appsubsec:analysis}

\subsubsection{Computational Level}
\label{appsubsec:computational}

\paragraph{Training Setups.}
\begin{table}[t!]
    \caption{Hyperparameter setting for additional computational and implementation-level analyses.}
    \centering
    \begin{tabular}{c|c}
        Name & Value \\
        \midrule
        Batch size & $64$ \\
        Epochs & $50$ \\
        Warmup Iters. & $10$ \\
        Learning rate & $5e-5$ \\
        Explanation lambda $\lambda_{\text{expl}}$ & $10.0$ \\
        Weight decay & $1e-3$ \\ 
        Sparsity lambda $\lambda_{\text{sparse}}$& $0.5$ \\
        \bottomrule
    \end{tabular}
    \label{tab:hyperparams_computational_implementation}
    \vspace{-3mm}
\end{table}
For the analysis, we trained an additional simple model as follows. 
We combined the frozen ViT-S as a feature extractor with our Bi-ICE module. 
The model leveraged the ground-truth global and spatial annotations during the training process to retrieve the concept importance scores for both global and spatial concepts. 
Table~\ref{tab:hyperparams_computational_implementation} shows the hyperparameter setting of the model.

\begin{figure}[t!]
\centering
\includegraphics[width=0.9\columnwidth]{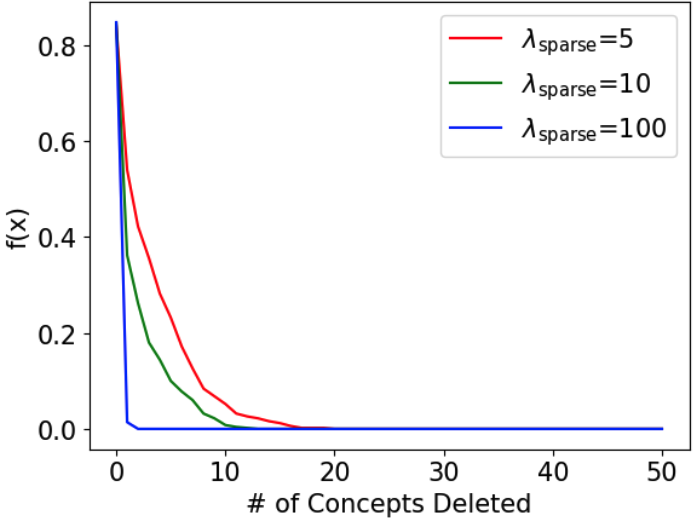}
\caption{C-deletion on ImageNet with the different $\lambda_{\text{sparse}}$. The higher the value, the sparser the concepts.}
\vspace{-3mm}
\label{fig:c_deletion_imagenet}
\end{figure}

\begin{figure*}[b!]
    \centering
    \includegraphics[width=0.93\textwidth]{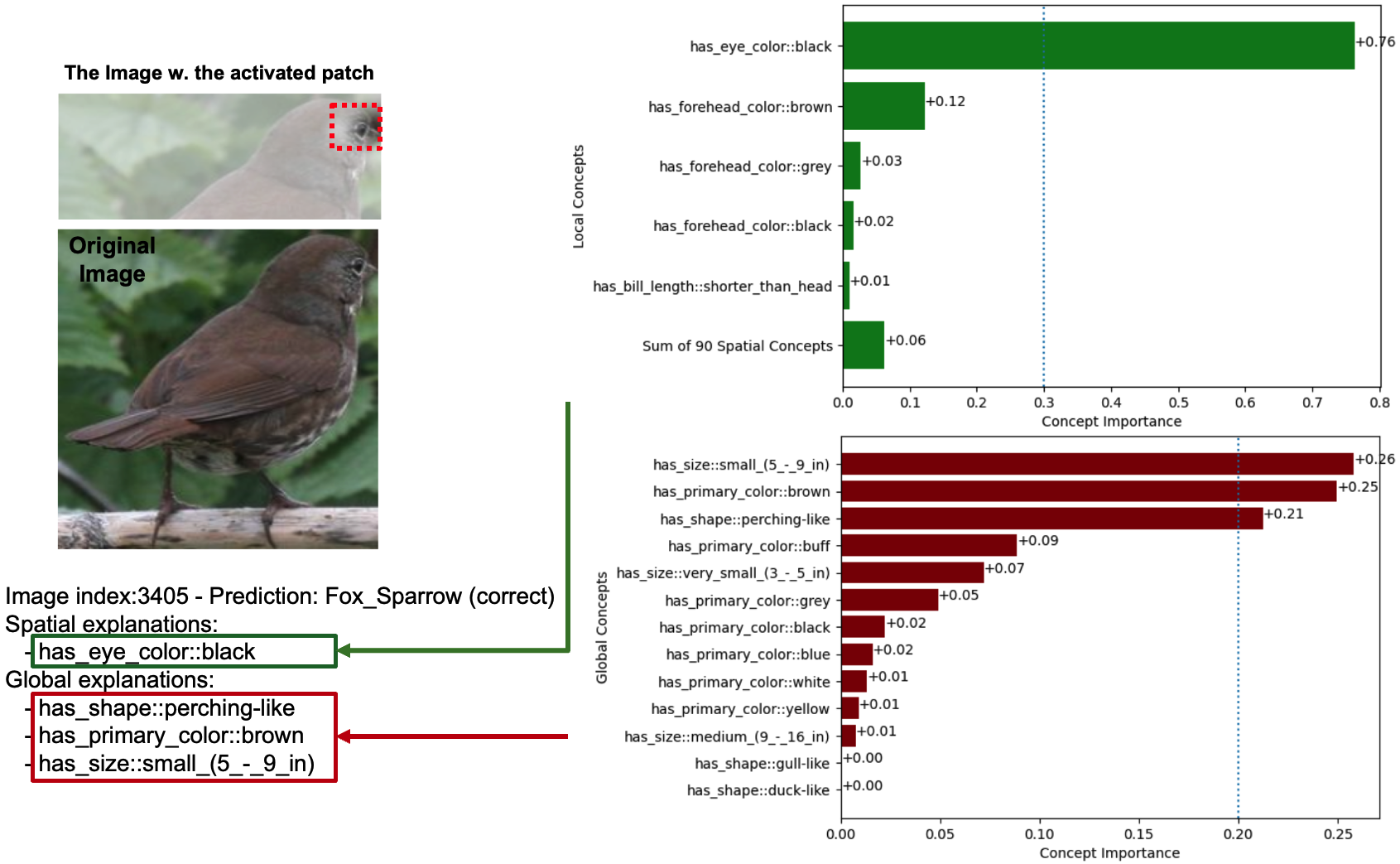}
    \caption{The additional example for the computational-level interpretability analysis. The concept contributions of the class \textit{`Fox Sparrow'} illustrate how the concepts influence the model's decision. The dashed red box indicates the activated patch ($>$ 0.6). Green for spatial, and red for global. 
    The activated patch represents the region of the bird's eye and the forehead area. 
    Accordingly, the spatial concept ``has\_eye\_color::black" holds the highest contribution score (0.76), reflecting the clear presence of black in the bird's eye.
    Similarly, ``has\_forehead\_color::brown" also shows the second highest importance, aligning with the brown coloration of the bird's forehead in the red box.
    In the global explanations, the bird's size and primary color were key concepts influencing the prediction.
    Concepts like ``has\_size::small\_(5\_-\_9\_in)" and ``has\_primary\_color::brown" rank the highest, with scores of 0.26 and 0.25, respectively, matching the Fox Sparrow's small size and dominant brown plumage.
    }
    \vspace{-3mm}
    \label{fig:concept_importance_score_example_02}
\end{figure*}

\begin{figure*}[t!]
    \centering
    \includegraphics[width=0.98\textwidth]{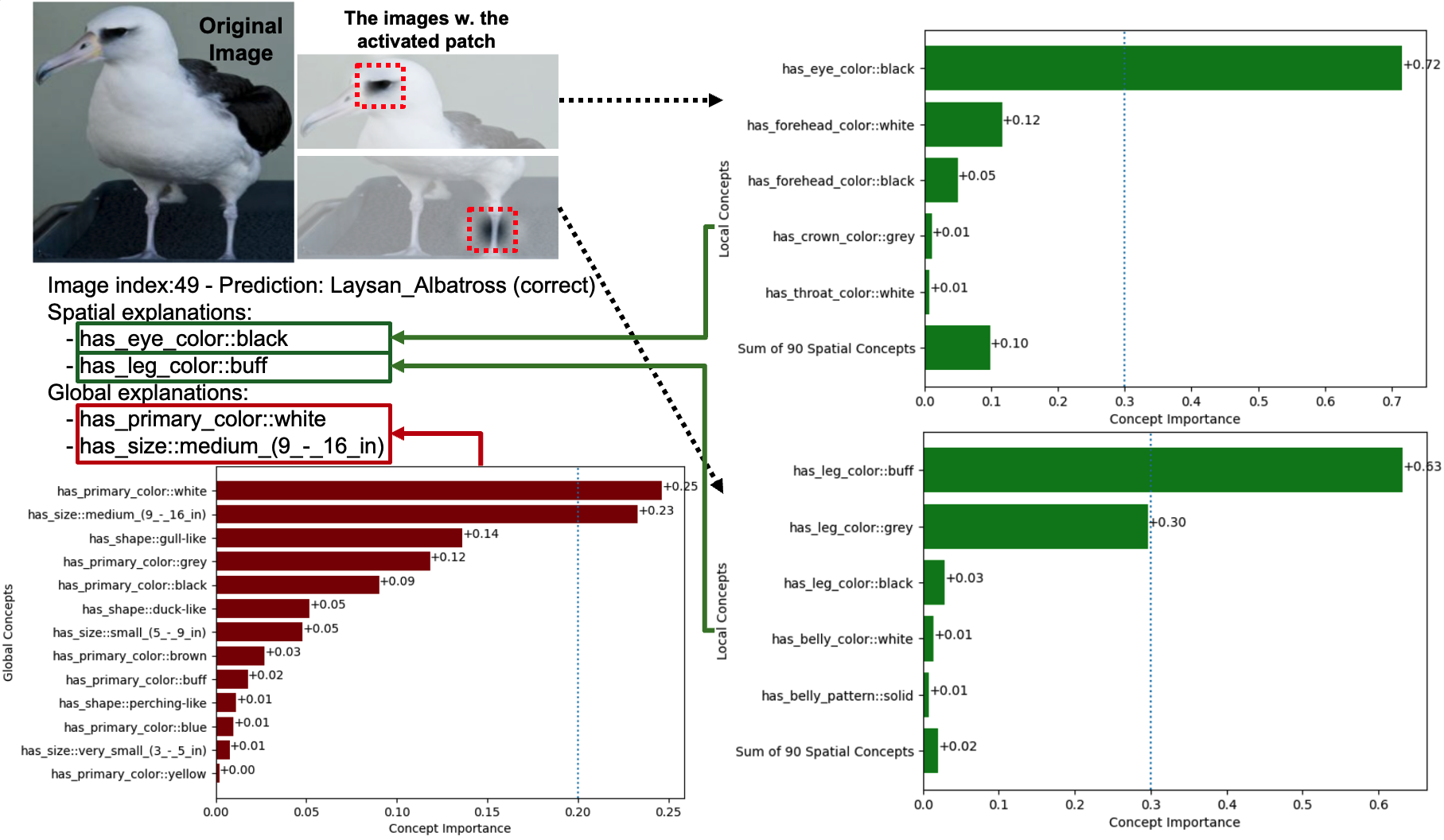}
    \caption{The additional example for the computational-level interpretability analysis. The concept Contributions of the class \textit{`Laysan Albatross'} illustrate how the concepts influence the model's decision. The dashed red boxes indicate the activated patches ($>$ 0.6). Green for spatial, and red for global.
    The activated patches represent the regions of the bird's eye and the leg areas, respectively. 
    Accordingly, the spatial concept ``has\_eye\_color::black" holds the highest contribution score (0.72) with the first patch, reflecting the dark eyes visible in the first patch.
    Similarly, ``has\_leg\_color::buff" contributes significantly with a score of 0.63, aligning with the distinct buff-colored legs captured in the second patch.
    In the global explanations, broader concepts like ``has\_primary\_color::white" and ``has\_size::medium\_(9\_\-\_16\_in)" are the most influential, with scores of 0.25 and 0.23, respectively, matching the Laysan Albatross's overall appearance and size.
    }
    \vspace{-3mm}
    \label{fig:concept_importance_score_example_01}
\end{figure*}

\begin{figure*}[t!]
\centering
\begin{subfigure}[t!]{0.9\textwidth}
   \includegraphics[width=1\linewidth]{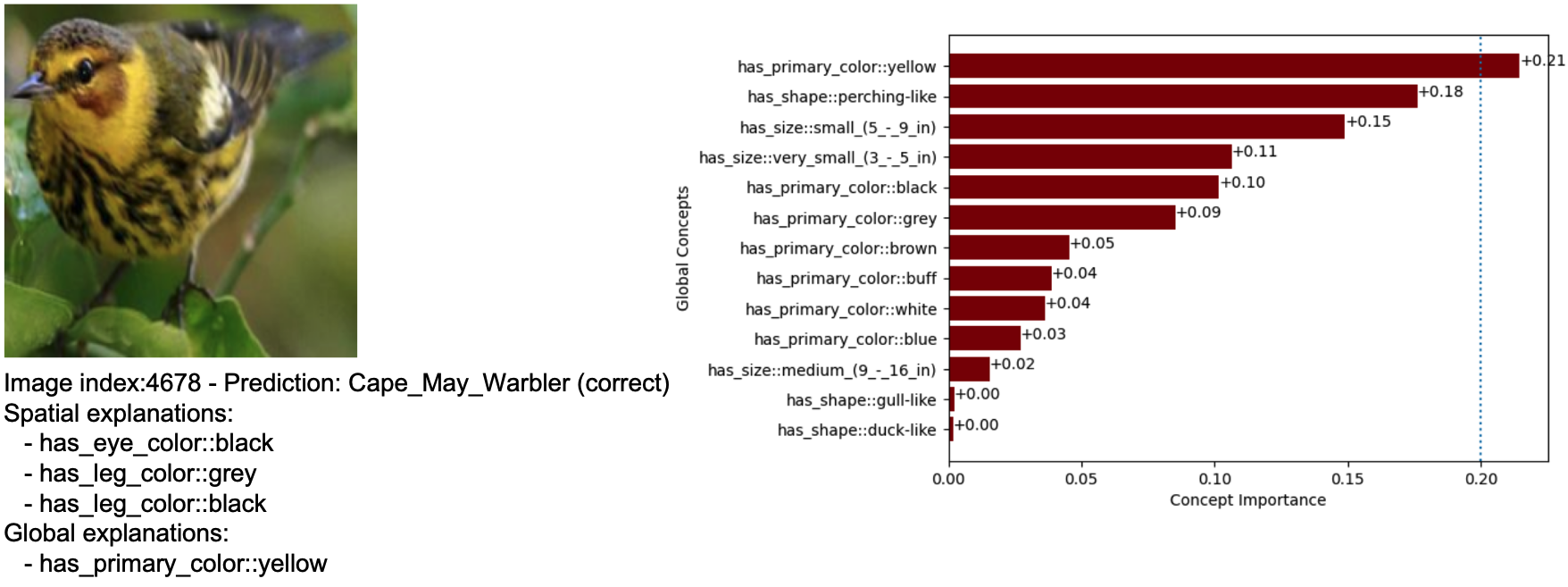}
   \caption{Concept importance score of global explanation. 
    Among the global concepts, ``has\_primary\_color::yellow" holds the highest contribution score (0.21), reflecting the bird's dominant yellow plumage.
    This is followed by ``has\_shape::perching-like" (0.18) and "has\_size::small\_(5\_-\_9\_in)" (0.15), which align with the bird's distinctive perching posture and small size.
    Other color-related concepts, such as ``has\_primary\_color::black" (0.10) and ``has\_primary\_color::grey" (0.09), further highlight the importance of color features in the model's prediction.}
   \label{fig:example_03_global} 
\end{subfigure}
\begin{subfigure}[t!]{0.98\textwidth}
   \includegraphics[width=1\linewidth]{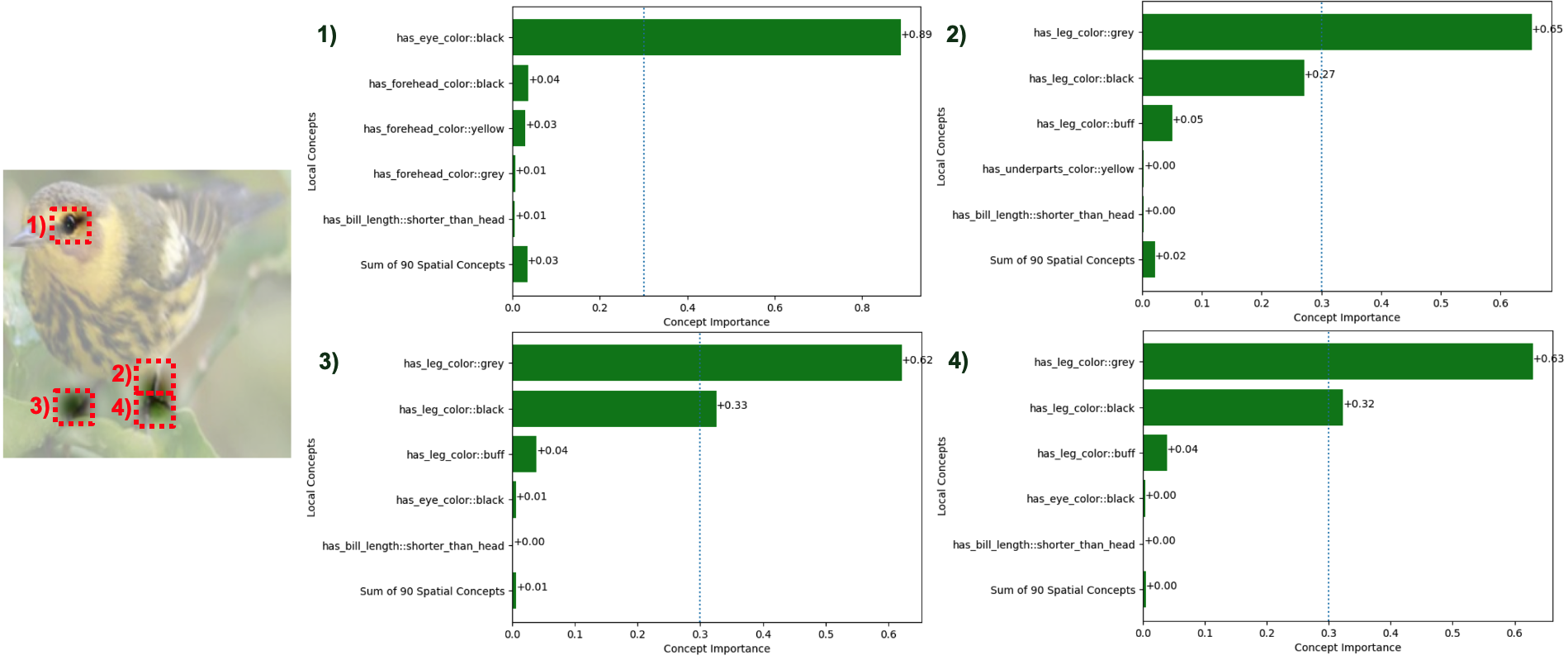}
   \caption{Concept importance score of spatial explanations with their activated patches (red boxes).
    \textbf{1)} Patch 1 (Head region):
    The spatial concept ``has\_eye\_color::black" contributes the highest score (0.89), accurately reflecting the bird's black eye, a prominent feature in this region.
    \textbf{2)} Patch 2 (Upper leg region): The concept ``has\_leg\_color::grey" holds the highest importance with a score of 0.65, emphasizing the grey coloration observed on the bird's upper leg.
    ``has\_leg\_color::black" also contributes with a score of 0.27, highlighting additional black tones in the same area.
    \textbf{3)} Patch 3 (Lower region): Similar to Patch 2, ``has\_leg\_color::grey" dominates with a score of 0.62, representing the grey-colored lower leg. ``has\_leg\_color::black" follows with a contribution of 0.33.
    \textbf{4)} Patch 4 (Feet region): Again, ``has\_leg\_color::grey" holds the highest contribution (0.63), accurately reflecting the grey feet of the bird.
    ``has\_leg\_color::black" is the second most important feature, with a score of 0.32. }
   \label{fig:example_03_spatial}
\end{subfigure}
\caption[Two numerical solutions]{The additional example for the computational-level interpretability analysis. The concept Contributions of the class \textit{`Cape May Warbler'} illustrate how the concepts influence the model's decision. The dashed red boxes indicate the activated patches ($>$ 0.6). Green for spatial, and red for global.
The global and spatial explanations reveal how the model integrates both overarching and localized features-such as color, size, and posture-to accurately classify the class `Cape May Warbler'.
}
\label{fig:concept_importance_score_example_03}
\end{figure*}

\paragraph{Additional Figures and Results.}
Fig~\ref{fig:c_deletion_imagenet} shows the C-deletion experiment on ImageNet with varying sparsity lambda $\lambda_\text{sparse}$.
Figs.~\ref{fig:concept_importance_score_example_02},~\ref{fig:concept_importance_score_example_01}, and~\ref{fig:concept_importance_score_example_03} present additional examples of concept importance scores with varying numbers of activated patches, generated using our trained model with the aforementioned setting.
Detailed explanations are included in the captions of each figure.

%


\subsubsection{Algorithmic Level}

\paragraph{Training Setups in the main text.}
For this experiment, we trained a simplified model with 15 concept vectors and 50 class labels using a ViT-S backbone without ground-truth concept annotations.
This configuration was specifically designed to ensure a focused and interpretable analysis, enabling clear observation of the convergence of concept vectors both within each class and across different instances.

\begin{figure*}[t!]
\centering
\begin{subfigure}[t!]{0.92\textwidth}
   \includegraphics[width=1\linewidth]{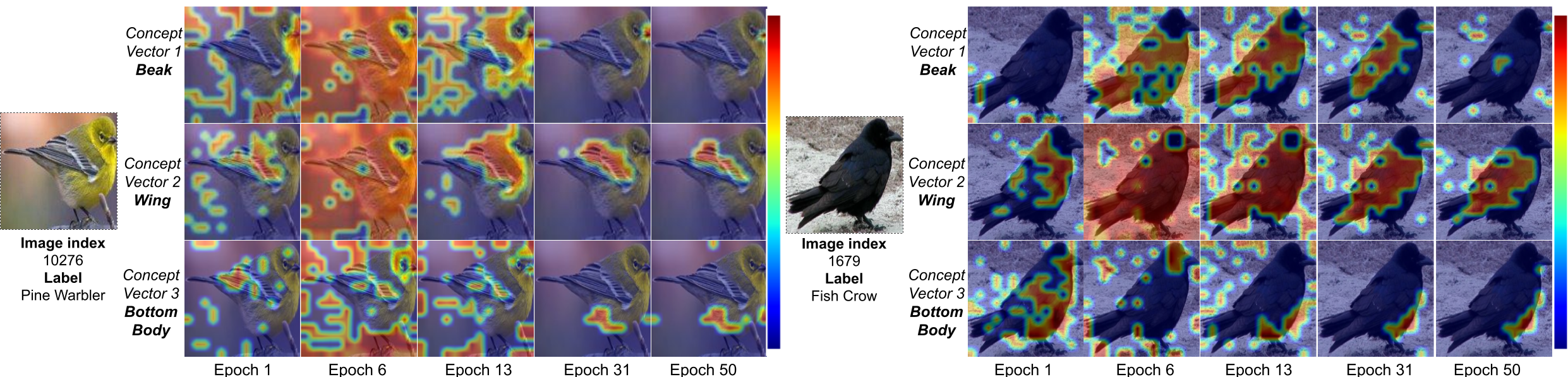}
   \caption{Enlarged view of concept vector convergence and refinement~(epoch 1 to 50). In the main text, a smaller version of this figure shows how concept vectors converge and refine from epoch 1 to 50. Here, we provide an enlarged version to better capture the details, highlighting the developmental process.}
   \label{fig:Algo_CUB_1} 
\end{subfigure}
\begin{subfigure}[t!]{0.92\textwidth}
   \includegraphics[width=1\linewidth]{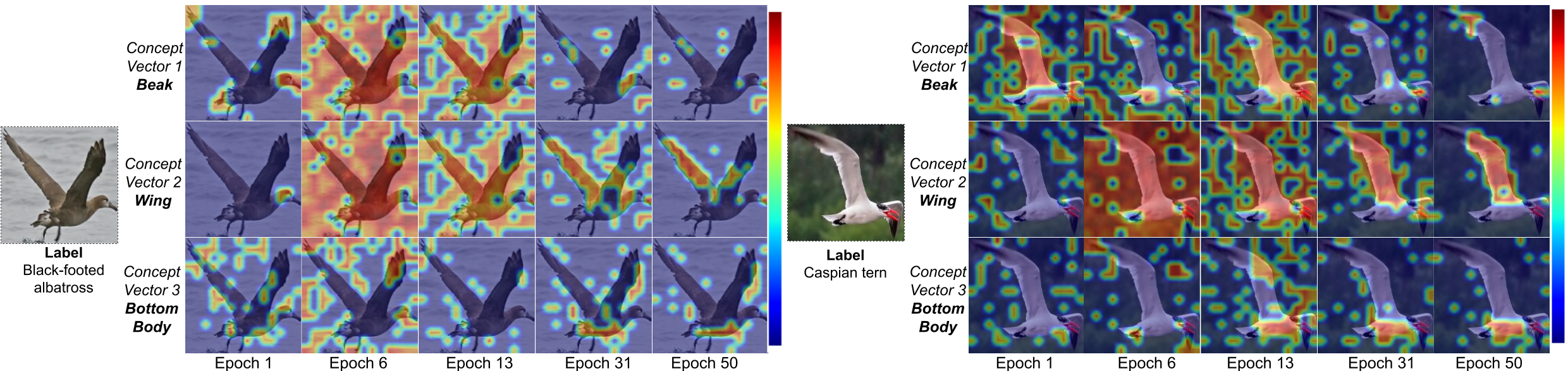}
   \caption{New examples showing concept vector convergence and refinement. This panel introduces additional examples not shown in the main text. The model exhibits a similar developmental trend with different instances, where each concept vector gradually converges and refines into specific regions, as seen in the earlier example.}
   \label{fig:Algo_CUB_2} 
\end{subfigure}
\begin{subfigure}[t!]{0.92\textwidth}
   \includegraphics[width=1\linewidth]{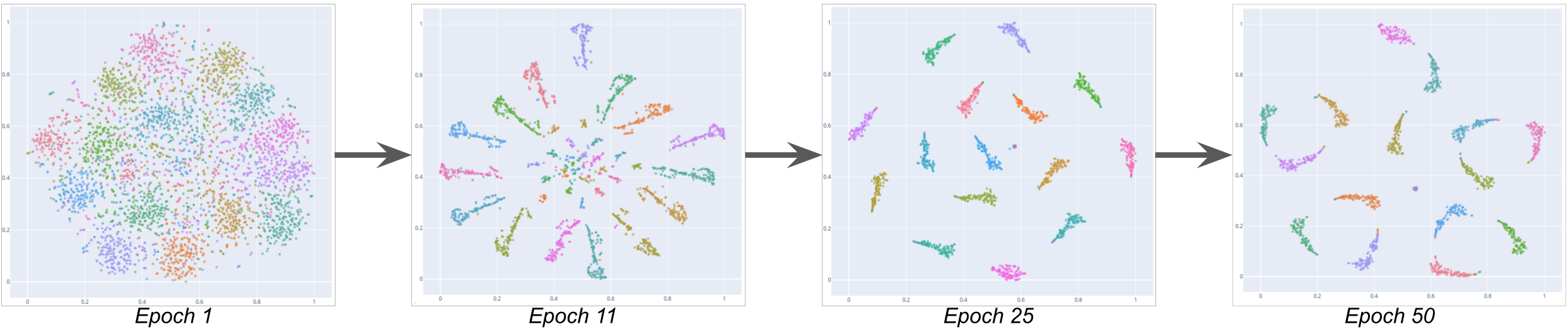}
   \caption{t-SNE of concept vector convergence over epochs. To analyze the internal structure further, t-SNE visualizations show how concept vectors converge and refine across epochs. The clusters form as epochs advance, and by later epochs, the concept vectors align closely within their respective clusters.}
   \label{fig:Algo_CUB_3}
\end{subfigure}
\begin{subfigure}[t!]{0.9\textwidth}
   \includegraphics[width=1\linewidth]{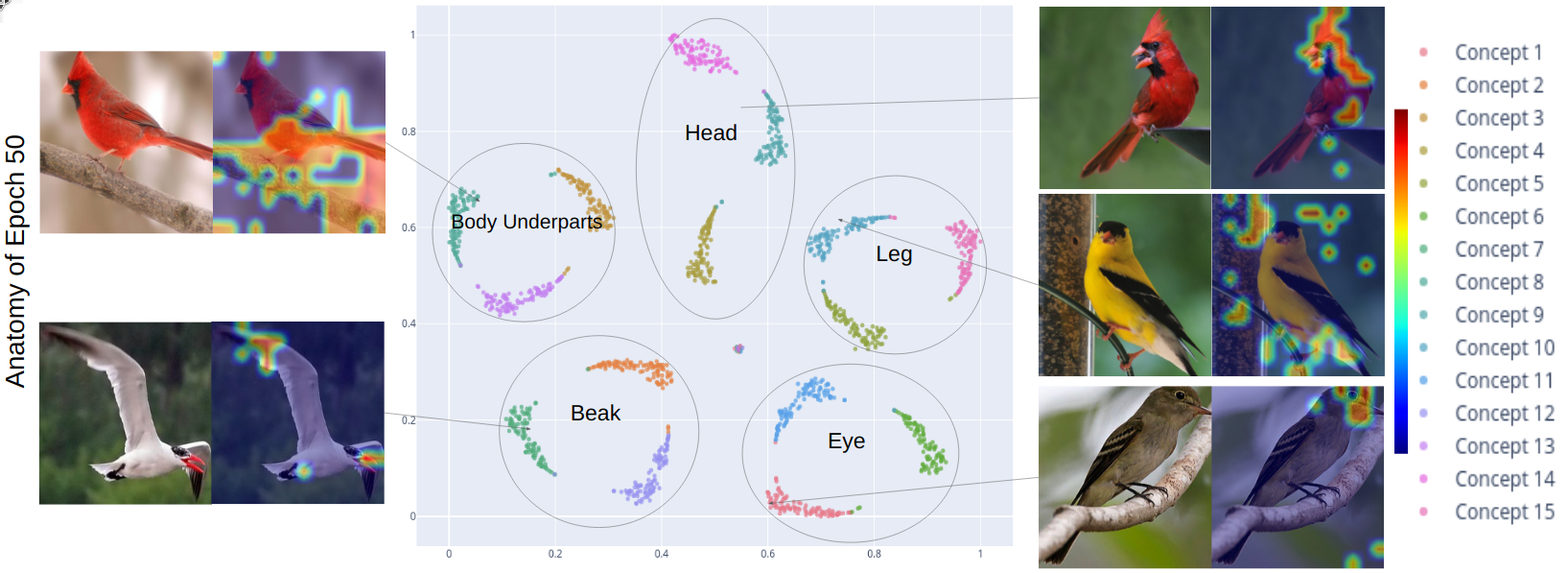}
   \caption{t-SNE at epoch 50 showing final clustering. t-SNE visualization for epoch 50~(an enlarged version of the last panel in Fig.\ref{fig:Algo_CUB_3}) demonstrates the fully developed state. At this stage, the clusters are well-formed, with each concept vector distinctly grouped.}
   \label{fig:Algo_CUB_4}
\end{subfigure}
\caption[Algo_CUB_figure_title]{Convergence and Refinement of Concept Vectors Across Epochs on CUB instances. This figure illustrates the developmental process of the model as it refines its concept vectors over training epochs, showing both individual instance-level convergence and the overall clustering behavior of internal representations.}
\label{fig:Algo_CUB_figure}
\end{figure*}

\paragraph{Additional Figures on CUB.}
In Fig.~\ref{fig:Algo_CUB_1}, we present an enlarged view of the concept vector convergence and refinement visualization initially shown in the main text~(Fig.~\ref{fig:2_algorithmic}), offering a clearer perspective on the developmental process.
This enlarged figure demonstrates how each concept vector focuses on specific semantic regions of bird images, such as the beak, wings, and lower body.
As shown in Fig.~\ref{fig:Algo_CUB_2}, the framework consistently converges across different instances to identify the same meaningful regions, underscoring its reliability in capturing interpretable concepts.

In the t-SNE analysis shown in Fig.~\ref{fig:Algo_CUB_3}, we used the experimental setup to include 50 class labels and 15 concepts, utilizing a ViT-small backbone.
This offers a comprehensive perspective on the clustering behavior of concept vectors.
Additionally, the t-SNE visualization at the final epoch~(epoch 50), presented in Fig.~\ref{fig:Algo_CUB_4}, shows that the model groups semantically similar features together.
Notably, we observed well-defined clusters for features such as body underparts, head, legs, eyes, and beak regions. 

\begin{figure*}[t!]
\centering
\begin{subfigure}[t!]{0.98\textwidth}
   \includegraphics[width=1\linewidth]{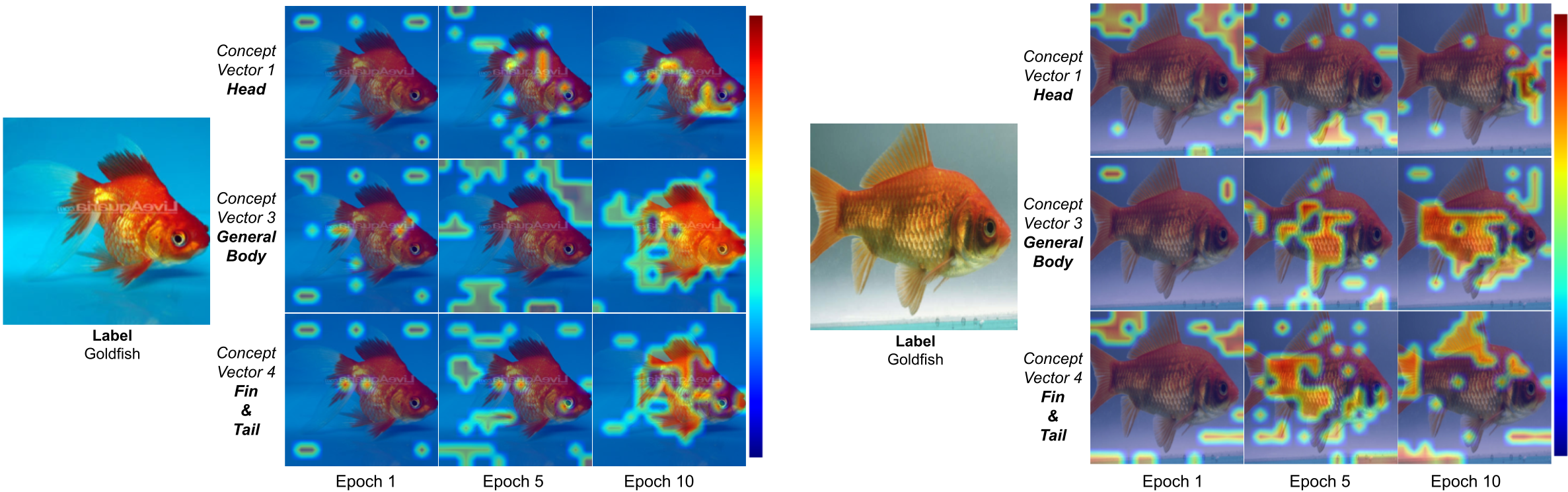}
   \caption{Instance : Goldfish – Concept Vector Convergence and Refinement. This panel shows the evolution of concept vectors from epoch 1 to 10 for an image labeled as``Goldfish.'' The concept vectors effectively identify different parts of the fish, such as the head, body, and fin with tail.}
   \label{fig:Algo_img_1_2} 
\end{subfigure}
\begin{subfigure}[t!]{0.98\textwidth}
   \includegraphics[width=1\linewidth]{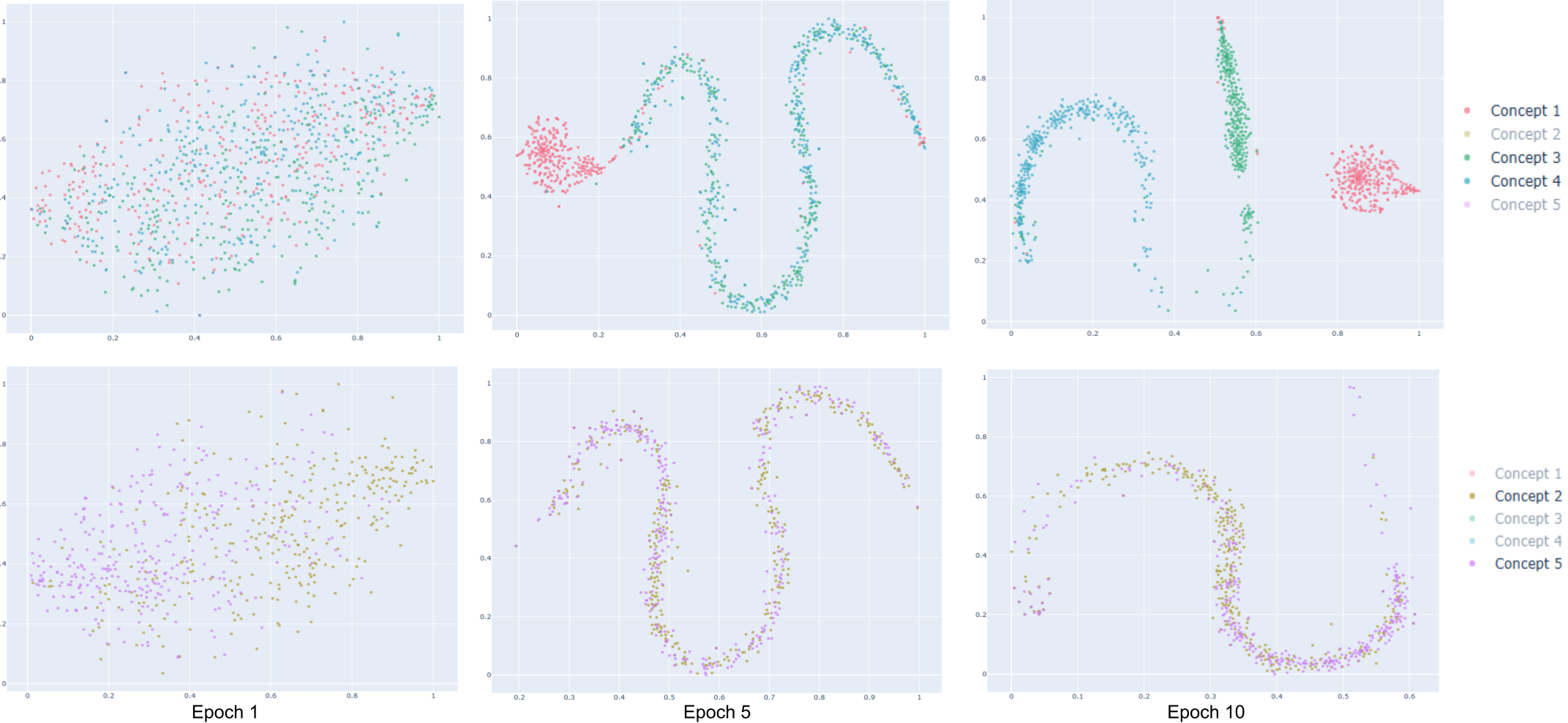}
   \caption{t-SNE Visualization of Concept Vector Clustering. This panel shows the t-SNE projections of five internal concept vectors across epochs. Clusters 1, 3, and 4 group well, indicating meaningful internal representations, while clusters 2 and 5 overlap significantly, highlighting redundancy or focus on irrelevant image regions. Over epochs, convergence and refinement of these vectors can be observed.}
   \label{fig:Algo_img_1_3} 
\end{subfigure}
\caption[Algo_img_1_figure_title]{Convergence and Refinement of Concept Vectors Across Epochs on ImageNet instances. This figure illustrates the developmental process of the model as it refines its concept vectors over training epochs, showing both individual instance-level convergence and the overall clustering behavior of internal representations. The submodel presented here was trained on ``Goldfish'' with other fish-like classes to capture semantically meaningful features, such as the fish's body, fins, and broader scene elements. 
}
\label{fig:Algo_img_1_figure}
\end{figure*}

\paragraph{Additional Figures on ImageNet.}
The algorithmic-level experiments on ImageNet are designed to analyze the behavior of concept vectors to demonstrate the applicability and versatility of our proposed method on a more challenging dataset.
In the first set of ImageNet results shown in Fig.~\ref{fig:Algo_img_1_figure}, we focus on fish-like classes, training the model with 20 class labels and five concepts using ViT-S backbone. 
Through out our t-SNE graph (Figs.~\ref{fig:Algo_img_1_3}), we observed that only three concept vectors are sufficient to classify fish-like classes: Concept Vector 1 emphasizes the head of the fish, Concept Vector 3 captures the general body, and Concept Vector 4 focuses on the fins and tail.

\begin{figure*}[t!]
\centering
\begin{subfigure}[t!]{0.98\textwidth}
   \includegraphics[width=1\linewidth]{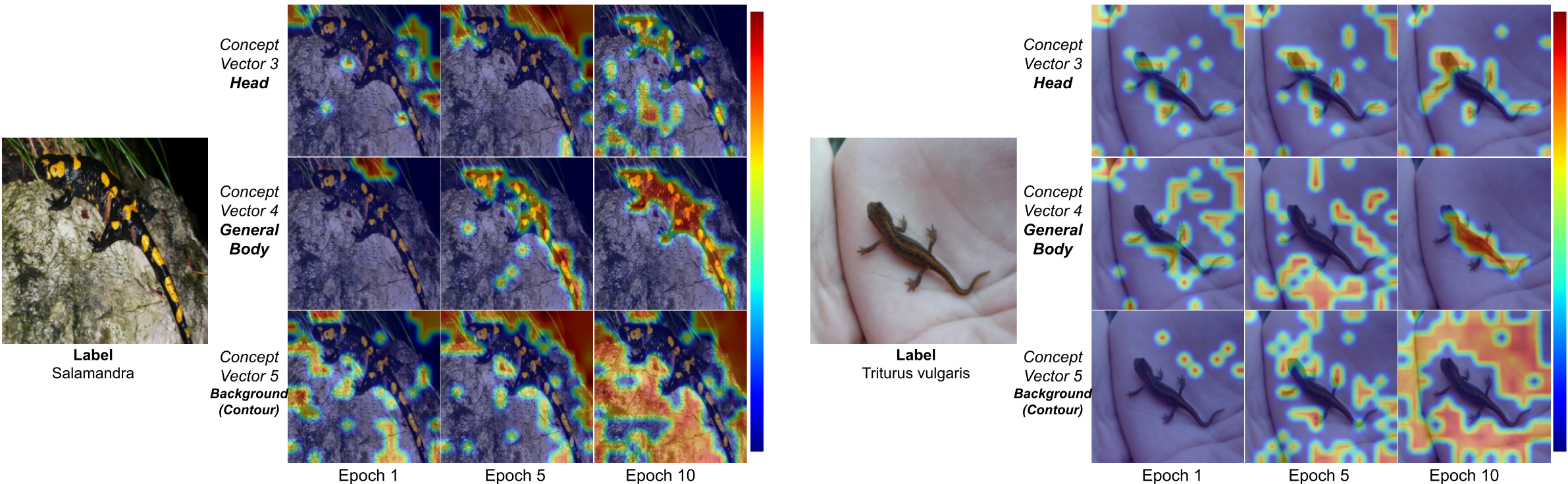}
   \caption{Instance: Amphibians~(Salamander and Triturus vulgaris) – Concept Vector Convergence and Refinement. This panel shows the evolution of concept vectors from epoch 1 to 10 for an image labeled as ``Salamander" and ``Triturus vulgaris." Each concept vector identifies distinct parts of the image: both head, its general body, and the background~(contours).}
   \label{fig:Algo_img_2_1} 
\end{subfigure}
\begin{subfigure}[t!]{0.98\textwidth}
   \includegraphics[width=1\linewidth]{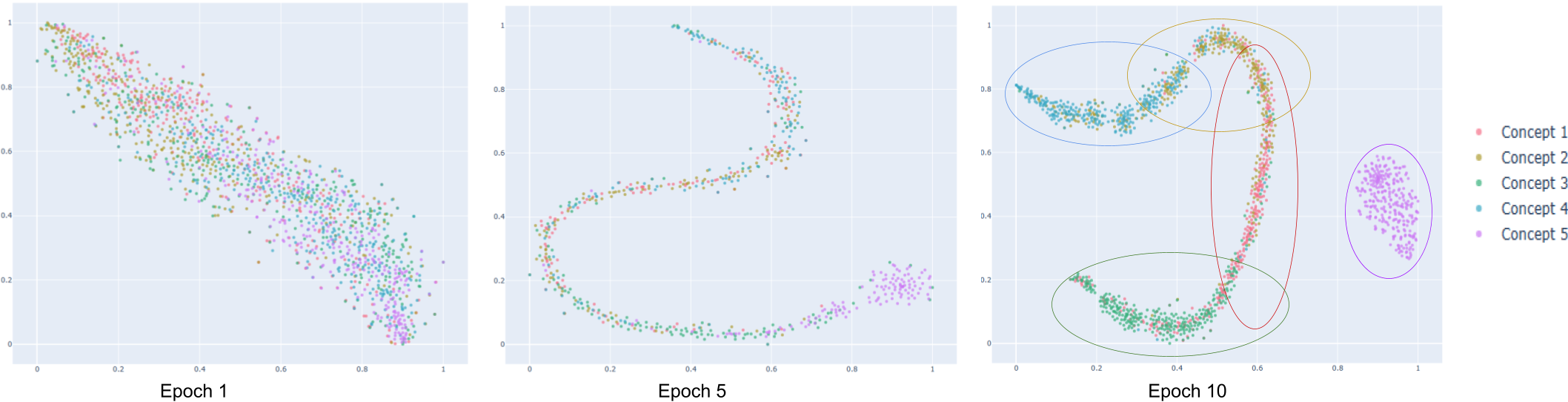}
   \caption{t-SNE Visualization: Concept Vector Separation Across Epochs. This panel displays the t-SNE visualization of concept vectors for the same instance above. As epochs progress, the separation of concept vectors improves, with minimal overlap between clusters. This demonstrates the model's ability to effectively disentangle and organize features without any ground truth guidance, although some overlap between concept vectors remains.}
   \label{fig:Algo_img_2_2} 
\end{subfigure}
\caption[Algo_img_2_figure_title]{Convergence and Refinement of Concept Vectors Across Epochs on ImageNet instances. This figure explores how concept vectors evolve and refine over epochs for two instances, alongside t-SNE projections highlighting separable and overlapping internal representations. The submodel presented here was trained on amphibian classes~(e.g., Salamander and Triturus vulgaris), isolating features like the head, body, and background contours.}
\label{fig:Algo_img_2_figure}
\end{figure*}

\begin{figure*}[t!]
\centering
\begin{subfigure}[t!]{0.98\textwidth}
   \includegraphics[width=1\linewidth]{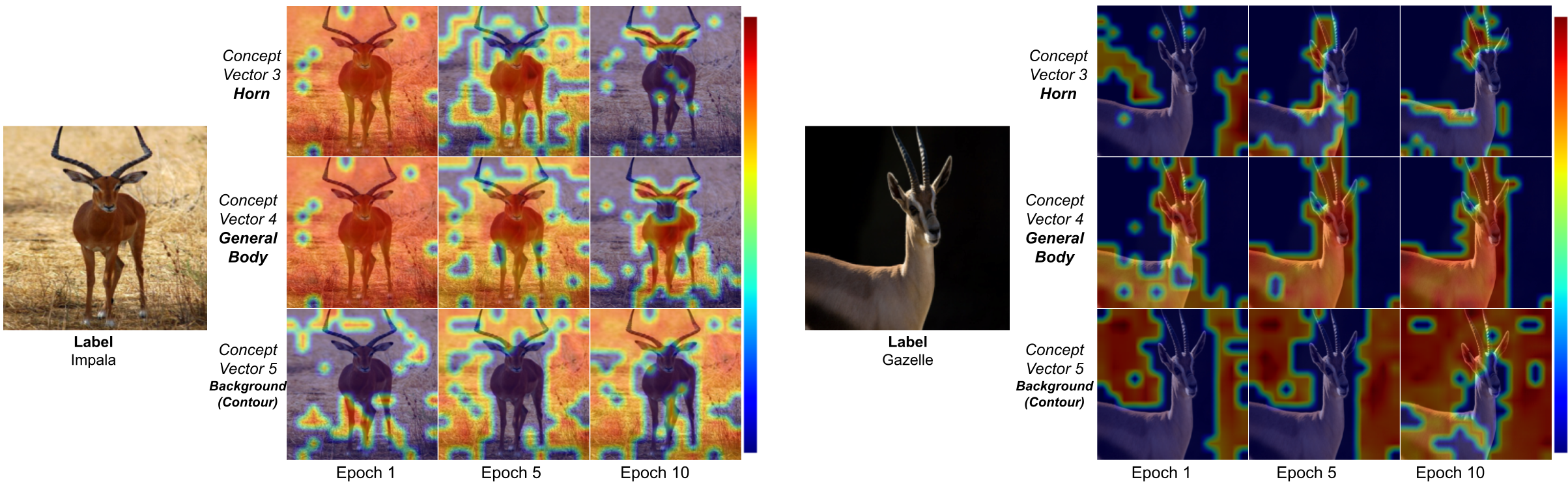}
   \caption{Instance : Impala and Gazelle – Concept Vector Convergence and Refinement. This panel shows the evolution of concept vectors from epoch 1 to 10 for an image labeled as "Impala" and "Gazelle". Each concept vector isolates distinct aspects of the image: the horns, the general body, and the background (contours).}
   \label{fig:Algo_img_3_1} 
\end{subfigure}
\begin{subfigure}[t!]{0.98\textwidth}
   \includegraphics[width=1\linewidth]{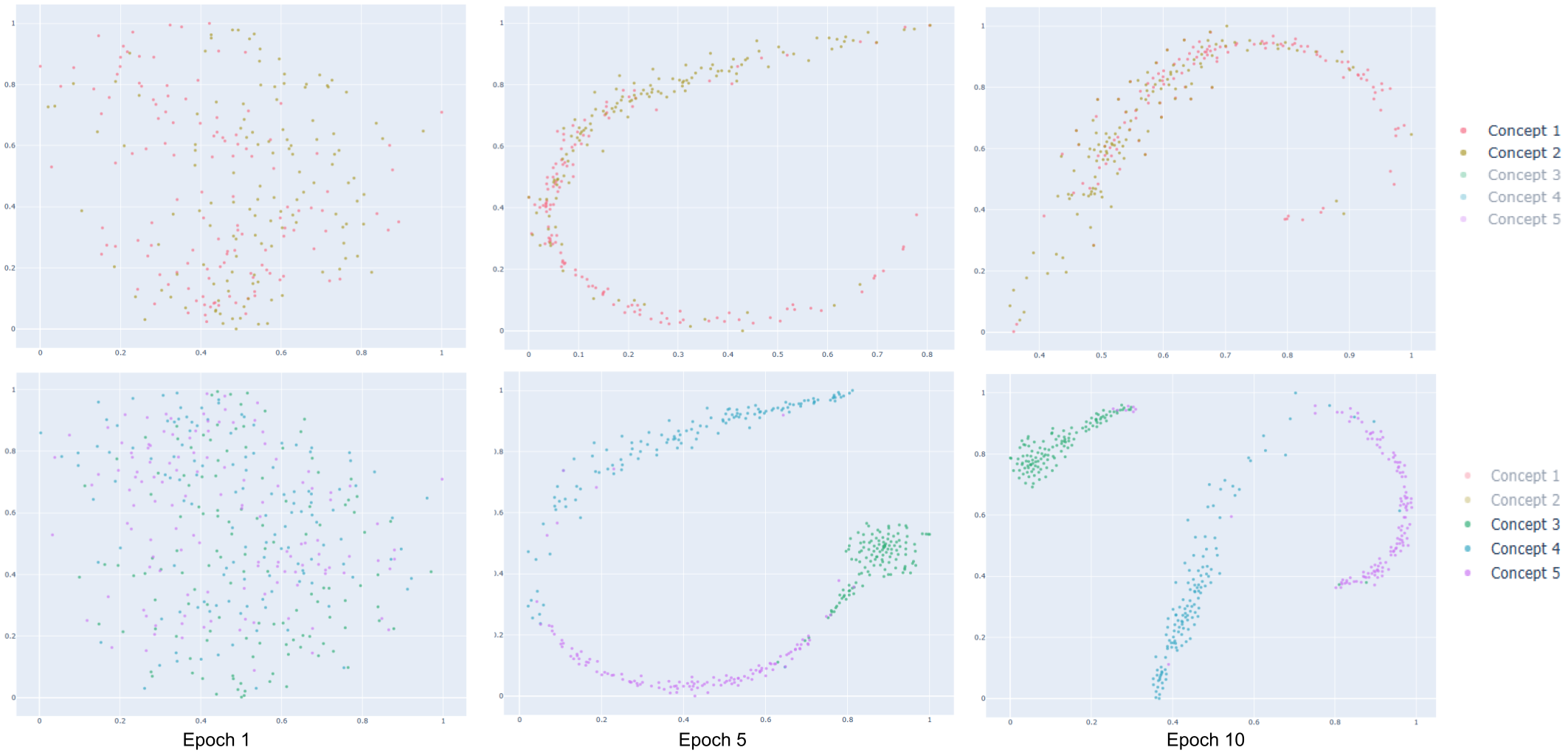}
   \caption{t-SNE Visualization of Concept Vector Clustering. This panel highlights the t-SNE projections of concept vectors across epochs. Concept vectors 1 and 2 overlap significantly, indicating redundancy and focus on less interpretable or irrelevant image areas. In contrast, concept vectors 3, 4, and 5 cluster separately, aligning well with meaningful features and demonstrating effective disentanglement over training epochs.}
   \label{fig:Algo_img_3_2} 
\end{subfigure}
\caption[Algo_img_3_figure_title]{Convergence and Refinement of Concept Vectors Across Epochs on ImageNet instances. This figure explores how concept vectors converge and refine over epochs for two instances, alongside t-SNE projections highlighting separable and overlapping internal representations. The submodel presented here was trained on antelope-like classes~(e.g., Impala and Gazelle), focusing on features such as horns, the general body, and the background.}
\label{fig:Algo_img_3_figure}
\end{figure*}

\begin{figure*}[t!]
\centering
\begin{subfigure}[t!]{0.98\textwidth}
   \includegraphics[width=1\linewidth]{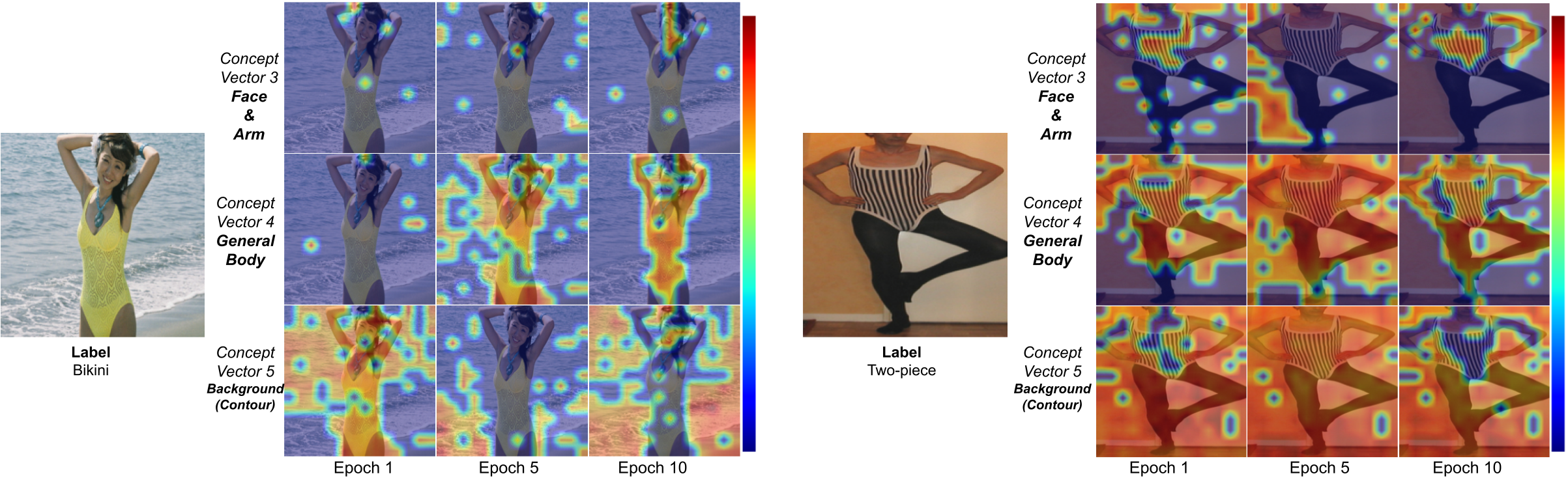}
   \caption{Instance : Bikini and Two-Piece – Concept Vector Convergence and Refinement. This panel shows the evolution of concept vectors from epoch 1 to 10 for an image labeled as "Bikini" and "Two-Piece." The vectors isolate distinct components: the face and arms, the general body, and the background (contours).}
   \label{fig:Algo_img_4_1} 
\end{subfigure}
\begin{subfigure}[t!]{0.98\textwidth}
   \includegraphics[width=1\linewidth]{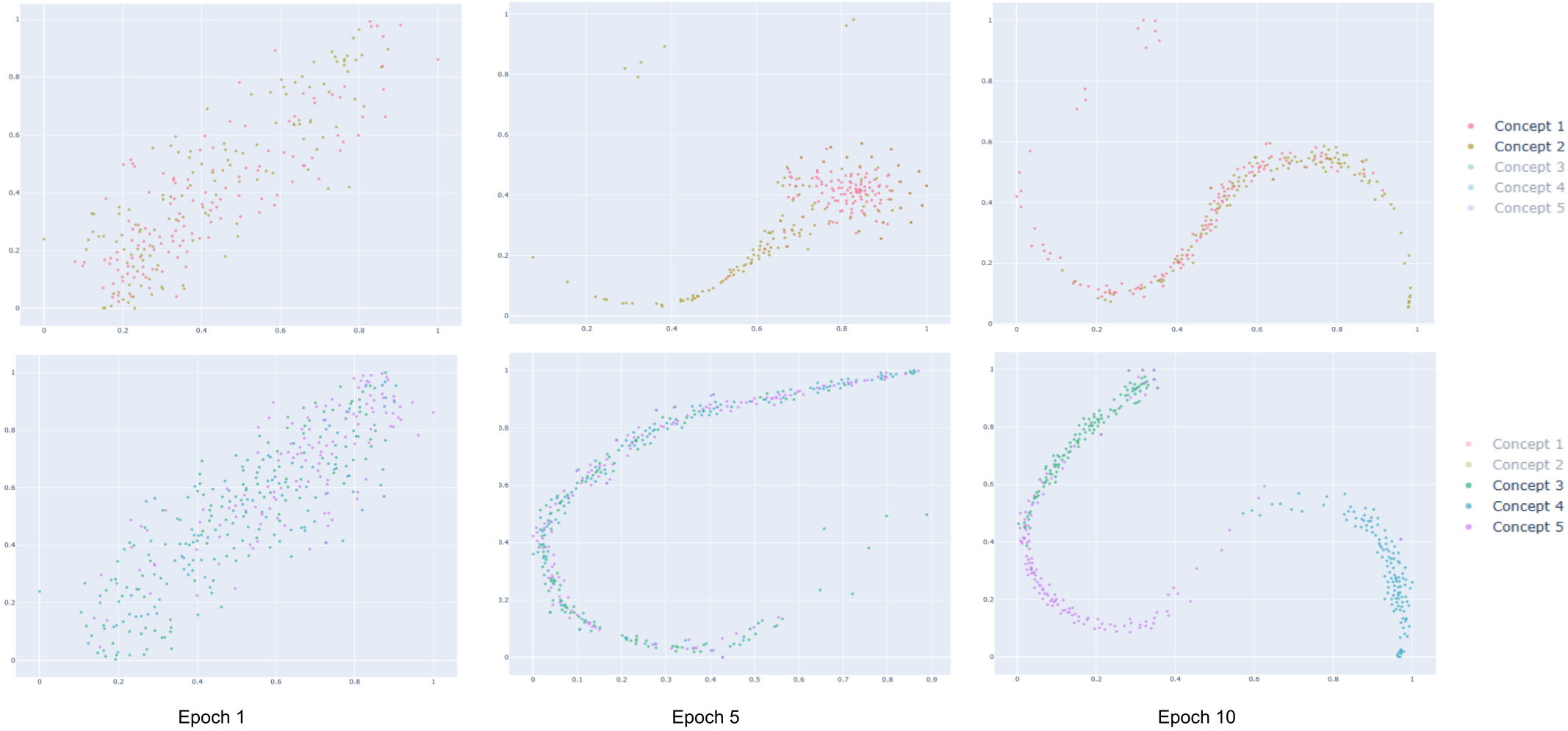}
   \caption{t-SNE Visualization of Concept Vector Clustering. This panel highlights the t-SNE projections of concept vectors across epochs. Concept vectors 1 and 2 overlap significantly, indicating redundancy and focus on less interpretable or irrelevant image areas. In contrast, concept vectors 3, 4, and 5 cluster separately, aligning well with meaningful features and demonstrating effective disentanglement over training epochs.}
   \label{fig:Algo_img_4_2} 
\end{subfigure}
\caption[Algo_img_4_figure_title]{Convergence and Refinement of Concept Vectors Across Epochs on ImageNet instances. This figure showcases the refinement of concept vectors over epochs for two instances involving human-centered images, alongside t-SNE visualizations illustrating the clustering behavior of these internal representations. The submodel presented here was trained on human-centered image classes~(e.g., Bikini and Two-Piece) to isolate specific body parts and their surrounding context.}
\label{fig:Algo_img_4_figure}
\end{figure*}

\begin{figure*}[t!]
\centering
\begin{subfigure}[t!]{0.98\textwidth}
   \includegraphics[width=1\linewidth]{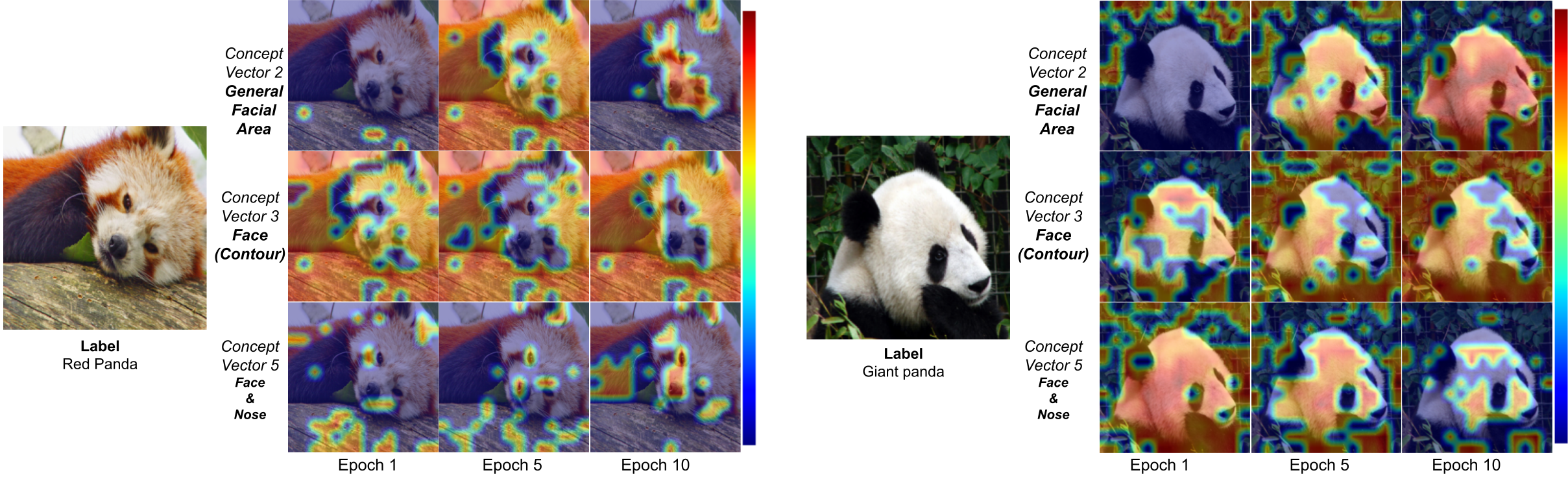}
   \caption{Instance : Red Panda and Giant Panda – Concept Vector Convergence and Refinement. This panel shows the evolution of concept vectors from epoch 1 to 10 for an image labeled as "Red Panda" and "Giant Panda." The model captures the general facial area, facial contour, and specific features such as the face and nose.}
   \label{fig:Algo_img_5_1} 
\end{subfigure}
\begin{subfigure}[t!]{0.98\textwidth}
   \includegraphics[width=1\linewidth]{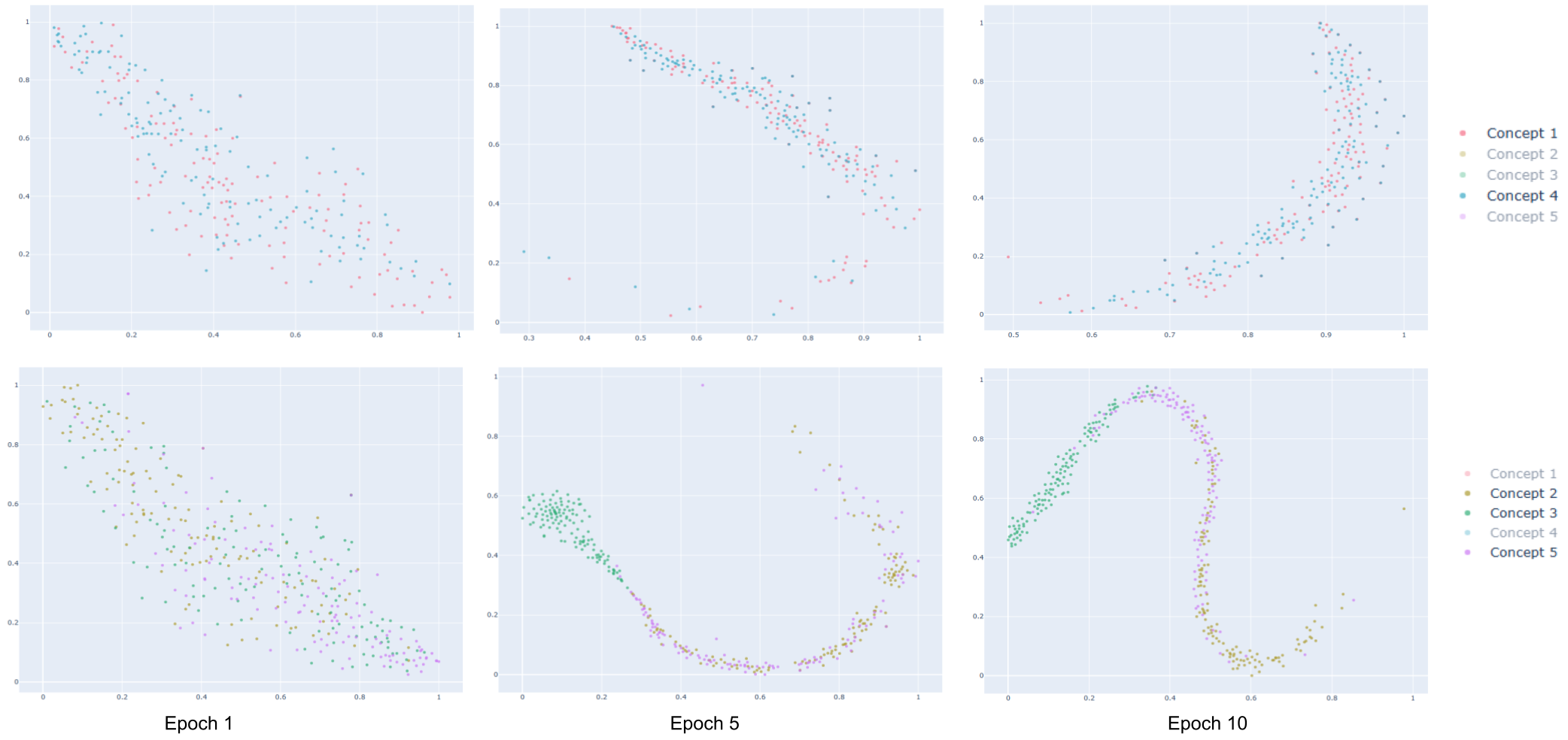}
   \caption{t-SNE Visualization of Concept Vector Clustering. This panel highlights the t-SNE projections of concept vectors across epochs. Concept vectors 1 and 4 overlap significantly, indicating redundancy and focus on less interpretable or irrelevant image areas. In contrast, concept vectors 2, 3, and 5 cluster separately, aligning well with meaningful features and demonstrating effective disentanglement over training epochs.}
   \label{fig:Algo_img_5_2} 
\end{subfigure}
\caption[Algo_img_5_figure_title]{Convergence and Refinement of Concept Vectors Across Epochs on ImageNet instances. This figure showcases the refinement of concept vectors over epochs for two instances, alongside t-SNE visualizations illustrating the clustering behavior of these internal representations. The submodel presented here was trained on panda-like classes~(e.g., Red Panda and Giant Panda) to capture distinctive features such as facial contours and specific markings.}
\label{fig:Algo_img_5_figure}
\end{figure*}

Across other coupled ImageNet examples~(Fig.~\ref{fig:Algo_img_2_figure}, Fig.~\ref{fig:Algo_img_3_figure}, Fig.~\ref{fig:Algo_img_4_figure}, and Fig.~\ref{fig:Algo_img_5_figure}), a consistent pattern emerges: concept vectors consistently capture semantically similar regions of the images, even when the labels vary.
This consistency highlights the ability of the model to disentangle meaningful components of the images.
Detailed explanations for each pair are provided in their respective captions.

\paragraph{Experimental Details and Submodel Differentiation on ImageNet.}
It is important to clarify that each of the experiments on ImageNet (Fig.~\ref{fig:Algo_img_1_figure} - Fig.~\ref{fig:Algo_img_5_figure}) corresponds to a distinct submodel trained independently.
These submodels were specifically designed to operate on subsets of ImageNet classes with closely related semantic meanings, ensuring that the concept vectors are tailored to capture features relevant to a coherent subset of labels.

For instance, the submodel in Fig.~\ref{fig:Algo_img_1_figure} was trained on fish-like classes (e.g., Tench and Goldfish) leveraging five concept vectors to capture semantically meaningful components such as the fish’s body, fins, and broader scene elements. Similarly, in Fig.~\ref{fig:Algo_img_2_figure}, the submodel focused on amphibians, such as Salamander and Triturus vulgaris, with concept vectors isolating parts like the head, body, and background contours.

This approach of using semantically related subclasses for training was consistent across all figures.
In Fig.~\ref{fig:Algo_img_3_figure}, the submodel was trained on antelope-like classes~(e.g., Impala and Gazelle), which allowed the concept vectors to disentangle features such as horns, the general body, and the background.
Similarly, in Fig.~\ref{fig:Algo_img_4_figure}, the focus was on human-centered image classes~(e.g., Bikini and Two-Piece), allowing the model to identify specific body parts and their surrounding context.
In Fig.~\ref{fig:Algo_img_5_figure}, the submodel targeted panda-like classes~(e.g., Red Panda and Giant Panda), capturing distinctive facial features and contours.

In total, we trained 30 submodels, each focusing on a distinct subset of semantically related ImageNet classes.
Among these, we selected and presented five representative submodels.
By focusing on these specific cases, we demonstrated the robustness and adaptability of concept vectors across various semantic contexts.
This methodology underscores the scalability of our approach when applied to a complex and diverse datasets like ImageNet.

\subsubsection{Implementation Level}
%
%
\begin{figure*}[p]
\centering
\begin{subfigure}{0.45\textwidth}
   \includegraphics[width=\linewidth]{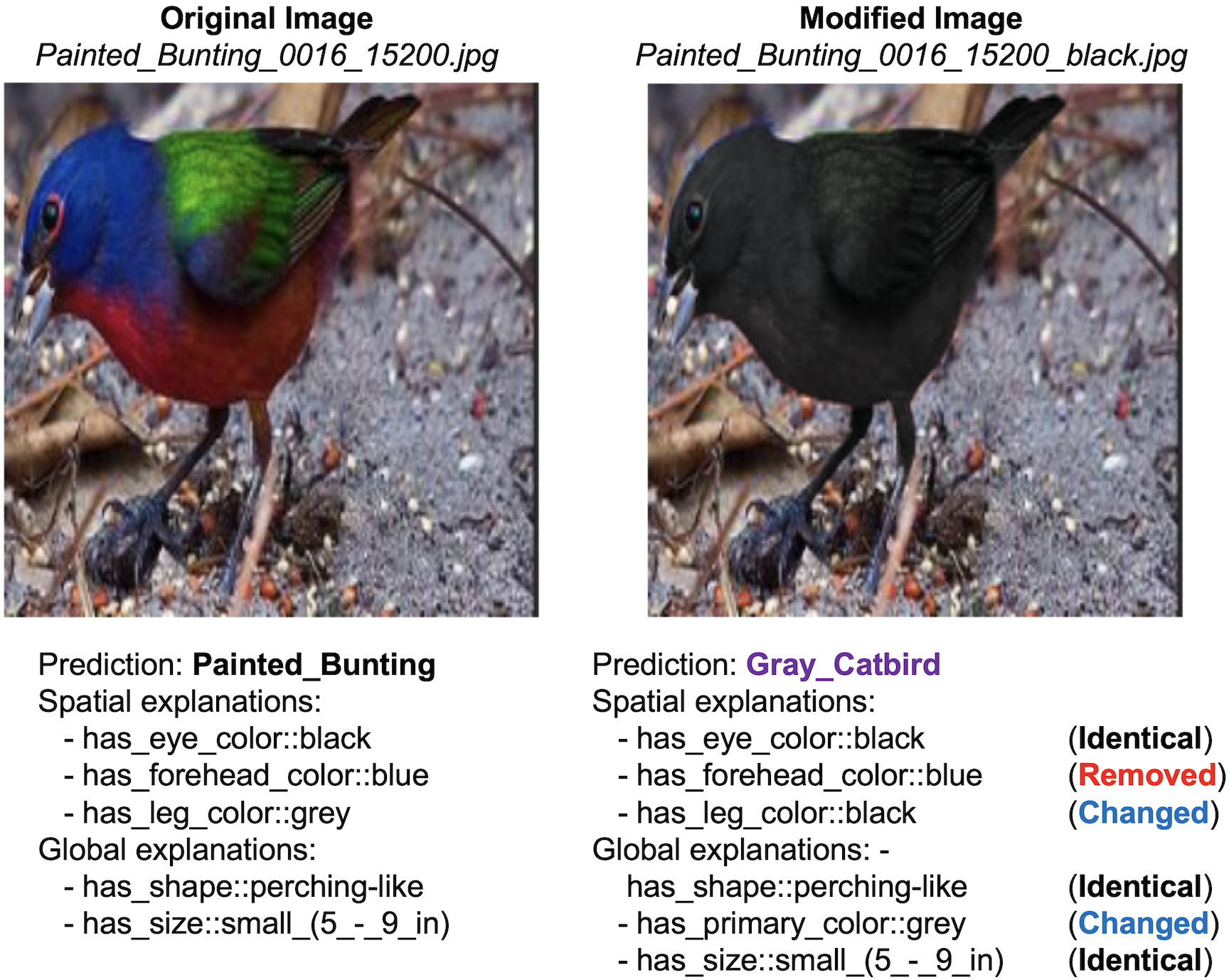}
   \caption{(\textbf{Left}) The original image of \textit{Painted Bunting} and its result, including prediction and explanations (\textbf{Right}) The modified image by coloring the body to black and its altered results. This artificially changing the color of the image input resulted in dramatic shifts by recognizing the concept \texttt{has\_primary\_color::grey} and updating others. Correspondingly, the bird originally classified as a \textit{Painted Bunting} is then classified as a \textit{Gray Catbird}.}
   \label{fig:cf_intervention_samples} 
\end{subfigure}
\begin{subfigure}{0.8\textwidth}
   \includegraphics[width=\textwidth]{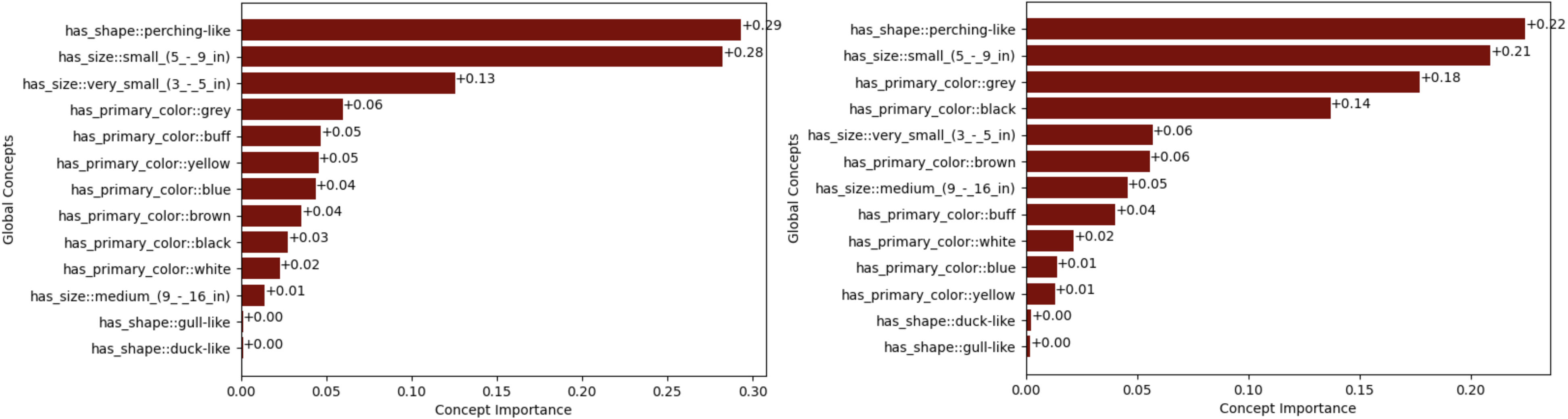}
   \caption{The comparison of global explanations (\textbf{Left}) from the original image (\textbf{Right}) from the modified image}
   \label{fig:cf_intervention_global_exp} 
\end{subfigure}
\begin{subfigure}{0.8\textwidth}
   \includegraphics[width=\textwidth]{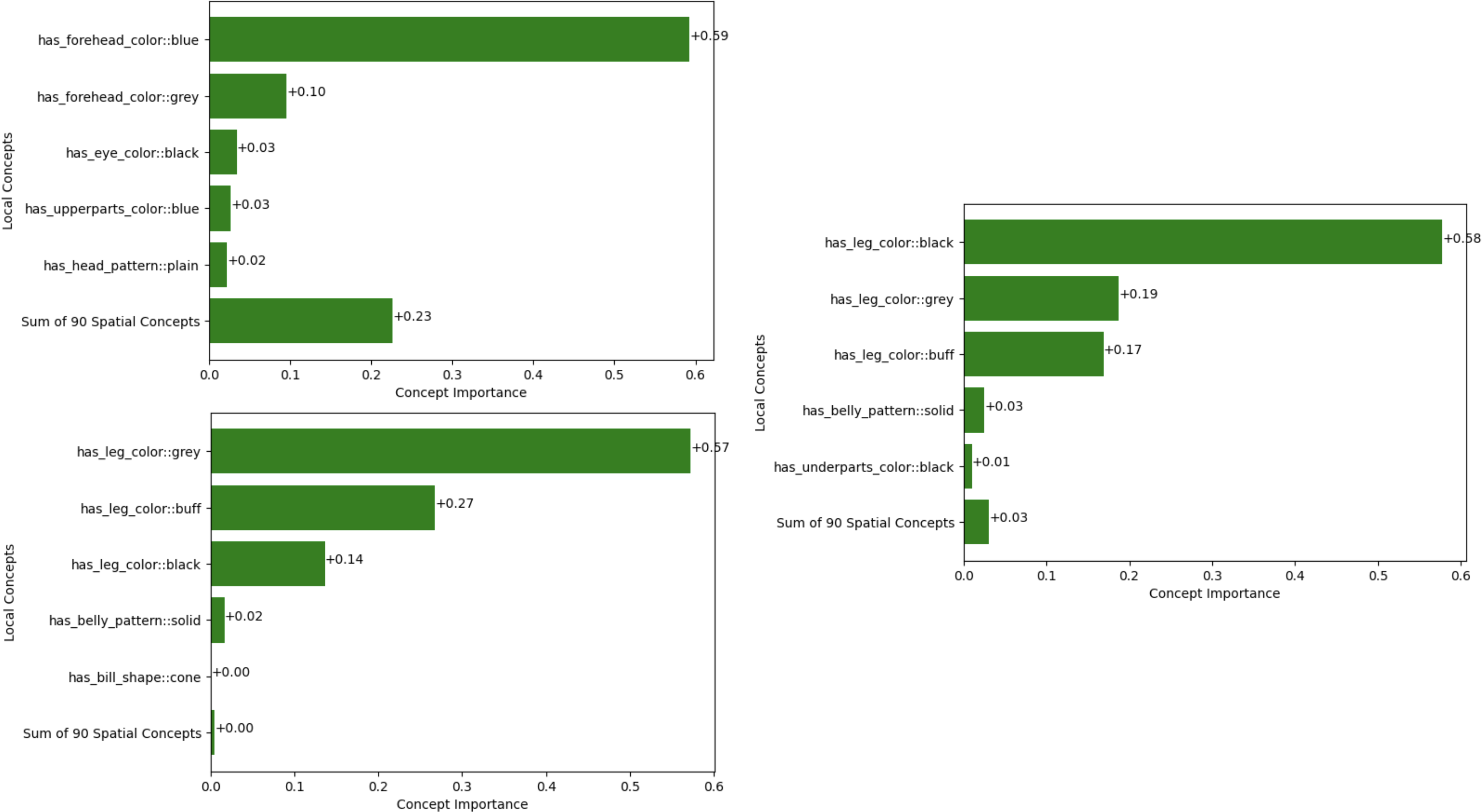}
   \caption{The comparison of spatial explanations (\textbf{Left}) from the original image (\textbf{Right}) from the modified image}
   \label{fig:cf_intervention_spatial_exp} 
\end{subfigure}
\caption[Short caption]{Counterfactual intervention on an exemplary CUB sample. Artificially changing the body color of a \textit{Painted Bunting} to black causes global and spatial concept shifts.}
\label{fig:counterfactual_intervention_example}
\end{figure*}

%

%
\begin{figure*}[t!]
    \centering
    \includegraphics[width=\linewidth]{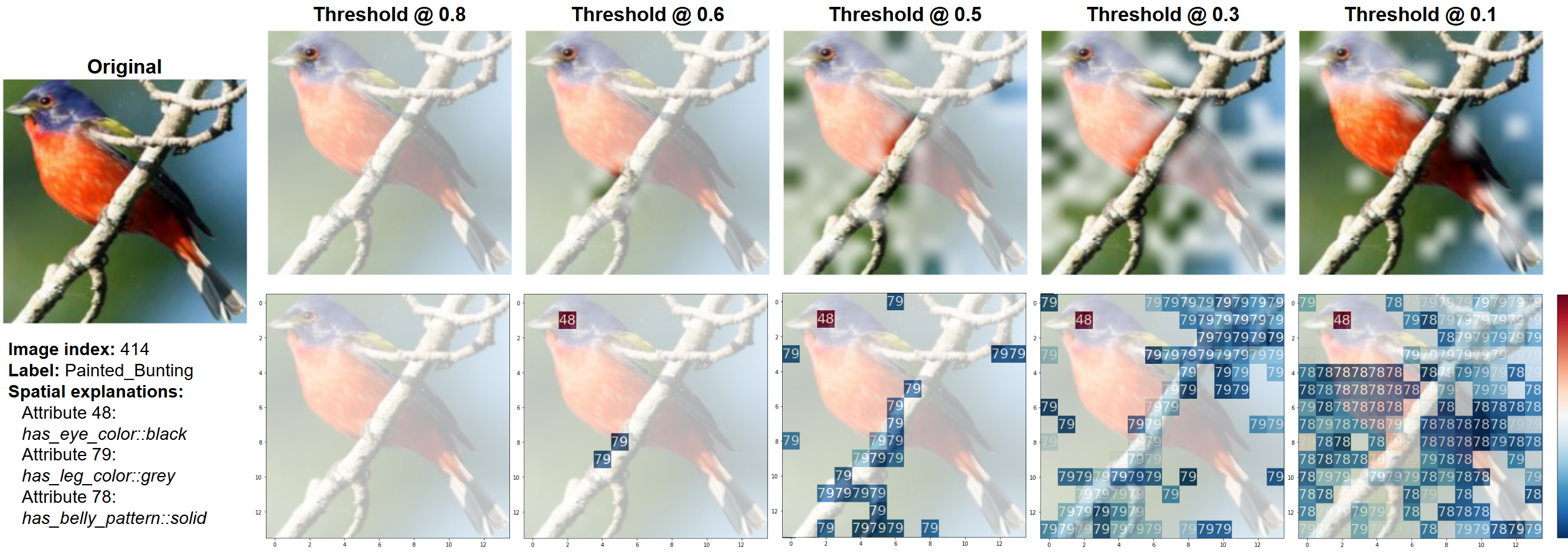}
    \caption{Visualization of thresholding variation for concept activations. 
    The top row shows whether concepts are activated at different thresholds, 
    while the bottom row illustrates the specific attributes and their activation 
    strengths~(red = stronger, blue/transparent = weaker). At 0.8, relevant 
    attributes~(e.g., eye color) are barely visible due to very low intensity, 
    while at lower thresholds~(e.g., 0.3 or 0.1), activations spread excessively 
    and become less interpretable. Empirically, we found that 0.6 provides the 
    most human-interpretable balance, which motivated our choice of this threshold 
    in the main experiments.}
    \label{fig:threshold_cub}
\end{figure*}
\begin{figure*}[t!]
    \centering
    \includegraphics[width=0.90\textwidth]{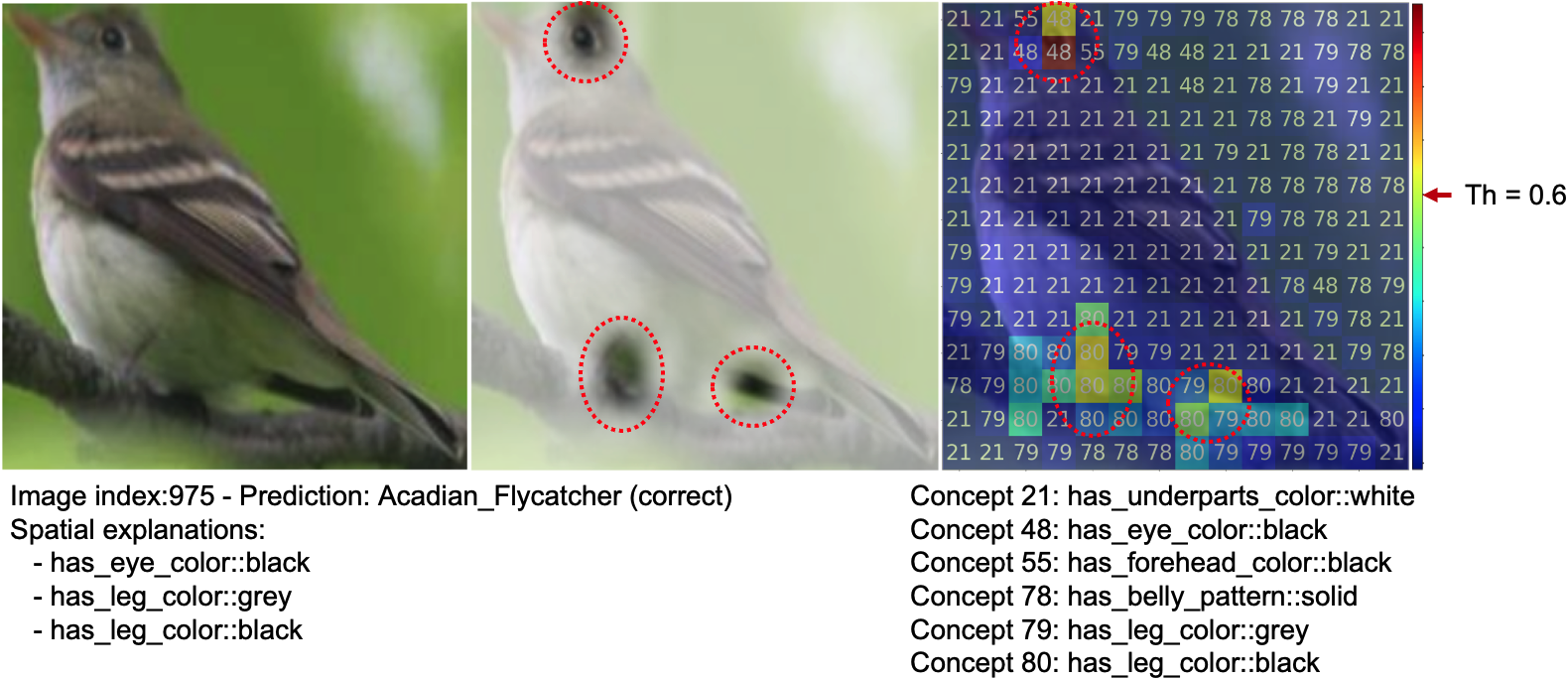}
    \caption{The detail of the example in Fig.~\ref{fig:3_implementation} in the main text. \textbf{Left}: The original image. \textbf{Middle}: The resultant image highlighting the activated patches. The active patches are displayed as their original image, while the rest are blurred. \textbf{Right}: The image shows the resultant concept localization with a 14 × 14 grid overlay. Each cell is color-coded with a colorbar, where red indicates a high concept contribution score, and blue indicates its low value. Each number in the cell indicates the corresponding concept of the highest concept importance value. The threshold determines whether each patch is active, contributing to the image in the middle.}
    \vspace{-3mm}
    \label{fig:concept_localization_example_01}
\end{figure*}
\begin{figure*}[t!]
\centering
\begin{subfigure}[t!]{0.98\textwidth}
   \includegraphics[width=1\linewidth]{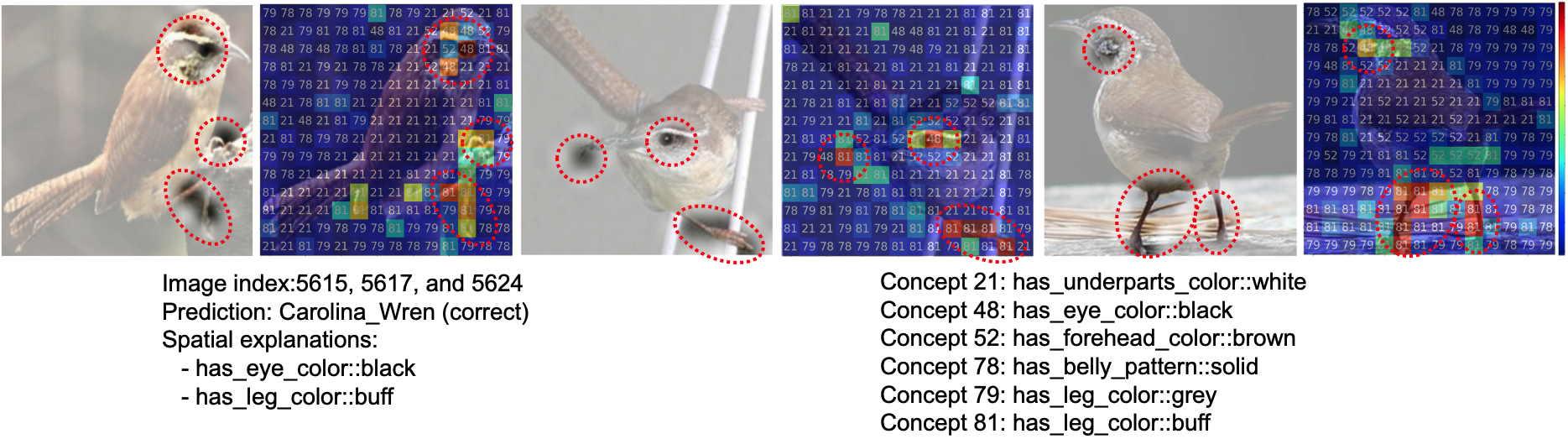}
   \caption{The set of examples of class \textit{`Carolina Wren'}}
   \label{fig:concept_localization_example_02} 
\end{subfigure}
\begin{subfigure}[t!]{0.98\textwidth}
   \includegraphics[width=1\linewidth]{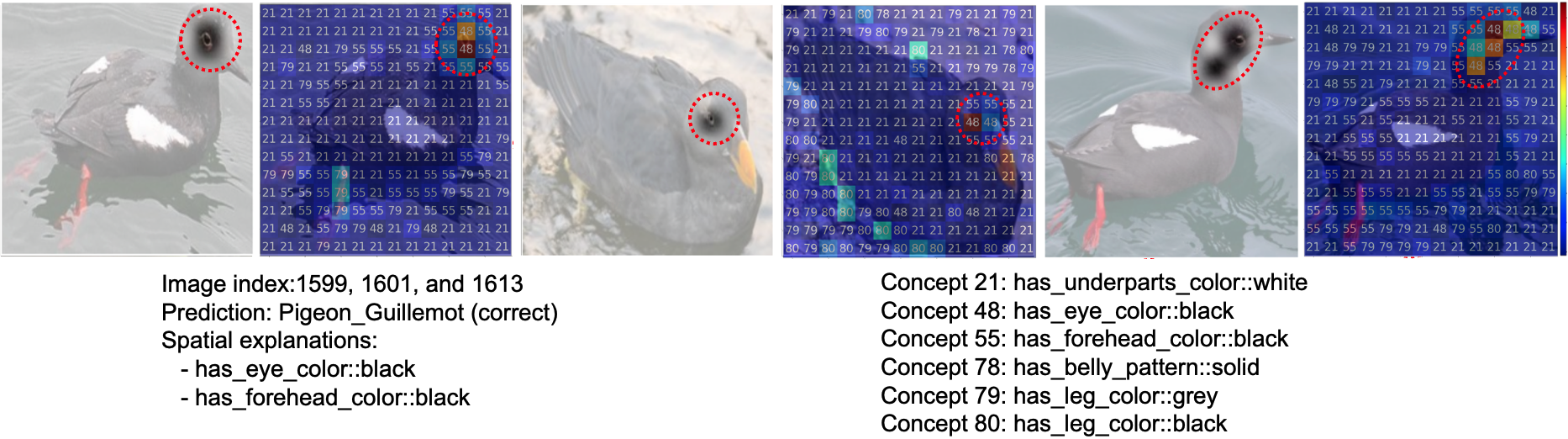}
   \caption{The set of examples of class \textit{`Pigeon Guillemot'}}
   \label{fig:concept_localization_example_03} 
\end{subfigure}
\begin{subfigure}[t!]{0.98\textwidth}
   \includegraphics[width=1\linewidth]{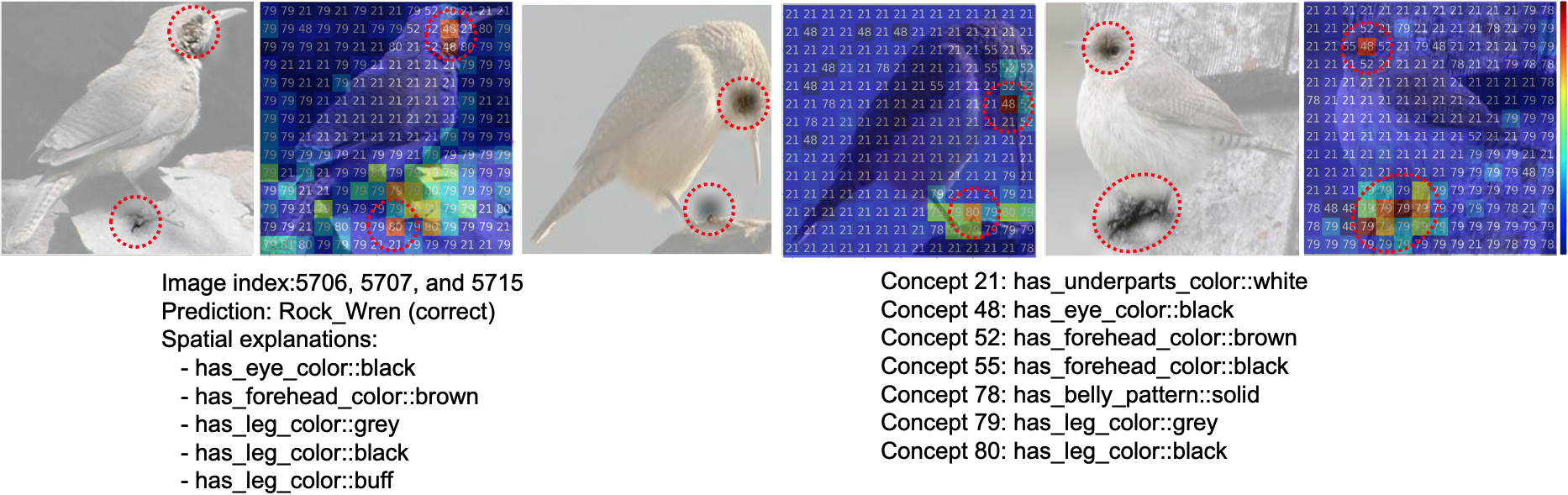}
   \caption{The set of examples of class \textit{`Rock Wren'}}
   \label{fig:concept_localization_example_04} 
\end{subfigure}
\caption[Two numerical solutions]{The additional concept localization results for three bird classes (Carolina Wren, Pigeon Guillemot, and Rock Wren) on CUB, highlighting specific regions associated with various concepts. Overall, the samples in the same class included consistent spatial explanations corresponding to semantically meaningful regions, highlighted by the dashed red circles.}
\label{fig:concept_localization_examples}
\end{figure*}

\paragraph{Training Setups.}
For the analysis, we trained an additional simple model, which follows. 
We used the frozen ViT-S as a feature extractor and combined it with our Bi-ICE module. 
The model leveraged the ground-truth spatial annotation only during the training, which is the key difference from the setup in Sec.~\ref{appsubsec:computational}.
This is because the concept localization is mainly determined by the ground-truth spatial annotations.
For the hyperparameter setting, we used the same setup in Table~\ref{tab:hyperparams_computational_implementation}.

\paragraph{Sensitivity Analysis of Thresholding the Most Activated Patches.}
Figure~\ref{fig:threshold_cub} illustrates how different threshold values affect the visualization of concept activations. The top row shows whether a concept is activated or not, while the bottom row highlights which attributes are activated and the corresponding strength of their activations~(with red indicating stronger activations and blue/transparent indicating weaker ones). At a high threshold of 0.8, for example, the eye region~(Attribute 48: \textit{has\_eye\_color::black}) is technically activated, but the intensity is so low that it is barely visible in the heatmap. At lower thresholds~(e.g., 0.5, 0.3 or 0.1), the activations spread excessively, leading to blurred and ambiguous visualizations that are less interpretable. Empirically, we found that a threshold of 0.6 provides the clearest and most human-interpretable balance: Relevant regions become visible without the excessive diffusion observed at lower thresholds. Therefore, we chose \emph{0.6} as the default threshold for our main experiments.

\paragraph{Additional Figures and Results.}
Firstly, 
Fig.~\ref{fig:concept_localization_example_01} depicts the detail of the example in Fig.~\ref{fig:3_implementation} in the main text.

Furthermore, Fig.~\ref{fig:concept_localization_examples} shows the additional concept localization results for three bird classes (Carolina Wren, Pigeon Guillemot, and Rock Wren), highlighting specific regions associated with various concepts.
Concept 48, has\_eye\_color::black, consistently aligns with eye regions across all samples, demonstrating accurate identification of black eyes.
Similarly, Concepts 79 (has\_leg\_color::grey) and 80 (has\_leg\_color::black) are localized to leg regions, emphasizing the module's ability to localize and differentiate leg colors.
These results demonstrate our module's capacity for localized feature detection, aiding interpretability and classification accuracy.

\section{Use cases}
\subsection{Use Case 1: Counterfactual Intervention on CUB.}
\label{appsubsec:usecase1}
Fig.~\ref{fig:counterfactual_intervention_example} demonstrates counterfactual intervention on an exemplary CUB sample.
The caption of the figure includes the details.

\subsection{Use Case 2: Empirical Evidence for the Number of Concepts.}
\label{appsubsec:usecase2}
The t-SNE visualizations from the ImageNet experiments~(Fig.~\ref{fig:Algo_img_1_3}, Fig.~\ref{fig:Algo_img_2_2}, Fig.~\ref{fig:Algo_img_3_2}, Fig.~\ref{fig:Algo_img_4_2}, and Fig.~\ref{fig:Algo_img_5_2}) provide empirical evidence for determining the optimal number of concepts.
By examining the results across different instance types—such as (i)~fish-like classes~(Fig.~\ref{fig:Algo_img_1_figure}), (ii)~amphibians~(Fig.~\ref{fig:Algo_img_2_figure}), (iii)~animals with horns~(Fig.~\ref{fig:Algo_img_3_figure}), (iv)~human-centered images~(Fig.~\ref{fig:Algo_img_4_figure}), and (v)~panda-like instances~(Fig.~\ref{fig:Algo_img_5_figure})—we observed a recurring pattern: out of the five concepts used in training~(except amphibian case in Fig.~\ref{fig:Algo_img_2_figure}), typically two concepts were redundant.
These redundant concepts either focused on irrelevant areas of the image or were non-interpretable~(e.g., capturing only a few random dots on the image).
Meanwhile, the remaining three concepts consistently captured relevant and human-interpretable features.

To further validate this, we extended this analysis by varying the number of concepts used in the training.
When we reduced the number of concepts to four, the same pattern persisted: two concepts remained redundant, while two concepts were useful.
Similarly, with three concepts, one was redundant while two remained meaningful.
In contrast, when we increased the number of concepts to ten, the number of redundant concepts grew to four, leaving only six as relevant.

These observations suggest that our framework can serve as a tool for determining the ``sweet spot" for the number of concepts.
For most experiments, setting the number of concepts to five provided a balanced trade-off between interpretability and representational capacity, avoiding unnecessary redundancy in the concept space.

By integrating these insights, our analysis underscores the utility of concept vectors in understanding not only the behavior of the model but also the fine-tuning of the model configurations for optimal performance and interpretability.

\section{User Study}
\label{app:user_study}

\subsection{User Study Design}
\label{appsubsec:user_study_design}
To evaluate the interpretability of unsupervised concepts on the CUB-200-2011 dataset, we conducted a user study under unsupervised setting. Specifically, we set the number of concept vectors to $K=20$ and did not provide any human expert knowledge during concept discovery. All other model configurations followed the same setup described in the main experiments. For each concept vector, we then extracted the top-5 most highly activated samples, which serve as the clearest evidence of what the concept vector has captured in the absence of supervision.

\begin{figure*}[t]
    \centering
    \includegraphics[width=0.98\textwidth]{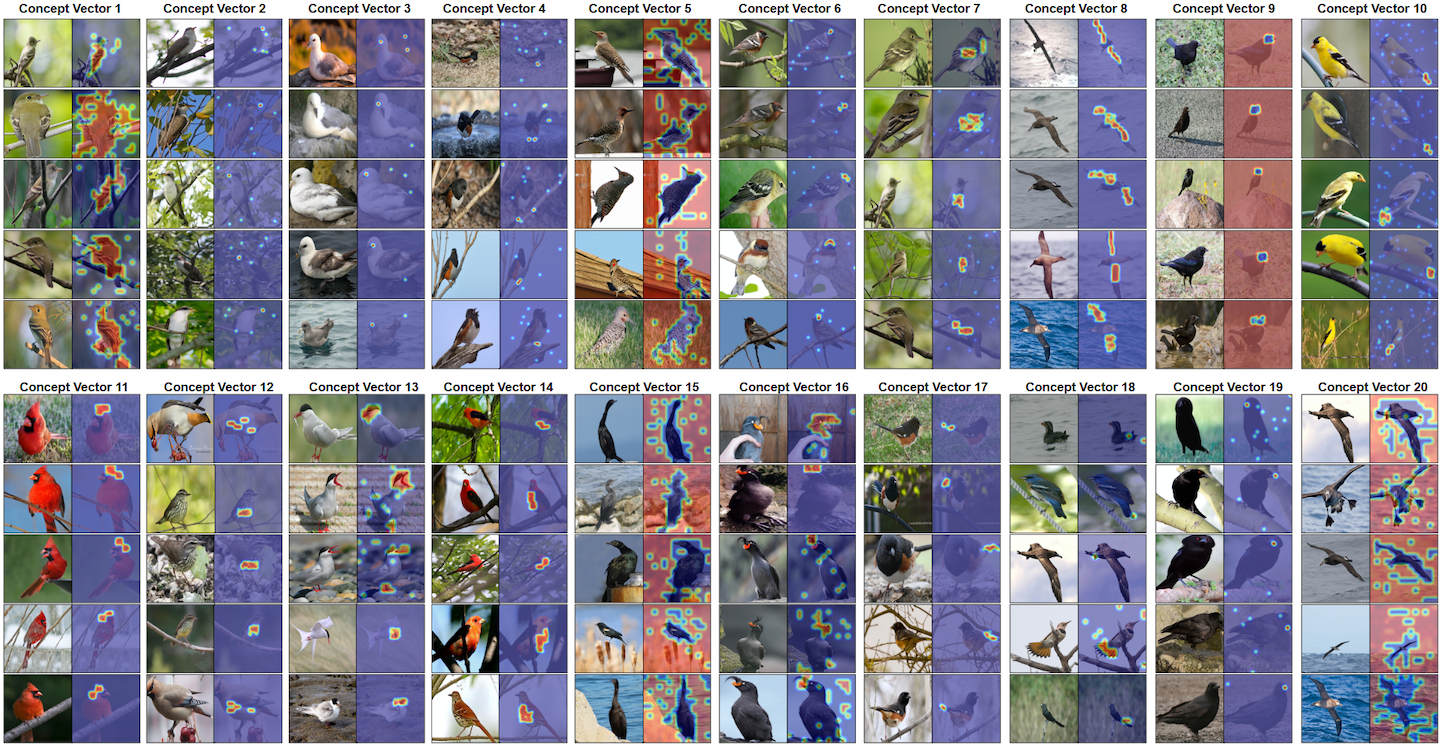}
    \caption{Top-5 activated samples for twenty discovered concept vectors in the unsupervised setting on CUB200 dataset. Each column corresponds to one concept vector, and each row shows images that most strongly activate the corresponding vector. Highlighted regions~(blue = low focus, red = stronger focus) indicate the spatial contributions to the activation. These examples illustrate the raw outputs of unsupervised concept discovery, which motivate the need for a user study: while the visual patterns appear meaningful, the crucial question is whether humans reach consistent and shared interpretations of these concepts.}
    \label{fig:top5_activated}
\end{figure*}

A natural idea would be to ask participants to freely describe each concept by examining these top-5 activated samples~(see Figure~\ref{fig:top5_activated}). While such visualizations can reveal recurring patterns, open-ended descriptions are difficult to compare and aggregate among participants. To address this and following the approach in~\cite{wang2023learning}, we provide participants with a controlled vocabulary derived from the ground-truth annotations of the CUB dataset, allowing them to select relevant terms.

\begin{figure*}[ht!]
    \centering
    \includegraphics[width=0.85\textwidth]{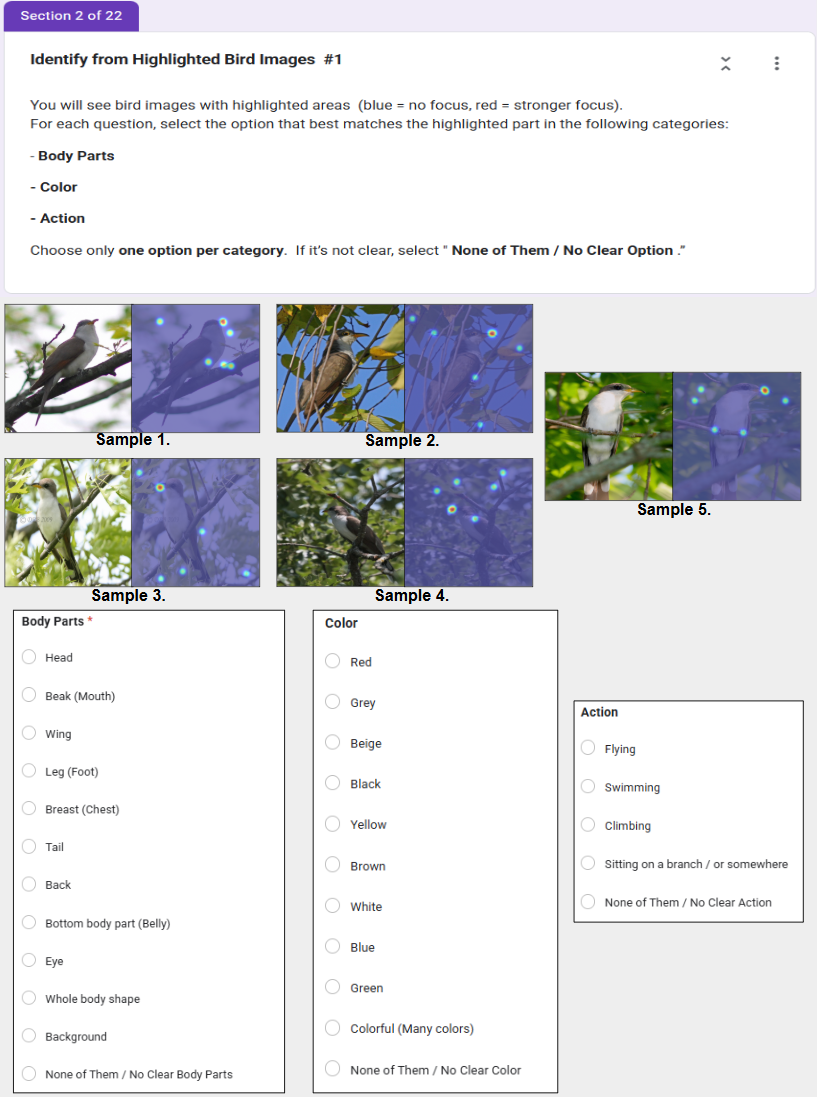}
    \caption{Example survey interface shown to participants. 
    Each question presents bird images with highlighted regions~(blue = low focus, red = stronger focus). Participants are asked to select 
    the most appropriate label in three categories: Body Parts, Color, and Action, 
    or choose ``None of Them / No Clear Option'' when no clear match is present.}
    \label{fig:survey_sample}
\end{figure*}

In each question, participants are shown bird images with highlighted regions, where red indicates stronger model attention. They are asked to assign the most appropriate labels in three categories: \emph{body parts, action}, and \emph{color}. The body part set includes head, beak, wing, leg, breast, tail, back, belly, eye, whole body shape, background, and a ``none'' option. The action set consists of flying, swimming, climbing, sitting, and ``none,'' while the color set covers red, gray, beige, black, yellow, brown, white, blue, green, colorful and ``none.'' Participants must select exactly one option per category, or ``none'' if no clear match is visible. Each participant answers twenty such questions. Figure~\ref{fig:survey_sample} illustrates the survey interface.

This design provides a structured and consistent protocol for evaluating concept vectors in an unsupervised setting. Inclusion of ``none'' options avoids bias from forced choices. More importantly, aggregating responses across annotators enables us to assess whether the discovered concepts support consistent human interpretation. High agreement would suggest that the concepts are coherent and semantically meaningful, while low agreement would reveal ambiguity and highlight the limitations of interpretability in the unsupervised setting.


\subsection{Metrics}
\label{appsubsec:user_study_metric}

We designed the following two metrics to evaluate learned concepts based on the user study.

\subsubsection{\texorpdfstring{$\chi^2$}{TEXT} tests of independence (per category).}
To test whether participants' vocabulary choices were systematically dependent on the concept being annotated, we used a $\chi^2$ test of independence~\cite{cohen2013statistical} for each category~(body part, color, action).

For each category, we created a contingency table with rows corresponding to the 20 concepts and columns corresponding to the vocabulary terms~(e.g., head, wing, tail for body part). Each entry in the table represents the number of participants who chose a given term for a given example.

The test evaluates whether the observed distribution of responses $O_{i,j}$ for cell $(i, j)$ differs from the distribution expected under independence $E_{i,j}$. The $\chi^2$ statistics is:
\begin{equation*}
    \chi^2 = \sum_{i,j} \frac{(O_{i,j}-E_{i,j})^2}{E_{i,j}},
\end{equation*}
with degrees of freedom $(R-1)(C-1)$, where $R$ and $C$ are the number of rows and columns in the contingency table. 
A large $\chi^2$ statistic with a small $p$-value indicates that the choice of vocabulary term is not independent of the concept; in other words, different concepts elicit different label distributions.

Furthermore, to quantify the strength of association, we used Cramer’s $V$~\cite{cohen2013statistical}:
\begin{equation*}
    V = \sqrt{\frac{\chi^2/n}{\operatorname{min}(R-1,C-1)}},
\end{equation*}
where $n$ is the total number of responses. 
Values closer to $0$ indicate weak associations, while values closer to $1$ indicate strong associations. 

Finally, beyond global significance, we inspected standardized residuals:
\begin{equation*}
    z = \frac{O_{i,j}-E_{i,j}}{\sqrt{E_{i,j}}}
\end{equation*}
to identify which specific (concept $\times$ vocabulary) cells contributed most. 
Large positive residuals indicate terms chosen disproportionately often for that example, while negative residuals indicate terms chosen less often than expected.

\subsubsection{Mutual Information between Concepts}

While $\chi^2$ tests assess global dependencies between concepts and responses within each category, they do not capture the pairwise similarity or redundancy between examples. 
To address this, we adapted the Mutual Information between Concepts~(MIC) metric from~\cite{wang2023learning}.

In our configuration, each of the 20 examples is treated as a concept.
For each concept, we collected the distribution of participant responses in a given category. 
Here, we consider body part category only.
This gives us a categorical random variable $L_{i}$ for each concept $i$.

For each pair of concept examples $i, j$, we computed the mutual information between $L_{i}$ and $L_{j}$:
\begin{equation*}
    \operatorname{MIC}(i,j) \triangleq \operatorname{MI}(L_{i};L_{j}) = \sum_{u,v} p(u,v)\log\frac{p(u,v)}{p(u)p(v)}, 
\end{equation*}
where $p(u,v)$ is the joint probability of participant assigning vocabulary $u$ to concept $i$ and vocabulary $v$ to another concept $j$. 

We report MIC in bits, which reflects the amount of information one example’s labels provide about another. For example, 1 bit corresponds to a binary distinction (yes/no).
To determine whether observed MIC values were greater than expected by chance, we used a permutation test: participant labels were randomly shuffled across concepts $10,000$ times, and MIC was recomputed for each shuffle. The permutation $p$-value is the proportion of shuffled MIC values greater than or equal to the observed MIC.

Low MIC indicates concepts are distinct, with participants assigning very different labels. While high MIC indicates redundancy, with participants using similar vocabularies for both concepts.

\subsection{Results}

\subsubsection{\texorpdfstring{$\chi^2$}{TEXT} Analysis}
\begin{figure}[ht!]
    \centering
    \includegraphics[width=0.98\columnwidth]{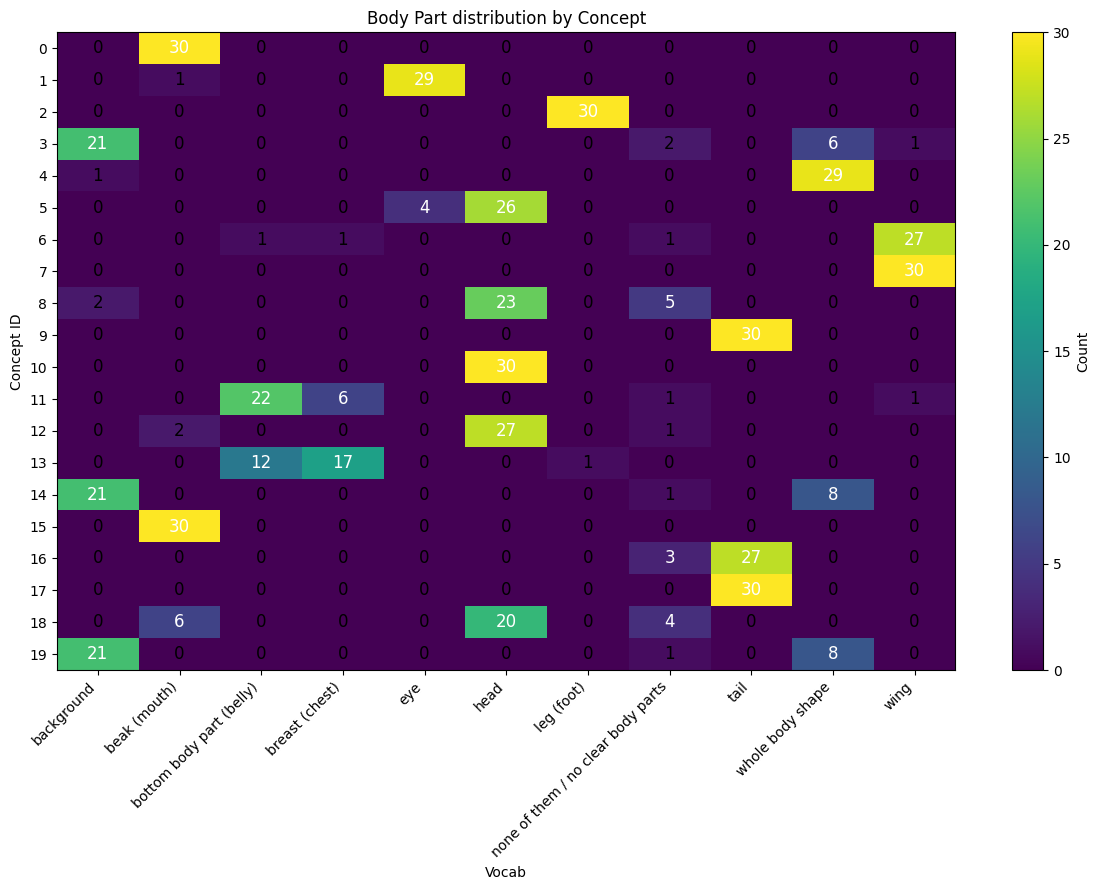}
    \caption{
    Contingency heatmap of body part responses by concept example. Most concepts are dominated by a single vocabulary term (e.g., \emph{beak}, \emph{wing}, \emph{tail}), producing highly concentrated response patterns. This strong alignment accounts for the extremely large $\chi^2$ statistic ($\chi^2(190) = 4171.64$) and effect size~(Cramer's $V = 0.82$).
    }
    \label{fig:user_study_body_map}
\end{figure}
\begin{figure}[ht!]
    \centering
    \includegraphics[width=0.98\columnwidth]{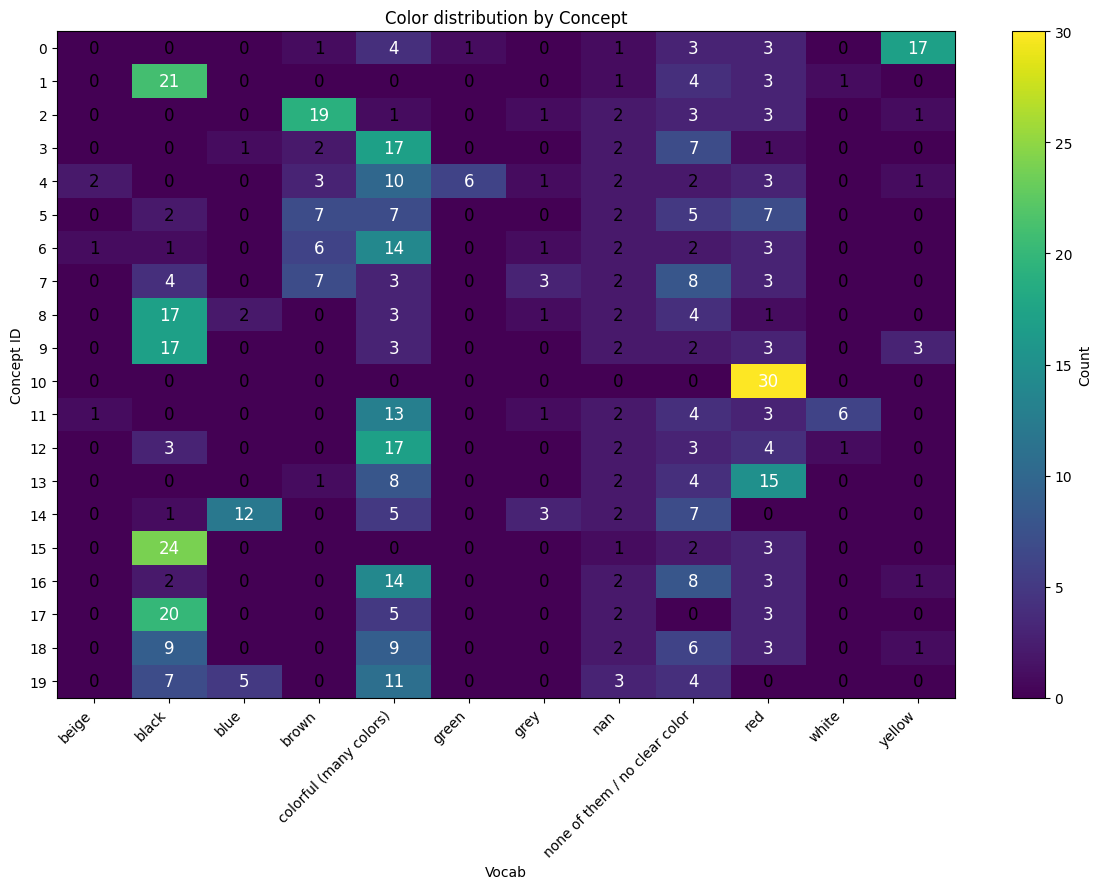}
    \caption{
    Contingency heatmap of color responses by concept example. While some concepts are clearly associated with single terms (e.g., \emph{black}, \emph{blue}), others show more distributed choices across multiple terms~(e.g., \emph{grey}, \emph{brown}). This greater variability explains the moderate effect size ($\chi^2(209) = 1308.05$, Cramer’s $V = 0.41$).
    }
    \label{fig:user_study_color_map}
\end{figure}
\begin{figure}[ht!]
    \centering
    \includegraphics[width=0.98\columnwidth]{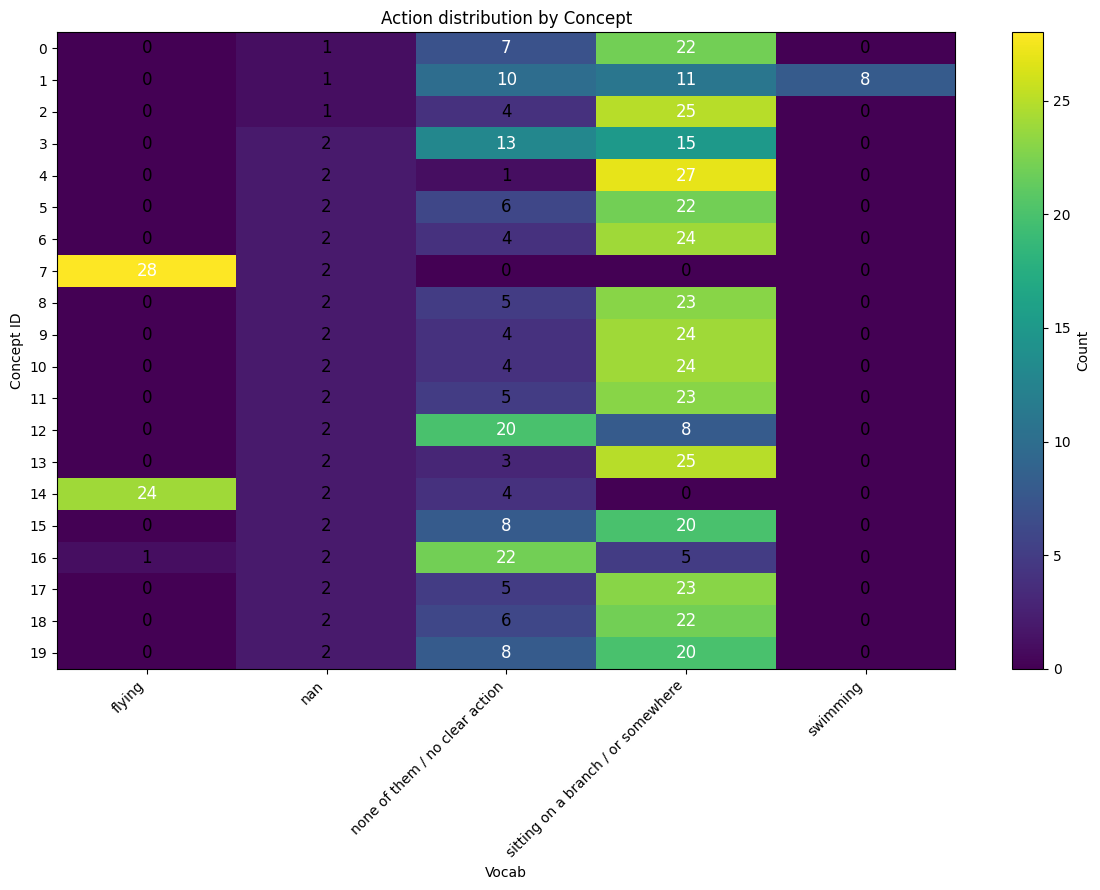}
    \caption{
    Contingency heatmap of action responses by concept example. Many concepts are dominated by \emph{sitting on a branch}, while others show clear associations with \emph{flying} or \emph{swimming}. Some ambiguity is present~(e.g., overlap with \emph{climbing}), producing a strong but not maximal association ($\chi^2(76) = 777.16$, Cramer’s $V = 0.54$).
    }
    \label{fig:user_study_action_map}
\end{figure}
\begin{table}[t!]
\centering
\caption{$\chi^2$ test results per category. df indicates degree of freedom.}
\resizebox{0.98\linewidth}{!}{%
\begin{tabular}{lrrrr}
    \toprule
    Category   & $\chi^2$ & df  & $p$ (asymptotic) & Cramer's $V$ \\
    \midrule
    Body part  & 4171.64  & 190 & $< 1\times 10^{-300}$ & 0.82 \\
    Color      & 1308.05  & 209 & $2.8 \times 10^{-158}$ & 0.41 \\
    Action     & 777.16   & 76  & $9.1 \times 10^{-117}$ & 0.54 \\
    \bottomrule
\end{tabular}
}
\label{tab:chi2_results}
\end{table}

Figures~\ref{fig:user_study_body_map},~\ref{fig:user_study_color_map} and~\ref{fig:user_study_action_map} depict the contingency heatmap for categories: \emph{Body Parts, Color} and \emph{Action}. 
Body part terms show the strongest discriminability, with most concepts dominated by a single label~(e.g., beak, tail), producing near one-to-one mappings. Action terms also show structured patterns, with many examples dominated by sitting on a branch or flying, though a few cases exhibit ambiguity. Color terms are comparatively more diffuse, with some examples clearly tied to a single color~(e.g., black, blue) but others displaying overlap across multiple terms. Together, these maps illustrate that body part cues provide the most robust separation, action terms offer strong but somewhat less distinct signals, and color terms are moderately discriminative.

Table~\ref{tab:chi2_results} summarizes the results, including $\chi^2$ statistic, degrees of freedom, p-values~(asymptotic), and effect sizes~(Cremer's $V$).
The $\chi^2$ tests revealed strong and highly significant dependencies between concept identity and vocabulary choices across all categories.
\begin{itemize}
    \item Body part yielded the largest $\chi^2$ statistic with Cramer's $V = 0.82$, indicating an extremely strong association. This means participants' body part choices were almost perfectly aligned with specific examples, reflecting the high diagnostic power of morphological cues. 
    \item Action also showed a strong association, demonstrating that behavioral descriptors consistently differentiated examples. 
    \item Color produced a smaller but still statistically robust effect. Although color was less discriminative than body part or action, it still contributed significantly to distinguishing among examples.
\end{itemize}

Taken together, these results confirm that all three vocabularies carried meaningful information for distinguishing examples, with body part as the most discriminative category, followed by action and then color. The effect sizes show that participants were not guessing, but instead made systematic and interpretable choices aligned with human-recognizable features.

\subsubsection{Mutual Information between Concepts}
\begin{figure}
    \centering
    \includegraphics[width=0.98\columnwidth]{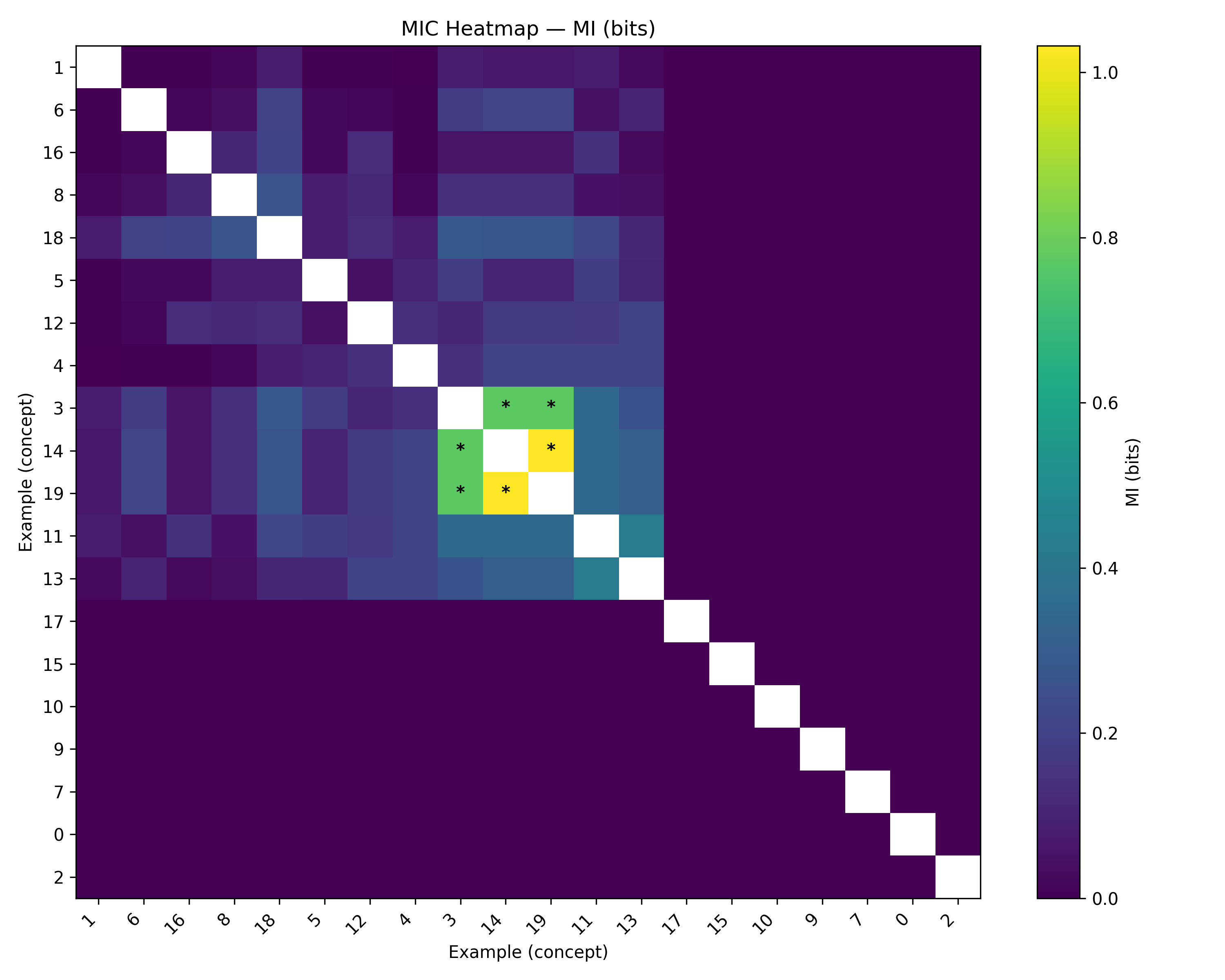}
    \caption{MIC heatmap (Mutual Information in bits) between concept examples using \textbf{Body part} annotations. 
    Most example pairs show low MI values ($<0.3$ bits, dark cells), indicating that body part labels (e.g., \emph{beak}, \emph{wing}, \emph{tail}) provide highly distinct information across examples. 
    A localized high-MI cluster (yellow and green cells with significance markers) emerges, corresponding to examples that participants consistently labeled as \emph{background} or \emph{whole body parts}. 
    This pattern highlights the dual role of body part vocabulary: it achieves strong global discriminability while also revealing interpretable redundancy in fallback categories.}
    \label{fig:user_study_body_part_MIC_NMI}
\end{figure}

The heatmap~(Fig.~\ref{fig:user_study_body_part_MIC_NMI}) presents the pairwise mutual information~(MI) between the 20 concepts using only Body part annotations.

Most cells in the matrix are dark~(low NMI $< 0.3$), indicating that body part responses for different examples are generally distinct. This aligns with the contingency map, where body part terms~(e.g., beak, tail, wing) mapped sharply to specific examples. 
A cluster in the matrix shows significantly elevated NMI values~(0.7–1.0, yellow cells with significance markers). On inspection, these examples were consistently labeled as \emph{background} or \emph{whole body parts}. This redundancy is desirable rather than problematic: it reflects that participants reserved these labels for examples where no finer morphological feature stood out. In other words, overlap here does not reflect confusion, but rather systematic use of "catch-all" terms.

Thus, this pattern strengthens the interpretability of the body part vocabulary. It shows not only that most examples are distinct but also that redundancy, where it exists, is aligned with intuitive and desirable fallback labels.

\end{document}